\documentclass{article}
\usepackage[table]{xcolor}%
\usepackage{placeins}
\usepackage{microtype}
\usepackage{graphicx}
\usepackage{subfigure}
\usepackage{booktabs} %
\usepackage{tikz}
\usetikzlibrary{fit}\usetikzlibrary{backgrounds}
\usepackage{hyperref}

\usepackage[accepted]{icml2025}
\definecolor{softgreen}{rgb}{0.5, 0.8, 0.5}  
\definecolor{softergreen}{rgb}{0.85, 0.95, 0.85}

\usepackage{amsmath}
\usepackage{amssymb}
\usepackage{mathtools}
\usepackage{amsthm}

\usepackage[capitalize,noabbrev]{cleveref}

\theoremstyle{plain}
\newtheorem{theorem}{Theorem}[section]

\newtheorem{lemma}[theorem]{Lemma}

\theoremstyle{definition}

\theoremstyle{remark}

\usepackage{bm} 

\newcommand{\norm}[1]{\left\|#1\right\|}

\def\Pr{\mathrm{P}}

\def\Exp{\mathbb{E}}

\def\R{\mathbb{R}}

\usepackage[textsize=tiny]{todonotes}

\newcommand{\ourMethod}{\textit{Mahalanobis++}}
\newcommand{\rourMethod}{\textit{relative Mahalanobis++}}

\icmltitlerunning{Mahalanobis++: Improving OOD Detection via Feature Normalization}
\begin{document}

\twocolumn[
\icmltitle{Mahalanobis++: Improving OOD Detection via Feature Normalization
}

\icmlsetsymbol{equal}{*}

\begin{icmlauthorlist}
\icmlauthor{Maximilian Müller}{yyy}
\icmlauthor{Matthias Hein}{yyy}
\end{icmlauthorlist}

\icmlaffiliation{yyy}{University of Tübingen and Tübingen AI Center}

\icmlcorrespondingauthor{Maximilian Müller}{maximilian.mueller@wsii.uni-tuebingen.de}

\icmlkeywords{Machine Learning, ICML}

\vskip 0.3in
]

\printAffiliationsAndNotice{} %

\begin{abstract}
Detecting out-of-distribution (OOD) examples is an important task for deploying reliable machine learning models in safety-critial applications. 
While post-hoc methods based on the Mahalanobis distance applied to pre-logit features are among the most effective for ImageNet-scale OOD detection, their performance varies significantly across models. We connect this inconsistency to strong variations in feature norms, indicating severe violations of the Gaussian assumption underlying the Mahalanobis distance estimation.
We show that simple $\ell_2$-normalization of the features mitigates this problem effectively, aligning better with the premise of normally distributed data with shared covariance matrix.
Extensive experiments on 44 models across diverse architectures and pretraining schemes show that $\ell_2$-normalization improves the conventional Mahalanobis distance-based approaches significantly and consistently, and outperforms other recently proposed OOD detection methods.
Code is available at \href{https://github.com/mueller-mp/maha-norm}{github.com/mueller-mp/maha-norm}.

\end{abstract}

\section{Introduction}\label{sec:intro}

\begin{figure}
    \centering
    \includegraphics[width=1.\linewidth]{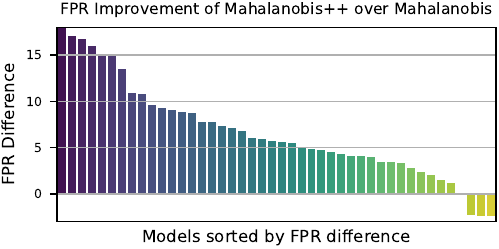}

    \caption{\textbf{Normalizing features improves OOD detection with the Mahalanobis distance consistently.} Shown is 
    the difference in false-positive rate at true positive rate of 95\% between unnormalized and normalized features for 44 ImageNet models, averaged over five OOD datasets of the OpenOOD benchmark.}
    \label{fig:teaser-barplot}
\vskip -0.1in
\end{figure}

\begin{figure*}[t]
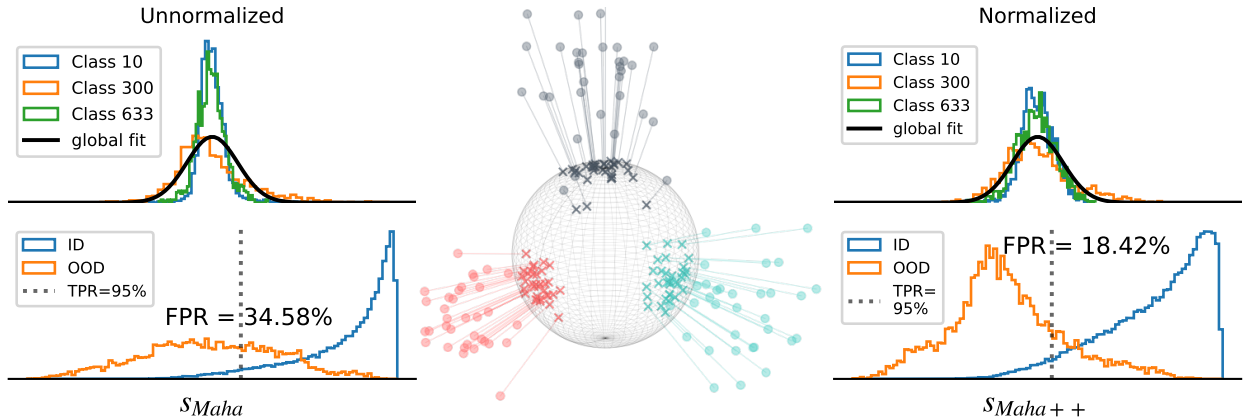

    \centering
    \vspace{-5mm}
\include{teaser}
    \vskip -4mm
    \caption{\textbf{Mahalanobis++:} We illustrate how to improve Mahalanobis-based OOD detection. \textbf{Left: } For unnormalized features, assuming a shared covariance matrix for all classes leads to suboptimal OOD detection (bottom) with the Mahalanobis score.
    \textbf{Center: } Normalizing the features, i.e. projecting them onto the unit sphere mitigates this problem effectively. \textbf{Right: } After normalization, the fit of the shared covariance matrix is tighter for all classes, 
    leading to improved OOD detection as in- and out-distribution are better separated. Shown are the \ourMethod{} scores for a pretrained ConvNextV2-L on NINCO, which achieves a new state-of-the-art FPR of 18.4\% (see Tab.~\ref{tab:fpr-big-NINCO}).}

    \label{fig:teaser-overview}
\end{figure*}

Deep neural networks have demonstrated remarkable performance across a variety of real-world tasks. However, when faced with inputs that fall outside their training distribution, they can behave unpredictably and even result in high-confidence predictions  \cite{hendrycks2017MSP, HeiAndBit2019}. These so-called out-of-distribution (OOD) inputs are often misclassified with high confidence as belonging to the in-distribution (ID) classes, creating significant risks for real-world deployments. 
OOD detectors aim to identify and reject such anomalous inputs — potentially prompting human intervention, transitioning to a safe state, or declining to provide a prediction — while still allowing genuine ID samples to pass through normally.
OOD detection methods are commonly divided into methods that require modifications to the training process and so-called post-hoc detection methods that can be applied to any pre-trained network. For many downstream tasks (not only OOD detection), the best results are achieved by models that have been pretrained on large datasets, some of which might not be publicly available. Since adjusting the training scheme for these networks is usually not feasible, simple post-hoc OOD detection is most often used in practice. 

Common post-hoc OOD detection methods 
are based on a scoring function that typically inputs either the logit/softmax outputs of a model \cite{hendrycks2017MSP,hendrycks22Scaling,liu2020energy}, or the pre-logit features \cite{LeeMahalanobis2018,RenRelMaha2021,sun2022knnood}, or both \cite{sun2021react,wang2022vim}.
VisionTransformers have shown particular success in this area \cite{OODFormerKoner21}. For large-scale settings where, e.g., ImageNet is the ID dataset, they perform particularly well \cite{anonymous2023COOD}, especially when paired with feature-based methods \cite{bitterwolf2023ninco}. Among those, the Mahalanobis distance \cite{LeeMahalanobis2018,RenRelMaha2021} stands out as a particularly effective and simple scoring function. However, despite leading for some models to state-of-the-art OOD performance, it fails for others and shows high performance variation across different models and pretraining schemes, and brittleness when confronted with supposedly easy noise distributions as OOD data \cite{bitterwolf2023ninco}. 

In this work, we observe that for models where the Mahalanobis distance does not work well as OOD detector, the assumptions underlying the method are often not well satisfied. In particular, the feature norms vary much more than expected when assuming a Gaussian model with a shared covariance matrix. To mitigate this problem, we provide a simple solution, called \ourMethod{}, which we visualize in Figure~\ref{fig:teaser-overview}: 
By projecting the features onto the unit sphere before estimating the Mahalanobis distance, we significantly reduce the class-dependent feature variability and obtain a better fit of the covariance matrix,
which ultimately leads to consistent improvements in OOD detection,
as demonstrated in Fig.~\ref{fig:teaser-barplot} or Tab.~\ref{tab:fpr-OpenOOD}.

In summary, our contributions are the following:
\begin{itemize}
    \item We observe that the assumptions underlying the Mahalanobis distance as OOD detection method, in particular that the features are normally distributed with a shared covariance matrix, are often not well satisfied
    \item We relate this to variations in the feature norm, which can vary strongly across and within classes, and correlates with the Mahalanobis distance
    \item We provide an easy solution,
    which we call \ourMethod: Normalizing the features by their $\ell_2$-norm before computing the Mahalanobis distance
    \item We evaluate \ourMethod{} across a large range of models with different pretraining schemes and architectures on ImageNet and Cifar datasets and find that it consistently outperforms the conventional Mahalanobis distance and other baseline methods, and improves the detection of far-OOD noise distributions
\end{itemize}

\section{Related Work}\label{sec:relwork}
\textbf{Mahalanobis distance.}
Most closely related to our work are the well-established OOD detection methods based on the Mahalanobis distance. \citet{LeeMahalanobis2018} proposed to estimate a class-conditional Gaussian distribution with a shared covariance matrix "\textit{with respect to (low- and upper-level) features}", and to use the minimal Mahalanobis distance to the respective mean vectors as OOD score. Since then, the community has transitioned to using only the pre-logit features. \citet{RenRelMaha2021} proposed to additionally estimate a class-agnostic mean and covariance matrix and use the difference between the two resulting scores as OOD score, called relative Mahalanobis distance. 
These methods have demonstrated broad applicability, spanning domains such as medical imaging \cite{Anthony23MahaMedical} and self-supervised OOD detection \cite{sehwag2021ssd}. 
Gaussian mixture models (GMMs) represent a more comprehensive framework for modelling feature distributions.
They have been applied to small-scale setups but require tweaks to the training process (e.g. spectral normalization) \cite{MukhotiCVPR23DDU}. Adapting them to ImageNet-scale setups as post-hoc OOD detectors has so far not been successful.

\textbf{Feature norm.} The role of the feature norm for OOD detection has been investigated in several works \cite{L2OODFace2020}. 
\citet{Park23UnderstandingFeatureNorm} underline that the norm of pre-logit features are equivalent to confidence scores and that the feature norms of OOD samples are typically smaller than those of ID samples. Their observations are mostly based on results obtained with strong over-training and simple networks. We will show that this observation does not hold generally.
\citet{Gia2023NormalizationContrastive} investigate the role of the $\ell_2$ norm in contrastive learning and OOD detection.
\citet{Regmi2024CVPRtf2norm} and \citet{haas2024l2norm} try to leverage the feature norm to discriminate between ID and OOD samples. In particular, they concurrently suggested training with $\ell_2$-normalized features and then using the norm of the unnormalized features as OOD score at inference time, similar to \citet{L2OODFace2020} and \citet{wei2022logitnorm}.

\textbf{Spherical embeddings.}
Spherical embeddings have been investigated and leveraged across several fields \cite{liu2018sphereface,AngVQ22,Sablayrolles2018SpreadingVF,yaras2022neural}, also within the OOD detection literature \cite{ADHypersphereZheng2022}. \citet{ming2023cider} proposed CIDER, a contrastive training scheme that creates well-separated hyperspherical embeddings via a dispersion loss and applies KNN as detection method at inference time. \citet{sehwag2021ssd} also train with a contrastive loss, and apply the Mahalanobis distance as OOD detection method on the normalized features at inference time. \citet{haas2023gmmnorm} observe that normalizing features during train and inference time improves performance on the DDU benchmark \cite{MukhotiCVPR23DDU}. They hypothesize that their training scheme induces early neural collapse, which might benefit out-of-distribution detection capabilities of networks. Importantly, all those methods are \textit{train-time} methods, i.e. require modifications to the training process, including feature normalization - either explicitly in the case of \citet{haas2023gmmnorm}, or implicitly through the contrastive loss in \citet{ming2023cider} and \citet{MukhotiCVPR23DDU}. They then apply normalization at inference time, because they also normalized at train time. 
In contrast, we highlight the benefits of feature normalization when applying the Mahalanobis distance as \textit{post-hoc} OOD detection method
in this work - which is non-obvious for generic pretraining schemes.

\textbf{Cosine-based detection scores.}
Many previous works have suggested using the angle, or more specifically, the cosine, for OOD detection, but those mostly require modifications to training or architecture \cite{techapanurak2020hyperparameter,tack2020csi}, or are used for unsupervised setups \cite{RadfordClip,ming2022delving}.
\citet{park2023nnguide} and \citet{sun2022knnood} use nearest neighbour search in the normalized feature-space, which amounts to a nearest neighbour search in the cosine space. We show that \ourMethod{} outperforms cosine-based OOD detection methods.

\section{Variations in feature norm degrade the performance of Mahalanobis-based OOD detectors}\label{sec:featurenorm}
In this Section, we investigate the assumptions underlying the Mahalanobis distance as OOD detection method. We report results for NINCO \cite{bitterwolf2023ninco} as OOD dataset. For all experiments, we use a pretrained ImageNet SwinV2-B-In21k model \cite{liu2022swintransformerv2scaling} with $87.1\%$ ImageNet accuracy. This strong model is a prototypical example where OOD detection on NINCO with Mahalanobis score performs significantly worse (FPR of 58.2\%) than for other similar models like the ViT-B16-In21k-augreg with 84.5\% accuracy \cite{steinerhowtotrainyourvit} but low FPR of 31.3\% using the Mahalanobis score. 

\subsection{Mahalanobis Distance}
The Mahalanobis distance is a simple, hyperparameter-free post-hoc OOD detector that has been suggested by \citet{LeeMahalanobis2018}.
Given the training set $(x_i,y_i)_{i=1}^n$ with input $x_i$ and class labels $y_i$ one estimates: i) the class-wise means $\hat{\mu}_c$ and ii) a shared covariance matrix $\hat{\Sigma}$:
\begin{align}
\hat{\mu}_{c} &= \frac{1}{N_{c}}\sum_{i:y_{i}=c}\phi(x_i) \\ \label{eq:mahavar}
\hat{{\Sigma}} &= \frac{1}{N}\sum_{c}^{C}\sum_{i:y_{i}=c}(\phi(x_i)-\hat{\mu}_{c})(\phi(x_i)-\hat{\mu}_{c})^{T}
\end{align}
where $\phi(x_i)$ are the pre-logit features of $x_i$,  $N_{c}$ the number of train samples in class $c$, $N$ the total number of train samples, and $C$ the total number of classes. The  Mahalanobis distance of a test sample $x_{\text{t}}$ to a class mean $\hat{\mu}_c$ is then
\begin{align}
    d_{Maha}(x_{\text{t}},\hat{\mu}_c)=(\phi(x_{\text{t}})-\hat{\mu}_{c})^{T}\hat{{\Sigma}}^{-1}(\phi(x_{\text{t}})-\hat{\mu}_{c})
\end{align}
and the final OOD-score $s_{\text{Maha}}(x_{\text{t}})$ of $x_{\text{t}}$ is the negative smallest distance to one of the class means:
\begin{equation}
s_{\text{Maha}}(x_{\text{t}})\,=\, %
-\min_c \,d_{Maha}(x_{\text{t}},\hat{\mu}_c)
\end{equation}
If $s_{\text{Maha}}(x_{\text{t}})\leq T$ then the sample is rejected as OOD, where for evaluation purposes $T$ is typically determined by fixing a TPR of 95\% on the in-distribution. The core assumption of \citet{lee2018training} is that ``the pre-trained features of the softmax neural classifier might
also follow the class-conditional Gaussian distribution''. Indeed, one implicitly uses a probabilistic model where each class is modelled as a Gaussian  $\mathcal{N}(\hat{\mu}_c,\hat{\Sigma})$ with a shared covariance matrix $\hat{\Sigma}$, which can be seen as a weighted average of the covariance matrices of the features of each class: $\hat{\Sigma}=\sum_{c=1}^C \frac{N_c}{N}\hat{\Sigma}_c$ with $\hat{\Sigma}_c=\frac{1}{N_c}\sum_{i:y_{i}=c}(\phi(x_i)-\hat{\mu}_{c})(\phi(x_i)-\hat{\mu}_{c})^{T}$, with the weight $N_c \backslash N$ being an estimate of $\Pr(Y=c)$.

The Mahanalobis score is a strong baseline for OOD detection as noted in \citet{bitterwolf2023ninco} where they report for a particular Vision Transformer (ViT)
trained with \textit{augreg} (a carefully selected combination of augmentation and regularization techniques) by \citet{steinerhowtotrainyourvit} state-of-the-art results on their NINCO benchmark comparing several models and OOD detection methods. On the other hand other ViTs like DeiT or Swin that are equally strong in terms of classification performance showed degraded OOD detection results. Moreover, \citet{bitterwolf2023ninco} report that the Mahalanobis-based OOD detector performs worse on their ``unit tests'' of simple far-OOD test sets than other methods.

In the remainder of this section, we will try to identify the reasons for the varying performance of Mahalanobis-based OOD detection. Our main hypothesis is that it is due to violations of its core assumptions:
\begin{itemize}
    \item \textbf{Assumption I:} the class-wise features $(\phi(x_i))_{y_i=c}$ follow a multivariate normal distribution $\mathcal{N}(\mu_c,\Sigma)$, 
    \item \textbf{Assumption II:} the covariance matrix $\hat{\Sigma}$ is the same for all classes. 
\end{itemize}
Below, we will show that these assumptions do not hold for some models, as indicated in Fig. \ref{fig:teaser-overview}. One strong indicator of this violation is the norm of the features, which turns out to be a strong confounder, ultimately degrading the OOD detection performance with Mahalanobis-based detectors. 

For completeness, we mention the Relative Mahalanobis score here, proposed by \citet{RenRelMaha2021}, also suggested as a fix to the Mahalanobis score. They argue that for the detection of near-OOD, one should use a likelihood ratio of two generative models compared to the likelihood used in the Mahalanobis method. Thus they fit a global Gaussian distribution with mean $\hat{\mu}_{\text{global}}$ and covariance matrix $\hat{\mathbf{\Sigma}}_{\text{global}}$, and use the difference between the class-conditional and the global Mahalanobis score as OOD score.

\begin{figure}[h]
    \centering
    \includegraphics[width=1.\linewidth]{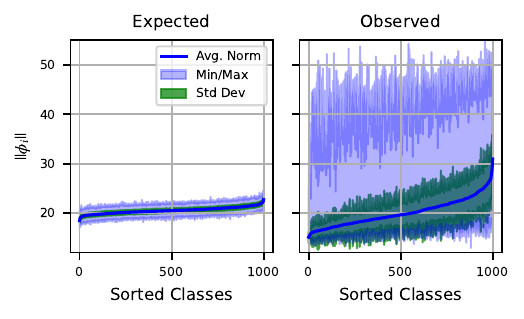}
    \caption{\textbf{The feature norms vary strongly across and within classes.} \textbf{Left:} We simulate how the feature norms per class would be distributed if they were sampled from Gaussians with the means and covariance matrix used for the Mahalanobis distance estimation. \textbf{Right:} The actual feature norm distribution observed in practice. Both the average norms across classes and the norms within each class vary much stronger than expected. 
    }
    \label{fig:fnorm-variations}
\end{figure}

\subsection{Is the Gaussian fit in feature space justified?}
As the features $\phi(x_i) \in \R^d$ for input $x$ are high-dimensional, e.g. $d=1024$ for a SwinV2-B, we expect some concentration of measure phenomena if the features $\phi(x)$ of a particular class are Gaussian distributed. In particular, the feature norm would be concentrated, as the following lemma shows.
\begin{lemma}\label{le:Gauss}
Let $\Phi(X) \sim \mathcal{N}(\mu,\Sigma)$. Then
\[ \Pr\left(|\norm{\Phi(X)}^2_2 -(\mathrm{tr}(\Sigma)+\norm{\mu}^2_2)| \geq \epsilon\right) \leq \frac{\mathrm{Var}\left(\norm{\Phi(X)}_2^2\right)}{\epsilon^2},\]
where $\mathrm{Var}(\norm{\Phi(X)}_2^2)\hspace{-0.5mm}:=\hspace{-0.5mm}\sum\limits_{i=1}^d (3\lambda_i^2+6 \mu_i^2 \lambda_i +\mu_i^4)- (\lambda_i+\mu_i^2)^2$ and $(\lambda_i)_{i=1}^d$ are the eigenvalues of $\Sigma$.
\end{lemma}

This implies that $\norm{\Phi(X)}_2$ should be concentrated around $\sqrt{\mathrm{tr}(\Sigma)+\norm{\mu}^2_2}$.
In the right part of Fig.~\ref{fig:fnorm-variations}, we show the distribution of the norms of the training features across classes for the SwinV2-B model, i.e. the feature norms of those samples that were used for estimating class means and covariance. In the left part of Fig. \ref{fig:fnorm-variations} we show the distribution of feature norms when sampling from $\mathcal{N}(\hat{\mu}_c,\hat{\Sigma})$ for every class $c$. As expected from the derived Lemma, the sampled norms vary little around their mean value. It is evident by the differences of the left and right part of Fig.~~\ref{fig:fnorm-variations}, that the fit with class-conditional means and shared covariance matrix does not represent the structure of the data well as the observed feature norms of SwinV2-B show heavy tails (right) which would not be present if the data was Gaussian (left). In Figure \ref{fig:feature-norm-dist-vits-four} we show that similar heavy-tailed feature norm distributions but with different skewness can be found even for the same ViT-architecture where the Mahalanobis score does not work well. 
This shows that Assumption I of the Mahalanobis score is not fulfilled across models, and models can deviate heavily from it.
In contrast, for the ViT-augreg \cite{steinerhowtotrainyourvit}, which has been shown to have very good OOD detection performance with the Mahalanobis score \cite{bitterwolf2023ninco}, the feature norms behave roughly as expected under the Gaussian assumption (right plot in Figure~\ref{fig:feature-norm-dist-vits-four}).

\begin{figure}
    \centering
    \includegraphics[width=1.\linewidth]{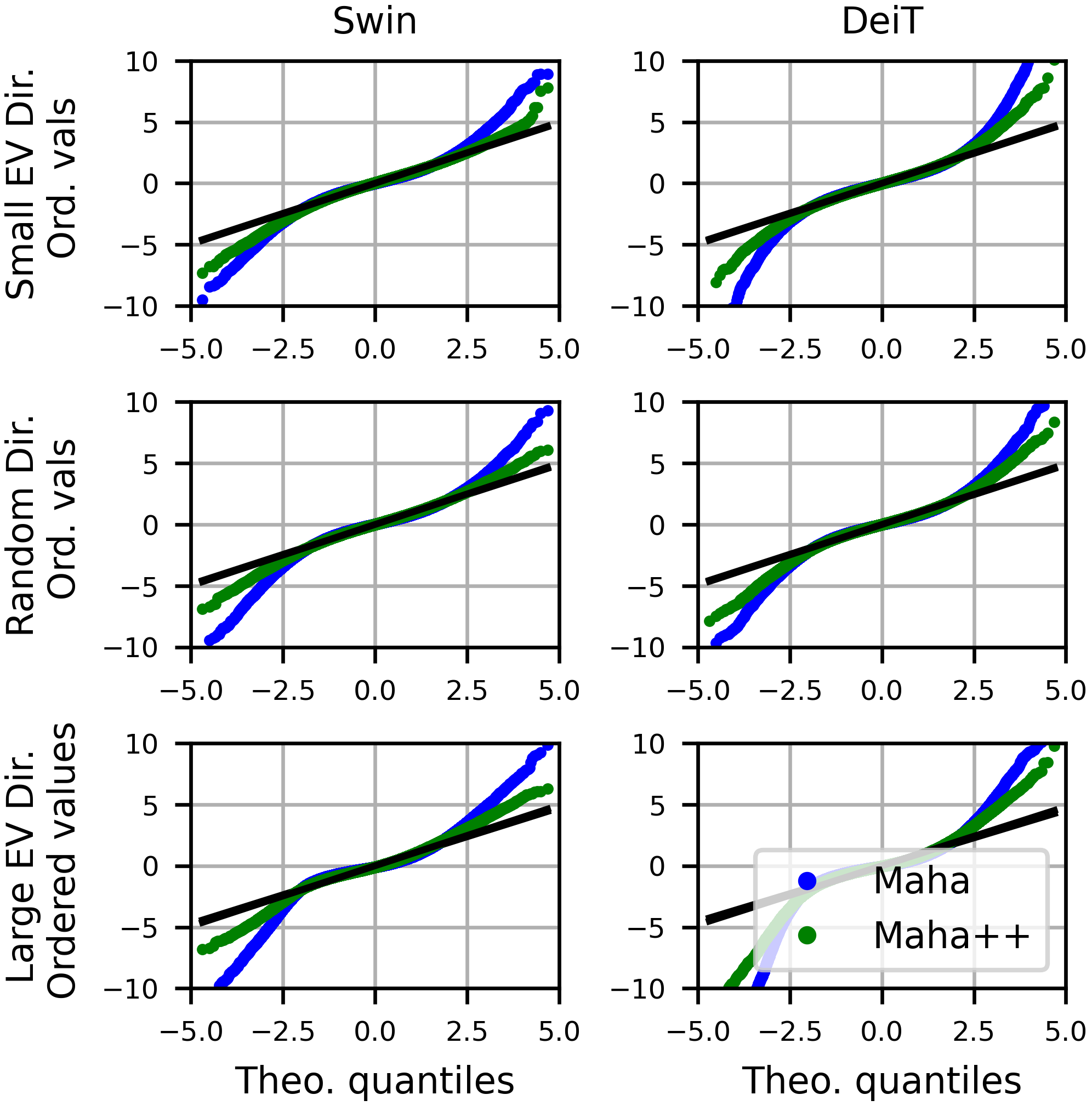}\vskip -0.in

    \caption{\textbf{QQ-plot: $\ell_2 -$normalization helps transform the features to be more aligned with a normal distribution.} For a SwinV2 and DeiT3 model (where the feature norms vary strongly across and within classes) normalization shifts the distribution towards a Gaussian (black line).}
    \label{fig:qq-plot-2models}\vskip -0.2in

\end{figure}

To further evaluate the adherence to Assumption I, we center training features of the SwinV2-B by their class means: $\phi^{\text{center}}(x_i) = \phi(x_i) - \mu_{c[i]}$.
These centered features, used for covariance estimation, should ideally follow a zero-mean multivariate normal distribution. To quantify deviations from normality, we use Quantile-Quantile (QQ) plots, a standard approach in statistics (see, e.g. \citet{qqplotWILK1968}) which compares sample quantiles against those of a theoretical distribution (here, the standard normal). A straight diagonal line indicates agreement with the theoretical distribution; deviations highlight mismatches.
To enable direct comparison between models (and later between normalized and unnormalized features), we standardize $\phi^{\text{center}}(x_i)$
by its empirical standard deviation. While standardization technically alters the distribution (as the empirical variance is sample-dependent), we expect this to be negligible due to the large dataset size ($>10^6$ samples). We report QQ-plots for three directions for a SwinV2-B and a DeiT3 (blue lines in Figure~\ref{fig:qq-plot-2models}), and observe strong deviations from the ideal diagonal line, indicating that the centered features have much stronger tails than expected if the features followed a Gaussian distribution, further refuting Assumption I. We observe similar heavy tails in QQ-plots of other models where the Mahalanobis score is not working well for OOD detection (see Fig.~\ref{fig:qq-many} in App.~\ref{app:featurenorm}). Only the ViT with augreg training has a QQ plot close to the expected one. 

To assess the validity of Assumption II, we measure how strongly the individual class variances deviate from the global variance. 
To this end, we compute the expected relative deviation over all directions:
\begin{align}\label{eq:expectdeviation}
\Exp_u[(u^TAu)^2] =\frac{2 \mathrm{tr}(A^2) + \mathrm{tr}(A)^2}{d(d+2)},
\end{align}

where $u$ has a uniform distribution on the unit sphere and $A=\hat{\Sigma}^{-\frac{1}{2}}(\hat{\Sigma}_i-\hat{\Sigma})\hat{\Sigma}^{-\frac{1}{2}}$ (see App.~\ref{app:proof-exp} for a derivation). We average over all classes $i$ and report the results for a SwinV2-B, a DeiT3-B and a ViT-augreg in Table~\ref{tab:gamma-mean}. We observe that the SwinV2 and DeiT3 show significantly larger deviations than the ViT-augreg, indicating that the class-specific variances differ more. More models in Tab.~\ref{tab:deviation-variances-avg} in the Appendix.

\begin{table}[]
    \centering
    \small
    \caption{\textbf{Variance alignment. } We measure how much the class-variances deviate from the global variance via the deviation score (see Eq.~\ref{eq:expectdeviation}). Lower values indicate better alignment. Normalization aligns the features of SwinV2 and DeiT3, but not ViT-augreg.}
    \label{tab:gamma-mean}
    \vspace{1mm}
\begin{tabular}{lcc}
\toprule
 &     unnormalized &      normalized \\
\midrule
          SwinV2-B-In21k &   0.26 &  0.12 \\       
          DeiT3-B16-In21k & 0.24 &      0.15 \\

   ViT-B16-In21k-augreg &   0.05 &  0.05 \\
\bottomrule
\end{tabular}
\end{table}

\subsection{Correlation of feature norm and $s_{\text{Maha}}$-score}
\begin{figure}
    \centering
    \includegraphics[width=1.\linewidth]{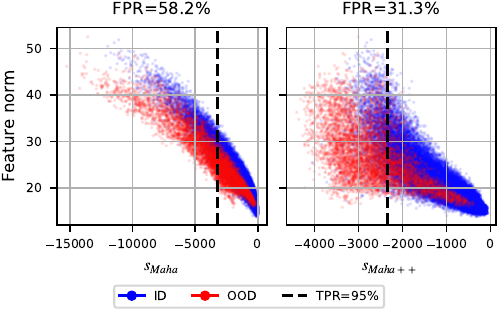}
    \vskip -0.1in
    \caption{\textbf{The feature norm correlates with the Mahalanobis score for SwinV2-B:} \textbf{Left:} The smaller the feature norm, the smaller the Mahalanobis OOD score $s_{Maha}$, irrespective of whether a sample is ID or not. OOD samples with small feature norms are systematically classified as ID. \textbf{Right:} After normalization, OOD samples with small feature norms can be detected, and OOD detection is significantly improved.
    }
    \label{fig:fnorm-maha}
\end{figure}

The strong variations within and across classes we observed in Figure~\ref{fig:fnorm-variations} indicate that the feature norm might impact the Mahalanobis estimation. To investigate this, we plot the feature norm against the Mahalanobis score $s_{Maha}$ assigned by the SwinV2-B model for ID and OOD test samples (i.e. samples that were not used for estimating means and covariance) in  Figure \ref{fig:fnorm-maha}. We observe a clear correlation: Samples with large feature norms consistently receive a large OOD score, and vice versa for samples with small feature norms - irrespective of whether they belong to the in or out distribution. Ideally, a detector should be able to distinguish ID from OOD samples irrespective of the norm of the OOD samples. Since a large fraction of the OOD samples have a comparably small feature norm, the resulting OOD detection performance is, however, poor. The reason for this strong correlation is the strongly different average feature norm across classes observed above. In Fig.~\ref{fig:fnorm-manymodels}, we observe the same correlation for other models (again, the ViT-B-augreg being an exception). In Fig.~\ref{fig:scaled-ood} in the Appendix, we substantiate this observation by artificially scaling the feature norm of OOD samples, leading to improved detection when the feature norm is increased and worse detection when the feature norm is decreased.

The heavy correlation between feature norm and OOD score implies that images yielding small feature norm are not detected as OOD (see Fig.~\ref{fig:scaled-ood} for a discussion). This also explains why the simple OOD unit tests in \citet{bitterwolf2023ninco}, using synthetic images of little variation, e.g. black or uni-colour images, often fail. These synthetic images contain little variation in color, which often results in small activations in the network, and thus small pre-logit features, see Figure~\ref{fig:fnorm-OOD-larger} for an analysis.

\section{Mahalanobis++: Normalize your features}\label{sec:maha++}
A challenge with Mahalanobis distance-based OOD detection is its sensitivity to feature norms, which can strongly correlate with Mahalanobis scores. We further find the feature distribution to strongly contradict the theoretical Gaussian assumption (with shared covariance), as empirical feature norms vary much more in practice than expected. To address this mismatch, we propose a simple but effective fix: Discarding the feature norm and leveraging only directional information in the features by $\ell_2$-normalization.

\textbf{Method.} 
Instead of the original features $\phi(x)$, we use $\ell_2$-normalized ones for computing the Mahalanobis score:
\begin{align}
    \hat{\phi}(x_i) = \phi(x_i)/\|\phi(x_i)\|_2,
\end{align}
The class-means and covariance matrix of the Mahalanobis score are estimated using the normalized features, and also test features are normalized when computing their score. We denote this simple modification as \ourMethod{}.

We note that $\ell_2$-normalization has been used with non-parametric post-hoc OOD detection methods like KNN \cite{sun2022knnood,park2023nnguide} or cosine similarity \cite{techapanurak2020hyperparameter}. 
With the Mahalanobis score, however,  $\ell_2$-normalization has - to the best of our knowledge - only been investigated for train-time methods like SSD+ \cite{sehwag2021ssd} or CIDER \cite{ming2023cider}. Those methods normalize their features for OOD detection because they also normalize during training. This is orthogonal to our work: The standard Mahalanobis method for OOD detection is a post-hoc method, where adjusting the pretraining scheme is not feasible. 
We show below that \ourMethod{} outperforms KNN and cosine similarity in all considered cases and, in particular, improves OOD detection consistently across tasks, architectures, training methods and OOD datasets as \textit{post-hoc} method.

\textbf{Improved normality. }
To evaluate how \ourMethod{} improves the adherence to the assumption of a Gaussian model with a shared covariance matrix, we compare the resulting feature distributions via QQ-plots to the unnormalized features. Like for the unnormalized features, we center $\hat{\phi}^{\text{center}}(x_i) = \hat{\phi}(x_i) - \hat{\mu}_{c[i]}$, divide by the empirical standard deviation, and plot the resulting quantiles against the quantiles of a standard normal. We observe that across all directions, normalization (green line in Figure~\ref{fig:qq-plot-2models}) shifts the feature quantiles closer to the diagonal line, 
confirming that \ourMethod{} better satisfies the Gaussian assumption of Mahalanobis-based detection.  We validate this for more models in Figure~\ref{fig:qq-many} in the Appendix.

\begin{table*}[ht]
\centering
    \caption{\textbf{ImageNet. } FPR (lower is better) on five OpenOOD datasets. \colorbox{softergreen}{Green} indicates that normalization improves over unnormalized features, \textbf{bold} indicates the best and \underline{underlined} the second best method. \ourMethod{} consistently improves over Maha and baselines.}
        \vskip 1.mm

    \resizebox{\textwidth}{!}{%

        \setlength{\tabcolsep}{2pt}

\begin{tabular}{l*{5}{c}|c
@{\hspace{10pt}}*{5}{c}
|c
@{\hspace{10pt}}*{5}{c}|c}
\toprule
model & \multicolumn{6}{c}{ConvNeXtV2-B-In21k} & \multicolumn{6}{c}{SwinV2-B-In21k} & \multicolumn{6}{c}{DeiT3-B16-In21k} \\
dataset &                                      NIN &                                   SSB &                                   TxT &                                  OpO &                                    iNat &                                   Avg &                                      NIN &                          SSB &                                   TxT &                                  OpO &                                 iNat &                                      Avg &                                      NIN &                                      SSB &                                      TxT &                                      OpO &                                    iNat &                                      Avg \\

MSP \cite{hendrycks2017MSP}                       &                                     41.4 &                                     60.1 &                                  47.4 &                                 24.6 &                                     8.7 &                                  36.5 &                                  48.2 &                         63.8 &                                  51.7 &                                 32.5 &                                 21.1 &                                  43.4 &                                     61.0 &                                     73.2 &                                     66.0 &                                     46.5 &                                    32.9 &                                     55.9 \\
MaxLogit \cite{hendrycks22Scaling}                 &                                     31.9 &                                     51.1 &                                  40.7 &                                 16.5 &                                     4.9 &                                  29.0 &                                  38.6 &                         52.6 &                                  47.7 &                                 24.6 &                                 13.0 &                                  35.3 &                                     55.2 &                                     67.2 &                                     61.9 &                                     41.4 &                                    34.3 &                                     52.0 \\
Energy     \cite{liu2020energy}               &                                     30.1 &                                     47.8 &                                  39.5 &                                 14.6 &                                     4.2 &                                  27.2 &                                  38.3 &                \textbf{47.8} &                                  50.9 &                                 26.3 &                                 13.9 &                                  35.5 &                                     55.9 &                                     65.2 &                                     63.2 &                                     43.3 &                                    45.6 &                                     54.7 \\
GEN \cite{Liu2023GEN}                       &                                     29.7 &                                     52.3 &                                  35.9 &                                 13.8 &                                     3.5 &                                  27.1 &                                  37.0 &                         57.0 &                                  38.7 &                                 17.6 &                                  8.9 &                                  31.8 &                                     45.2 &                            \textbf{61.5} &                                     50.1 &                                     24.8 &                                    13.7 &                                     39.0 \\
Energy+React  \cite{sun2021react}            &                                     29.5 &                                     48.0 &                                  38.6 &                                 13.9 &                                     3.7 &                                  26.7 &                                  35.1 &                         48.8 &                                  44.8 &                                 18.5 &                                  7.2 &                                  30.9 &                                     50.9 &                                     63.8 &                                     55.2 &                                     32.1 &                                    27.3 &                                     45.9 \\
fDBD \cite{liu2024fdbd}                      &                                     37.0 &                                     60.4 &                                  37.9 &                                 15.4 &                                     3.8 &                                  30.9 &                                  50.5 &                         74.5 &                                  41.9 &                                 19.1 &                                  6.1 &                                  38.4 &                                     53.5 &                                     70.9 &                                     50.7 &                                     24.8 &                                    11.8 &                                     42.4 \\
ViM  \cite{wang2022vim}                     &                                     26.9 &                                     47.7 &                      \underline{28.1} &                      \underline{7.9} &                            \textbf{1.1} &                      \underline{22.4} &                                  50.4 &                         75.7 &                                  35.4 &                                 15.4 &                      \underline{1.7} &                                  35.7 &                                     55.3 &                                     75.7 &                                     48.8 &                                     21.1 &                                     4.8 &                                     41.1 \\
KNN \cite{sun2022knnood}                       &                                     40.9 &                                     59.0 &                                  32.9 &                                 16.5 &                                     6.5 &                                  31.2 &                                  57.2 &                         82.3 &                                  35.4 &                                 18.5 &                                  6.8 &                                  40.0 &                                     52.6 &                                     73.7 &                                     43.7 &                                     21.1 &                                     9.7 &                                     40.2 \\
Neco \cite{ammar2023neco}                   &                                     27.7 &                            \textbf{45.9} &                                  35.0 &                                 12.5 &                                     2.8 &                                  24.8 &                      \underline{32.6} &             \underline{48.7} &                                  39.8 &                                 17.6 &                                  5.8 &                      \underline{28.9} &                                     51.5 &                                     64.8 &                                     57.4 &                                     34.1 &                                    24.2 &                                     46.4 \\
NNguide \cite{park2023nnguide}                 &                                     31.7 &                                     53.5 &                                  31.6 &                                 12.7 &                                     3.3 &                                  26.6 &                                  42.7 &                         72.5 &                      \underline{33.0} &                     \underline{12.3} &                                  3.5 &                                  32.8 &                                     46.4 &                                     68.4 &                                     44.3 &                                     19.3 &                                     9.3 &                                     37.5 \\
Rel.-Mahalanobis \cite{RenRelMaha2021}      &                                     28.1 &                                     54.6 &                                  33.2 &                                 11.7 &                                     2.0 &                                  25.9 &                                  48.2 &                         74.0 &                                  39.7 &                                 19.3 &                                  3.5 &                                  36.9 &                                     47.4 &                                     69.8 &                                     46.2 &                                     20.1 &                                     6.0 &                                     37.9 \\
Rel.-Mahalanobis++ &  \cellcolor{softergreen}\underline{24.7} &              \cellcolor{softergreen}51.6 &           \cellcolor{softergreen}32.1 &          \cellcolor{softergreen}11.3 &                                     2.2 &           \cellcolor{softergreen}24.4 &           \cellcolor{softergreen}34.4 &  \cellcolor{softergreen}62.5 &           \cellcolor{softergreen}36.8 &          \cellcolor{softergreen}15.7 &                                  3.9 &           \cellcolor{softergreen}30.6 &     \cellcolor{softergreen}\textbf{38.3} &  \cellcolor{softergreen}\underline{61.6} &  \cellcolor{softergreen}\underline{42.5} &  \cellcolor{softergreen}\underline{17.1} &  \cellcolor{softergreen}\underline{4.0} &  \cellcolor{softergreen}\underline{32.7} \\
Mahalanobis \cite{LeeMahalanobis2018}               &                                     30.3 &                                     53.8 &                                  30.4 &                                  9.4 &                                     1.4 &                                  25.0 &                                  58.2 &                         81.4 &                                  41.5 &                                 23.2 &                                  3.5 &                                  41.6 &                                     52.5 &                                     72.8 &                                     47.0 &                                     21.4 &                                     5.5 &                                     39.8 \\
Mahalanobis++          &     \cellcolor{softergreen}\textbf{22.4} &  \cellcolor{softergreen}\underline{46.9} &  \cellcolor{softergreen}\textbf{26.5} &  \cellcolor{softergreen}\textbf{7.8} &  \cellcolor{softergreen}\underline{1.3} &  \cellcolor{softergreen}\textbf{21.0} &  \cellcolor{softergreen}\textbf{31.3} &  \cellcolor{softergreen}62.0 &  \cellcolor{softergreen}\textbf{28.7} &  \cellcolor{softergreen}\textbf{9.7} &  \cellcolor{softergreen}\textbf{1.6} &  \cellcolor{softergreen}\textbf{26.7} &  \cellcolor{softergreen}\underline{38.8} &              \cellcolor{softergreen}62.8 &     \cellcolor{softergreen}\textbf{42.0} &     \cellcolor{softergreen}\textbf{15.6} &     \cellcolor{softergreen}\textbf{3.1} &     \cellcolor{softergreen}\textbf{32.5} \\
\bottomrule

\end{tabular}

}
\vspace{-3mm}

    \label{tab:per-dataset}
\end{table*}

\textbf{Variance alignment. } In Table~\ref{tab:gamma-mean}, we observe lower variance deviation scores for normalized features of the SwinV2 model compared to unnormalized features, indicating that normalization aligns the class variances in \ourMethod{}. We illustrate this effect in Figure~\ref{fig:teaser-overview}, which visualizes centered training features for three selected classes along a random direction. Without normalization, class feature variances differ substantially, and the shared covariance matrix fails to jointly capture their distributions. After normalization, class variances become more consistent, making the shared variance assumption more appropriate. 
To further validate this, 
we examine which in-distribution test samples are flagged as OOD at a 95\% true-positive rate: unnormalized Mahalanobis rejects samples from 634 classes, while \ourMethod{} rejects samples from 728 classes. In an ideal setting with a perfect covariance fit, one would expect samples to be drawn uniformly from all 1,000 classes. The increase from 634 to 728 classes suggests that normalization reduces bias in the covariance estimation, better aligning with the shared variance assumption. We substantiate our observations in Figure~\ref{fig:var-class-global} and Table~\ref{tab:deviation-variances-avg} in the Appendix for more models. We find that class variances are more similar to the global variances after normalization for all models - except the ViT-augreg.

\textbf{Decoupling of feature norm and OOD score.} In Figure~\ref{fig:fnorm-maha} on the right, we plot the feature norm of ID and OOD samples against their OOD scores
obtained via \ourMethod{}. In contrast to the conventional Mahalanobis score, the correlation between OOD score and feature norm (before normalization) is much weaker. In particular, OOD samples with small feature norm are now also detected as OOD, which was not the case for unnormalized Mahalanobis.

\vspace{-2mm}

\section{Experiments}

\textbf{ImageNet. }Our main goal is to investigate the effectiveness of \ourMethod{} across a large pool of architectures, model sizes and training schemes for ImageNet-scale OOD detection, as this is where the conventional Mahalanobis distance showed the most varied results in previous studies \cite{bitterwolf2023ninco,mueller2024trainvitooddetection}. To this end, we use 44 publicly available model checkpoints from timm \cite{rw2019timm} and \url{huggingface.co}.
Following the OpenOOD setup  \cite{yang2022openood}, we report results on Ninco \cite{bitterwolf2023ninco}, iNaturalist \cite{van2018inaturalist}, SSB-hard \cite{360OSR}, OpenImages-O \cite{OpenImages2} and Texture \cite{cimpoi2014describing}.
We report the false positive rate at a true positive rate of 95\% (FPR) as the OOD detection metric and refer to the appendix for other metrics, such as AUC, details on the model checkpoints, baseline methods, and extended results. In addition to \ourMethod{}, we also report \rourMethod{}, i.e the relative Mahalanobis distance with $\ell_2$ normalization.

We report results on the five OOD datasets in Table~\ref{tab:per-dataset} using three pretrained base-size models: ConvNextV2 \cite{Woo2023ConvNeXtV2}, 
SwinV2 \cite{liu2022swintransformerv2scaling} and DeiT3 \cite{Touvron2022DeiTIR}. For all models, \ourMethod{} outperforms the conventional Mahalanobis distance consistently across datasets, and is the best-performing method on average, and in most cases also per dataset. Also the \rourMethod{} outperforms its counterpart across models and datasets, but is slightly worse on average. 
In Table~\ref{tab:fpr-OpenOOD}, we show the results averaged over the five datasets for 44 models with different training schemes, model sizes and network types. With the exception of three models (two of which are trained with \textit{augreg}), \ourMethod{} outperforms its counterparts in \textit{all} cases. \rourMethod{} outperforms its counterpart in 39/44 cases. 
In 30/44 cases, the best performing method is \ourMethod{} (in 6/44 cases it is \rourMethod{})
and the differences to the baseline methods are often large. Averaged over models, \ourMethod{} is the best method, followed by \rourMethod{} and outperforming the previously best method ViM by 7 FPR points. 
We note that \ourMethod{} is particularly effective for the best-performing models, as it is the best method for 4 of the top-5 models.

\setlength{\tabcolsep}{3.8pt}
\begin{table}[htb]
\centering
\vspace{-3mm}
    \caption{\textbf{CIFAR100. } \colorbox{softergreen}{Green} indicates that normalization improves the baseline, \textbf{bold} and \underline{underlined} indicate the best and second best method. We report FPR averaged over OpenOOD datasets. Maha++ is the best method. The best FPR is achieved by Maha++ for ViT-S16-21k highlighted in {\color{blue}blue}.}
    \vskip 1mm
\resizebox{\columnwidth}{!}{%

    \begin{tabular}{lccccccc}

Model & MSP & Ash & ML & KNN & ViM & MD & MD++ \\
\hline
SwinV2-S-1k & 47.28 & 92.66 & 40.96 & 36.27 & \underline{34.02} & 40.10 & \cellcolor{softergreen}{\textbf{26.01}} \\
Deit3-S-21k & 48.92 & 94.47 & 42.37 & \underline{36.81} & 39.99 & 41.99 & \cellcolor{softergreen}{\textbf{31.72}} \\
ConvN-T-21k & 60.60 & 92.11 & 57.44 & \underline{51.16} & 51.18 & 52.48 & \cellcolor{softergreen}{\textbf{42.69}} \\
ViT-B32-21k & 48.02 & 93.98 & 31.28 & 26.49 & 27.14 & \underline{26.28} & \cellcolor{softergreen}{\textbf{18.94}} \\
ViT-S16-21k & 52.17 & 80.45 & 37.63 & 31.91 & \underline{24.90} & 25.51 & \cellcolor{softergreen}{\color{blue}\textbf{18.58}} \\
RN18 & 80.59 & 78.98 & 79.87 & \underline{76.61} & 79.61 & 79.48 & \cellcolor{softergreen}{\textbf{72.92}} \\
RN34 & 76.93 & 78.27 & 75.33 & \textbf{74.44} & 77.17 & 76.63 & \cellcolor{softergreen}{\underline{74.51}} \\
RNxt29-32 & 82.31 & \underline{72.59} & 82.30 & 73.17 & 76.40 & 77.67 & \cellcolor{softergreen}{\textbf{67.71}} \\
\hline
Average & 62.10 & 85.44 & 55.90 & \underline{50.86} & 51.30 & 52.52 & \cellcolor{softergreen}{\textbf{44.13}} \\
\end{tabular}

    }
    \label{tab:fpr-cifar100}
    \vspace{-1mm}
\end{table}

We further note that NNguide \cite{park2023nnguide} and KNN \cite{sun2022knnood}, both of which operate in a normalized feature space, are consistently outperformed by \ourMethod{}. The most competitive baseline method that is not based on the Mahalanobis distance is ViM \cite{wang2022vim}, which for certain models shows similar or slightly better performance than \ourMethod{} (e.g. for EVA and DeiT networks). For several other networks (e.g., ConvNexts, Mixer, ResNets, EfficientNets, Swins,...), differences are, however, larger and often in the range of 8-15\% FPR. We note that most of the OOD datasets in OpenOOD show contamination with ID samples, as reported in \citet{bitterwolf2023ninco}. Therefore, we report results on Ninco, which has been cleaned from ID data, separately in Table~\ref{tab:fpr-big-NINCO}, and find even clearer improvements of \ourMethod{}. 

\setlength{\tabcolsep}{3.5pt}
\begin{table*}[htb]
\vspace{-3mm}
\centering
\scriptsize 
    \caption{FPR on OpenOOD datasets, \colorbox{softergreen}{Green} indicates that a normalized method is better than its unnormalized counterpart, \textbf{bold} indicates the best and \underline{underlined} the second best method. Maha++ improves over Maha on average by 7.6\% in FPR over all models. Similarly, rMaha++ is, on average, 2.9\% better in FPR than rMaha. In total, Maha++ improves the SOTA compared to the strongest competitor rMaha among all OOD methods by 6.9\%, which is a significant improvement. The lowest FPR is achieved by Maha++ for the EVA02-L14-M38m-In21k highlighted in {\color{blue}blue}. }\vspace{1mm}
        \begin{tabular}{lccccccccccccccccc}
Model & Val Acc & MSP & E & E+R & ML & ViM & AshS & KNN & NNG & NEC & GMN & GEN & fDBD & Maha & Maha++ & rMaha & rMaha++ \\
\hline
ConvNeXt-B-In21k & 86.3 & 41.7 & 40.1 & 36.0 & 37.3 & \underline{29.5} & 88.5 & 37.2 & 31.8 & 31.4 & 54.2 & 32.6 & 37.9 & 33.6 & \cellcolor{softergreen}{\textbf{24.3}} & 31.7 & \cellcolor{softergreen}{29.5} \\
ConvNeXt-B & 84.4 & 61.4 & 90.9 & 86.9 & 70.2 & 52.8 & 99.5 & 58.7 & 51.2 & 66.5 & 73.9 & 60.1 & 60.3 & 54.2 & \cellcolor{softergreen}{\textbf{44.6}} & 50.0 & \cellcolor{softergreen}{\underline{45.4}} \\
ConvNeXtV2-T-In21k & 85.1 & 44.7 & 37.3 & 37.1 & 38.6 & \textbf{27.0} & 96.7 & 41.6 & 36.4 & 33.2 & 47.2 & 36.5 & 42.3 & 32.5 & \cellcolor{softergreen}{\underline{28.6}} & 34.6 & \cellcolor{softergreen}{33.4} \\
ConvNeXtV2-B-In21k & 87.6 & 36.5 & 27.2 & 26.7 & 29.0 & \underline{22.4} & 95.3 & 31.2 & 26.6 & 24.8 & 38.9 & 27.1 & 30.9 & 25.0 & \cellcolor{softergreen}{\textbf{21.0}} & 25.9 & \cellcolor{softergreen}{24.4} \\
ConvNeXtV2-L-In21k & 88.2 & 35.0 & 27.0 & 26.5 & 28.5 & 28.7 & 95.6 & 30.8 & 25.9 & 24.1 & 32.9 & 26.4 & 31.6 & 27.8 & \cellcolor{softergreen}{\textbf{18.8}} & 25.8 & \cellcolor{softergreen}{\underline{23.0}} \\
ConvNeXtV2-T & 83.5 & 60.5 & 66.1 & 58.6 & 58.9 & 49.9 & 99.2 & 72.1 & 62.8 & 54.1 & 73.9 & 53.6 & 61.8 & 55.4 & \cellcolor{softergreen}{\textbf{44.4}} & 48.9 & \cellcolor{softergreen}{\underline{44.6}} \\
ConvNeXtV2-B & 85.5 & 58.8 & 70.8 & 64.1 & 59.5 & 46.8 & 99.6 & 53.4 & 47.9 & 55.9 & 71.2 & 48.6 & 53.7 & 46.3 & \cellcolor{softergreen}{\textbf{37.5}} & 43.0 & \cellcolor{softergreen}{\underline{39.1}} \\
ConvNeXtV2-L & 86.1 & 58.6 & 68.0 & 60.1 & 58.3 & 48.4 & 99.1 & 48.9 & 44.7 & 55.9 & 63.8 & 46.3 & 48.5 & 41.7 & \cellcolor{softergreen}{\textbf{36.2}} & 39.0 & \cellcolor{softergreen}{\underline{38.0}} \\
DeiT3-S16-In21k & 84.8 & 60.5 & 53.3 & 50.4 & 54.4 & 47.6 & 99.2 & 49.8 & 47.5 & 51.6 & 52.0 & 47.8 & 54.4 & 50.2 & \cellcolor{softergreen}{\textbf{42.4}} & 48.9 & \cellcolor{softergreen}{\underline{43.6}} \\
DeiT3-B16-In21k & 86.7 & 55.9 & 54.7 & 45.9 & 52.0 & 41.1 & 99.2 & 40.2 & 37.5 & 46.4 & 46.2 & 39.0 & 42.4 & 39.8 & \cellcolor{softergreen}{\textbf{32.5}} & 37.9 & \cellcolor{softergreen}{\underline{32.7}} \\
DeiT3-L16-In21k & 87.7 & 55.0 & 45.5 & 38.3 & 46.9 & 36.4 & 98.1 & 35.0 & 32.1 & 38.5 & 36.8 & 34.6 & 38.2 & 34.9 & \cellcolor{softergreen}{\underline{30.1}} & 33.4 & \cellcolor{softergreen}{\textbf{29.9}} \\
DeiT3-S16 & 83.4 & 56.9 & 54.0 & 58.1 & 52.2 & \textbf{43.3} & 85.6 & 69.8 & 48.2 & 52.2 & 64.1 & 46.6 & 54.2 & 52.4 & \cellcolor{softergreen}{46.5} & 49.1 & \cellcolor{softergreen}{\underline{44.9}} \\
DeiT3-B16 & 85.1 & 59.7 & 82.3 & 88.4 & 64.4 & \textbf{44.7} & 99.2 & 66.1 & 71.3 & 63.5 & 63.5 & 46.0 & 54.8 & 51.5 & \cellcolor{softergreen}{46.7} & 48.3 & \cellcolor{softergreen}{\underline{45.0}} \\
DeiT3-L16 & 85.8 & 60.3 & 80.5 & 89.3 & 64.0 & 46.1 & 78.4 & 54.0 & 72.5 & 64.3 & 56.9 & 45.1 & 52.4 & 45.3 & \cellcolor{softergreen}{\underline{39.7}} & 42.7 & \cellcolor{softergreen}{\textbf{38.6}} \\
EVA02-B14-In21k & 88.7 & 32.4 & 26.8 & 26.2 & 28.8 & \underline{22.0} & 87.9 & 29.6 & 25.8 & 25.0 & 36.0 & 24.3 & 28.6 & 25.5 & \cellcolor{softergreen}{\textbf{21.0}} & 26.2 & \cellcolor{softergreen}{23.8} \\
EVA02-L14-M38m-In21k & 90.1 & 27.0 & 22.6 & 22.4 & 24.3 & \underline{18.0} & 91.0 & 25.8 & 22.8 & 21.8 & 39.9 & 20.3 & 23.9 & 19.7 & \cellcolor{softergreen}{\textbf{\color{blue}17.7}} & 21.1 & \cellcolor{softergreen}{20.4} \\
EVA02-T14 & 80.6 & 64.8 & 66.2 & 66.8 & 63.1 & \underline{49.3} & 98.4 & 60.8 & 57.1 & 55.4 & 54.1 & 57.9 & 66.4 & 51.0 & \cellcolor{softergreen}{\textbf{48.1}} & 52.6 & \cellcolor{softergreen}{50.7} \\
EVA02-S14 & 85.7 & 52.2 & 53.4 & 53.1 & 49.5 & \textbf{34.8} & 99.1 & 44.1 & 40.3 & 42.9 & 43.0 & 41.7 & 48.9 & 36.6 & \cellcolor{softergreen}{\underline{35.4}} & 38.1 & \cellcolor{softergreen}{36.8} \\
EffNetV2-S & 83.9 & 59.3 & 71.0 & 58.7 & 61.1 & 52.2 & 99.4 & 45.6 & 45.2 & 59.3 & 77.9 & 49.7 & 54.0 & 47.3 & \cellcolor{softergreen}{\textbf{40.2}} & 43.6 & \cellcolor{softergreen}{\underline{40.4}} \\
EffNetV2-L & 85.7 & 57.1 & 74.2 & 57.4 & 58.8 & 48.9 & 99.2 & 48.9 & 47.1 & 56.0 & 58.1 & 44.7 & 49.2 & 41.3 & \cellcolor{softergreen}{\textbf{34.6}} & 38.0 & \cellcolor{softergreen}{\underline{34.8}} \\
EffNetV2-M & 85.2 & 57.0 & 69.3 & 56.7 & 57.3 & 54.7 & 99.5 & 51.3 & 48.6 & 54.9 & 66.8 & 45.2 & 52.6 & 46.0 & \cellcolor{softergreen}{\underline{37.1}} & 41.1 & \cellcolor{softergreen}{\textbf{36.8}} \\
Mixer-B16-In21k & 76.6 & 71.5 & 83.0 & 83.5 & 75.0 & 71.8 & 95.8 & 77.8 & 83.9 & 75.5 & 61.2 & 67.7 & 71.8 & 63.3 & \cellcolor{softergreen}{\textbf{52.5}} & 60.0 & \cellcolor{softergreen}{\underline{52.9}} \\
SwinV2-B-In21k & 87.1 & 43.4 & 35.5 & 30.9 & 35.3 & 35.7 & 77.0 & 40.0 & 32.8 & \underline{28.9} & 57.7 & 31.8 & 38.4 & 41.6 & \cellcolor{softergreen}{\textbf{26.7}} & 36.9 & \cellcolor{softergreen}{30.6} \\
SwinV2-L-In21k & 87.5 & 40.4 & 35.9 & 31.2 & 34.5 & 39.0 & 85.1 & 38.9 & 32.9 & 29.0 & 48.8 & 31.5 & 37.5 & 41.8 & \cellcolor{softergreen}{\textbf{24.7}} & 36.2 & \cellcolor{softergreen}{\underline{28.7}} \\
SwinV2-S & 84.2 & 61.2 & 68.1 & 62.1 & 60.9 & 51.1 & 99.9 & 58.6 & 52.9 & 56.0 & 61.2 & 52.8 & 60.7 & 52.4 & \cellcolor{softergreen}{\textbf{38.9}} & 48.7 & \cellcolor{softergreen}{\underline{39.3}} \\
SwinV2-B & 84.6 & 62.4 & 66.2 & 58.2 & 60.5 & 49.9 & 99.1 & 55.0 & 51.1 & 55.4 & 56.1 & 49.9 & 56.9 & 47.9 & \cellcolor{softergreen}{\underline{40.1}} & 45.2 & \cellcolor{softergreen}{\textbf{39.7}} \\
ResNet101 & 81.9 & 67.7 & 82.8 & 99.6 & 70.7 & 50.5 & 80.2 & 53.6 & 51.4 & 70.6 & 82.3 & 62.5 & 71.3 & \underline{45.9} & \cellcolor{softergreen}{\textbf{43.5}} & 55.6 & 66.8 \\
ResNet152 & 82.3 & 66.4 & 82.1 & 99.5 & 70.0 & 49.7 & 80.0 & 52.0 & 46.8 & 69.1 & 77.2 & 60.3 & 69.3 & \underline{44.4} & \cellcolor{softergreen}{\textbf{38.3}} & 51.8 & 64.7 \\
ResNet50 & 80.9 & 72.0 & 95.9 & 99.4 & 75.8 & 53.1 & 80.3 & 67.8 & 64.1 & 76.6 & 89.5 & 65.4 & 74.8 & \textbf{49.5} & \underline{52.0} & 62.5 & 70.4 \\
ResNet50-supcon & 78.7 & 54.0 & 47.3 & 42.1 & 48.4 & 72.0 & \textbf{40.6} & 47.0 & \underline{41.9} & 47.8 & 78.8 & 53.5 & 48.0 & 95.5 & \cellcolor{softergreen}{44.5} & 90.2 & \cellcolor{softergreen}{63.7} \\
ViT-T16-In21k-augreg & 75.5 & 70.7 & 55.3 & \underline{48.4} & 58.3 & 51.1 & 94.9 & 76.2 & 71.0 & 52.8 & 58.2 & 64.7 & 58.5 & 55.5 & \cellcolor{softergreen}{\textbf{48.0}} & 59.2 & \cellcolor{softergreen}{57.7} \\
ViT-S16-In21k-augreg & 81.4 & 57.0 & 38.9 & 42.5 & 41.7 & \underline{33.4} & 76.7 & 55.6 & 48.9 & 38.1 & 44.3 & 46.5 & 44.0 & 36.7 & \cellcolor{softergreen}{\textbf{31.7}} & 43.0 & \cellcolor{softergreen}{40.6} \\
ViT-B16-In21k-augreg2 & 85.1 & 55.3 & 45.9 & 41.1 & 47.5 & 53.9 & 98.6 & 47.5 & 42.2 & 43.7 & 60.1 & 42.9 & 51.4 & 54.2 & \cellcolor{softergreen}{\textbf{38.2}} & 47.0 & \cellcolor{softergreen}{\underline{39.1}} \\
ViT-B16-In21k-augreg & 84.5 & 46.5 & 33.7 & 36.0 & 34.6 & \underline{26.9} & 94.9 & 54.3 & 45.6 & 32.4 & 38.6 & 36.5 & 36.4 & \textbf{25.7} & 28.3 & 30.8 & 31.5 \\
ViT-B16-In21k-orig & 81.8 & 44.6 & 30.7 & 30.9 & 33.1 & \underline{29.0} & 62.6 & 38.6 & 35.4 & 30.5 & 48.8 & 38.4 & 35.7 & 30.9 & \cellcolor{softergreen}{\textbf{27.5}} & 35.4 & \cellcolor{softergreen}{33.9} \\
ViT-B16-In21k-miil & 84.3 & 48.0 & 35.0 & 34.6 & 38.8 & 37.8 & 96.9 & 45.0 & 38.5 & \underline{33.9} & 57.1 & 38.3 & 44.6 & 47.1 & \cellcolor{softergreen}{\textbf{30.4}} & 43.6 & \cellcolor{softergreen}{36.7} \\
ViT-L16-In21k-augreg & 85.8 & 40.2 & 29.4 & 25.0 & 30.0 & \underline{23.6} & 94.5 & 50.6 & 41.2 & 28.0 & 41.6 & 30.7 & 30.4 & \textbf{21.0} & 23.9 & 25.2 & 25.8 \\
ViT-L16-In21k-orig & 81.5 & 40.8 & 29.3 & \underline{29.2} & 31.1 & 30.4 & 49.3 & 34.0 & 31.6 & 29.4 & 47.9 & 35.2 & 33.0 & 30.9 & \cellcolor{softergreen}{\textbf{26.8}} & 33.6 & \cellcolor{softergreen}{32.6} \\
ViT-S16-augreg & 78.8 & 64.8 & 59.0 & 60.6 & 60.0 & 68.1 & 96.9 & 71.5 & 68.9 & 60.2 & 61.8 & 61.8 & 64.8 & 49.3 & \cellcolor{softergreen}{49.2} & \underline{48.4} & \cellcolor{softergreen}{\textbf{48.2}} \\
ViT-B16-augreg & 79.2 & 64.3 & 59.6 & 56.2 & 60.1 & 63.4 & 90.2 & 65.5 & 64.1 & 59.9 & 60.3 & 61.5 & 63.5 & 49.6 & \cellcolor{softergreen}{48.0} & \underline{47.6} & \cellcolor{softergreen}{\textbf{46.7}} \\
ViT-B16-CLIP-L2b-In12k & 86.2 & 42.2 & 37.7 & 35.5 & 37.2 & 35.5 & 99.5 & 35.6 & \underline{31.6} & 34.0 & 41.4 & 33.4 & 38.0 & 43.2 & \cellcolor{softergreen}{\textbf{28.1}} & 38.0 & \cellcolor{softergreen}{32.4} \\
ViT-L14-CLIP-L2b-In12k & 88.2 & 31.5 & 25.2 & 24.6 & 26.5 & \textbf{21.5} & 97.6 & 29.9 & \underline{22.3} & 26.3 & 36.3 & 24.3 & 27.3 & 28.2 & \cellcolor{softergreen}{22.4} & 27.1 & \cellcolor{softergreen}{25.4} \\
ViT-H14-CLIP-L2b-In12k & 88.6 & 32.0 & 26.5 & 26.1 & 27.7 & \underline{22.3} & 97.8 & 31.2 & 23.4 & 27.5 & 53.0 & 24.6 & 28.9 & 27.1 & \cellcolor{softergreen}{\textbf{22.0}} & 26.8 & \cellcolor{softergreen}{25.1} \\
ViT-so400M-SigLip & 89.4 & 45.5 & 47.1 & 39.4 & 41.8 & 30.6 & 93.5 & 28.7 & 26.1 & 39.6 & 64.3 & 28.3 & 29.9 & 28.8 & \cellcolor{softergreen}{\textbf{24.5}} & 27.3 & \cellcolor{softergreen}{\underline{25.5}} \\
\hline
\textbf{Average} & 84.4 & 52.7 & 53.0 & 51.0 & 49.0 & 41.9 & 90.7 & 48.9 & 44.8 & 46.0 & 56.3 & 43.6 & 47.8 & 42.5 & \cellcolor{softergreen}{\textbf{34.9}} & 41.8 & \cellcolor{softergreen}{\underline{38.9}} \\
\end{tabular}

    \label{tab:fpr-OpenOOD}
    \vspace{-2mm}
\end{table*}

Two of the three models for which \ourMethod{} does not bring an improvement are ViTs trained with \textit{augreg} by \citet{steinerhowtotrainyourvit}. Those are the models that showed state-of-the-art performance in \citet{bitterwolf2023ninco}. 
We extend our observations from the previous Sections regarding these models in 
Appendix \ref{app:featurenorm}, where we show that 
their feature norms are already well-behaved; therefore, $\ell_2$ normalization does not improve the normality assumptions.

\citet{bitterwolf2023ninco} reported that Mahalanobis-based detectors sometimes fail to detect supposedly easy-to-detect noise distributions (called \textit{"unit tests"}). In Section~\ref{sec:featurenorm}, we connected this to the small feature norm those samples obtain. In Table~\ref{tab:unit-tests} we report the number of "failed" unit tests (a unit test counts as failed when a detector shows FPR values above $10\%$) and observe that normalization, in particular \ourMethod{} remedies this effectively. For results on all models, we refer to Table~\ref{tab:unit-tests-big} Appendix. 

\setlength{\tabcolsep}{8pt}
\begin{table}[htb]
\small 
\vspace{-2mm}
    \centering
    \caption{\textbf{Normalization improves robustness against noise distributions.}
We report the number of failed unit tests (noise distributions with FPR values $\geq10\%$) from \citet{bitterwolf2023ninco}. Normalization remedies the brittleness of Mahalanobis-based detectors. Full Table in Appendix~\ref{app:extended-results}.}
    \vskip 1.mm
    \label{tab:unit-tests}
    
\begin{tabular}{lccc}
\toprule
model            &  ConvNeXtV2 &  SwinV2 &  ViT-CLIP \\
\midrule
Maha      &                 5/17 &            10/17 &                    14/17 \\
Maha++ &                 \textbf{0/17} &             \textbf{0/17} &                     \textbf{0/17} \\
\bottomrule
\end{tabular}

\end{table}

\setlength{\tabcolsep}{3.5pt}
\begin{table*}[htb]
\centering
\scriptsize 
    \caption{FPR on NINCO, \colorbox{softergreen}{Green} indicates that normalized method is better than its unnormalized counterpart, \textbf{bold} indicates the best method, and \underline{underlined} indicates second best method. Maha++ improves over Maha on average by 10.9\% in FPR over all models. Similarly, rMaha++ is 6.0\% better in FPR than rMaha. In total, Maha++ improves the SOTA compared to the strongest competitor rMaha among all OOD models by 6.3\% which is significant. The lowest FPR is achieved by Maha++ for the ConvNeXtV2-L-In21k highlighted in {\color{blue}blue}. }\vspace{1mm}
        \begin{tabular}{lccccccccccccccccc}
Model & Val Acc & MSP & E & E+R & ML & ViM & AshS & KNN & NNG & NEC & GMN & GEN & fDBD & Maha & Maha++ & rMaha & rMaha++ \\
\hline
ConvNeXt-B-In21k & 86.3 & 46.2 & 43.0 & 40.0 & 40.5 & 41.2 & 92.7 & 51.6 & 41.2 & 35.2 & 53.8 & 38.0 & 48.3 & 48.7 & \cellcolor{softergreen}{\textbf{28.8}} & 40.2 & \cellcolor{softergreen}{\underline{32.5}} \\
ConvNeXt-B & 84.4 & 64.1 & 89.4 & 86.2 & 71.4 & 64.7 & 99.6 & 70.1 & 62.2 & 68.2 & 68.3 & 65.9 & 68.6 & 64.6 & \cellcolor{softergreen}{\underline{50.5}} & 59.5 & \cellcolor{softergreen}{\textbf{49.7}} \\
ConvNeXtV2-T-In21k & 85.1 & 51.6 & 42.4 & 42.4 & 44.1 & \underline{34.5} & 97.5 & 54.1 & 45.0 & 38.7 & 52.5 & 44.2 & 51.4 & 40.9 & \cellcolor{softergreen}{\textbf{32.8}} & 40.0 & \cellcolor{softergreen}{36.8} \\
ConvNeXtV2-B-In21k & 87.6 & 41.4 & 30.1 & 29.5 & 31.9 & 26.9 & 95.8 & 40.9 & 31.7 & 27.7 & 40.6 & 29.7 & 37.0 & 30.3 & \cellcolor{softergreen}{\textbf{22.4}} & 28.1 & \cellcolor{softergreen}{\underline{24.7}} \\
ConvNeXtV2-L-In21k & 88.2 & 38.7 & 30.7 & 29.8 & 31.5 & 37.2 & 96.0 & 38.7 & 29.9 & 27.0 & 43.4 & 29.0 & 36.6 & 34.6 & \cellcolor{softergreen}{\textbf{\color{blue}18.4}} & 27.7 & \cellcolor{softergreen}{\underline{21.4}} \\
ConvNeXtV2-T & 83.5 & 66.1 & 73.3 & 68.4 & 65.7 & 64.4 & 99.2 & 82.3 & 73.9 & 61.7 & 72.3 & 64.1 & 71.6 & 66.1 & \cellcolor{softergreen}{\underline{52.3}} & 58.6 & \cellcolor{softergreen}{\textbf{49.7}} \\
ConvNeXtV2-B & 85.5 & 62.9 & 73.9 & 69.8 & 63.5 & 61.1 & 99.4 & 67.3 & 60.3 & 60.7 & 69.2 & 57.1 & 65.1 & 58.9 & \cellcolor{softergreen}{\underline{44.7}} & 52.4 & \cellcolor{softergreen}{\textbf{44.1}} \\
ConvNeXtV2-L & 86.1 & 63.8 & 72.3 & 66.8 & 63.6 & 62.4 & 99.5 & 62.4 & 56.9 & 61.9 & 71.2 & 55.6 & 59.7 & 53.8 & \cellcolor{softergreen}{\underline{43.0}} & 47.1 & \cellcolor{softergreen}{\textbf{42.1}} \\
DeiT3-S16-In21k & 84.8 & 68.7 & 61.8 & 59.8 & 62.6 & 60.6 & 99.7 & 62.9 & 59.3 & 60.1 & 58.5 & 58.7 & 65.8 & 62.4 & \cellcolor{softergreen}{\textbf{50.8}} & 59.8 & \cellcolor{softergreen}{\underline{50.9}} \\
DeiT3-B16-In21k & 86.7 & 61.0 & 55.9 & 50.9 & 55.2 & 55.3 & 99.5 & 52.6 & 46.4 & 51.5 & 52.2 & 45.2 & 53.5 & 52.5 & \cellcolor{softergreen}{\underline{38.8}} & 47.4 & \cellcolor{softergreen}{\textbf{38.3}} \\
DeiT3-L16-In21k & 87.7 & 59.7 & 46.2 & 41.8 & 48.9 & 45.7 & 98.4 & 43.8 & 37.8 & 42.3 & 43.7 & 38.2 & 46.6 & 42.0 & \cellcolor{softergreen}{\underline{33.9}} & 38.1 & \cellcolor{softergreen}{\textbf{32.8}} \\
DeiT3-S16 & 83.4 & 64.3 & 63.0 & 63.8 & 60.6 & 54.3 & 84.1 & 75.6 & 57.8 & 60.6 & 66.4 & 57.4 & 65.0 & 61.1 & \cellcolor{softergreen}{\underline{53.5}} & 56.3 & \cellcolor{softergreen}{\textbf{50.5}} \\
DeiT3-B16 & 85.1 & 66.7 & 87.8 & 89.9 & 72.6 & 59.7 & 99.1 & 74.5 & 80.1 & 71.9 & 66.7 & 57.3 & 67.1 & 63.7 & \cellcolor{softergreen}{\underline{57.2}} & 58.9 & \cellcolor{softergreen}{\textbf{53.2}} \\
DeiT3-L16 & 85.8 & 67.8 & 82.3 & 86.6 & 70.5 & 57.9 & 81.1 & 67.2 & 77.9 & 70.8 & 62.9 & 58.4 & 64.4 & 57.0 & \cellcolor{softergreen}{\underline{50.4}} & 52.0 & \cellcolor{softergreen}{\textbf{46.6}} \\
EVA02-B14-In21k & 88.7 & 35.8 & 28.2 & 27.4 & 30.9 & 28.7 & 92.7 & 37.6 & 30.0 & 27.3 & 39.0 & \underline{25.8} & 32.3 & 31.7 & \cellcolor{softergreen}{\textbf{23.8}} & 30.3 & \cellcolor{softergreen}{25.9} \\
EVA02-L14-M38m-In21k & 90.1 & 29.0 & 24.3 & 24.1 & 25.7 & 21.4 & 94.8 & 30.3 & 26.1 & 22.9 & 39.5 & 20.3 & 26.0 & 22.2 & \cellcolor{softergreen}{\textbf{18.6}} & 22.1 & \cellcolor{softergreen}{\underline{20.1}} \\
EVA02-T14 & 80.6 & 72.7 & 74.5 & 75.2 & 72.1 & 67.7 & 98.8 & 74.5 & 71.0 & 68.4 & 65.6 & 70.5 & 73.5 & 65.9 & \cellcolor{softergreen}{\textbf{64.0}} & 65.9 & \cellcolor{softergreen}{\underline{64.4}} \\
EVA02-S14 & 85.7 & 61.2 & 61.4 & 61.5 & 57.8 & 51.6 & 98.9 & 60.0 & 54.0 & 53.2 & 51.9 & 53.1 & 60.0 & 49.3 & \cellcolor{softergreen}{\underline{48.0}} & 49.1 & \cellcolor{softergreen}{\textbf{47.8}} \\
EffNetV2-S & 83.9 & 67.7 & 77.5 & 73.3 & 69.5 & 74.0 & 99.7 & 60.9 & 59.6 & 69.4 & 79.9 & 62.9 & 67.9 & 67.5 & \cellcolor{softergreen}{59.9} & \underline{59.2} & \cellcolor{softergreen}{\textbf{52.1}} \\
EffNetV2-L & 85.7 & 63.7 & 77.2 & 68.8 & 64.3 & 69.4 & 98.9 & 62.5 & 60.1 & 63.5 & 64.4 & 56.3 & 62.4 & 58.4 & \cellcolor{softergreen}{\underline{47.8}} & 50.8 & \cellcolor{softergreen}{\textbf{44.3}} \\
EffNetV2-M & 85.2 & 63.4 & 75.1 & 69.1 & 63.8 & 72.3 & 99.6 & 63.1 & 60.6 & 63.3 & 67.5 & 56.3 & 64.5 & 61.7 & \cellcolor{softergreen}{\underline{50.0}} & 52.2 & \cellcolor{softergreen}{\textbf{45.3}} \\
Mixer-B16-In21k & 76.6 & 77.4 & 83.4 & 83.5 & 79.5 & 78.0 & 94.8 & 85.8 & 83.7 & 79.8 & 66.7 & 75.9 & 80.3 & 73.4 & \cellcolor{softergreen}{\underline{65.4}} & 70.3 & \cellcolor{softergreen}{\textbf{63.1}} \\
SwinV2-B-In21k & 87.1 & 48.2 & 38.3 & 35.1 & 38.6 & 50.4 & 86.0 & 57.2 & 42.7 & \underline{32.6} & 56.3 & 37.0 & 50.5 & 58.2 & \cellcolor{softergreen}{\textbf{31.3}} & 48.2 & \cellcolor{softergreen}{34.4} \\
SwinV2-L-In21k & 87.5 & 45.9 & 38.7 & 35.4 & 38.6 & 55.3 & 89.9 & 55.1 & 41.7 & 32.3 & 63.1 & 36.5 & 50.5 & 57.8 & \cellcolor{softergreen}{\textbf{28.3}} & 47.6 & \cellcolor{softergreen}{\underline{32.2}} \\
SwinV2-S & 84.2 & 67.6 & 71.7 & 70.1 & 66.7 & 66.8 & 99.8 & 73.1 & 66.8 & 62.7 & 61.0 & 63.8 & 73.4 & 68.0 & \cellcolor{softergreen}{\underline{49.8}} & 63.7 & \cellcolor{softergreen}{\textbf{48.5}} \\
SwinV2-B & 84.6 & 69.5 & 72.6 & 69.3 & 67.4 & 66.6 & 97.8 & 69.4 & 65.2 & 64.2 & 60.7 & 62.0 & 70.5 & 63.3 & \cellcolor{softergreen}{\underline{52.2}} & 59.1 & \cellcolor{softergreen}{\textbf{50.2}} \\
ResNet101 & 81.9 & 73.4 & 85.2 & 100.0 & 76.1 & 75.8 & 89.9 & 74.9 & 66.4 & 77.2 & 83.5 & 72.5 & 84.5 & 66.8 & \cellcolor{softergreen}{\textbf{50.4}} & 55.8 & \cellcolor{softergreen}{\underline{53.5}} \\
ResNet152 & 82.3 & 71.2 & 83.2 & 100.0 & 74.4 & 74.6 & 88.1 & 72.0 & 61.6 & 75.0 & 79.5 & 69.9 & 82.4 & 64.9 & \cellcolor{softergreen}{\textbf{46.5}} & 52.7 & \cellcolor{softergreen}{\underline{52.2}} \\
ResNet50 & 80.9 & 76.0 & 94.7 & 99.9 & 78.6 & 79.6 & 89.9 & 83.7 & 75.0 & 80.0 & 89.1 & 75.0 & 85.7 & 69.9 & \cellcolor{softergreen}{61.0} & \underline{58.2} & \cellcolor{softergreen}{\textbf{56.9}} \\
ResNet50-supcon & 78.7 & 60.6 & 57.0 & \textbf{56.1} & \underline{56.8} & 84.6 & 59.1 & 65.8 & 58.4 & 56.9 & 80.3 & 60.0 & 63.9 & 98.3 & \cellcolor{softergreen}{59.6} & 90.7 & \cellcolor{softergreen}{61.8} \\
ViT-T16-In21k-augreg & 75.5 & 79.0 & 72.9 & 69.6 & 74.0 & 66.4 & 90.1 & 81.7 & 81.9 & 71.1 & 69.4 & 78.4 & 72.3 & \textbf{61.6} & \underline{63.2} & 67.6 & 68.0 \\
ViT-S16-In21k-augreg & 81.4 & 67.0 & 53.3 & 55.4 & 54.7 & 47.4 & 85.0 & 70.9 & 64.0 & 51.9 & 58.3 & 61.4 & 57.6 & \underline{44.8} & \cellcolor{softergreen}{\textbf{44.6}} & 51.1 & \cellcolor{softergreen}{50.5} \\
ViT-B16-In21k-augreg2 & 85.1 & 62.1 & 52.1 & 49.4 & 54.6 & 71.0 & 98.7 & 64.0 & 57.0 & 51.3 & 65.2 & 53.4 & 64.4 & 69.8 & \cellcolor{softergreen}{\underline{45.9}} & 58.5 & \cellcolor{softergreen}{\textbf{44.5}} \\
ViT-B16-In21k-augreg & 84.5 & 56.8 & 45.2 & 48.9 & 45.4 & 38.1 & 94.1 & 67.7 & 59.0 & 43.1 & 50.1 & 48.2 & 49.8 & \textbf{31.3} & 35.7 & \underline{35.2} & 37.1 \\
ViT-B16-In21k-orig & 81.8 & 52.2 & 39.2 & 39.0 & 41.0 & \underline{35.6} & 71.4 & 52.7 & 47.6 & 38.3 & 56.9 & 48.2 & 44.5 & 35.6 & \cellcolor{softergreen}{\textbf{31.6}} & 38.3 & \cellcolor{softergreen}{36.5} \\
ViT-B16-In21k-miil & 84.3 & 57.2 & 46.4 & 46.5 & 49.3 & 46.1 & 98.0 & 59.6 & 51.7 & 43.6 & 59.4 & 50.0 & 58.1 & 56.2 & \cellcolor{softergreen}{\textbf{35.4}} & 48.6 & \cellcolor{softergreen}{\underline{40.2}} \\
ViT-L16-In21k-augreg & 85.8 & 47.0 & 39.0 & 27.3 & 37.7 & 31.7 & 95.2 & 68.6 & 58.9 & 35.4 & 52.7 & 37.6 & 40.3 & \textbf{24.2} & 28.9 & \underline{26.5} & 28.1 \\
ViT-L16-In21k-orig & 81.5 & 46.2 & 37.3 & 37.1 & 37.7 & 42.2 & 58.5 & 45.8 & 40.7 & 36.4 & 55.8 & 42.1 & 40.1 & 39.4 & \cellcolor{softergreen}{\textbf{32.4}} & 37.6 & \cellcolor{softergreen}{\underline{36.1}} \\
ViT-S16-augreg & 78.8 & 72.8 & 72.8 & 73.6 & 72.5 & 80.7 & 97.0 & 82.1 & 80.0 & 72.7 & 71.0 & 72.8 & 75.4 & 63.2 & \cellcolor{softergreen}{63.1} & \underline{59.2} & \cellcolor{softergreen}{\textbf{58.9}} \\
ViT-B16-augreg & 79.2 & 72.2 & 71.7 & 69.6 & 71.1 & 73.5 & 90.9 & 77.6 & 75.9 & 70.9 & 65.5 & 72.0 & 73.8 & 62.9 & \cellcolor{softergreen}{61.3} & \underline{58.4} & \cellcolor{softergreen}{\textbf{57.4}} \\
ViT-B16-CLIP-L2b-In12k & 86.2 & 49.7 & 44.7 & 42.6 & 44.0 & 49.6 & 99.9 & 49.4 & 42.3 & 40.9 & 50.5 & 41.4 & 48.4 & 57.2 & \cellcolor{softergreen}{\textbf{35.8}} & 49.0 & \cellcolor{softergreen}{\underline{38.9}} \\
ViT-L14-CLIP-L2b-In12k & 88.2 & 35.5 & 28.8 & 28.1 & 29.9 & \textbf{24.5} & 97.9 & 39.5 & \underline{25.4} & 29.7 & 41.5 & 26.8 & 31.4 & 35.3 & \cellcolor{softergreen}{25.4} & 30.3 & \cellcolor{softergreen}{27.2} \\
ViT-H14-CLIP-L2b-In12k & 88.6 & 36.4 & 31.1 & 30.8 & 31.6 & \underline{24.9} & 97.0 & 41.7 & 27.4 & 31.5 & 53.4 & 27.4 & 33.6 & 33.5 & \cellcolor{softergreen}{\textbf{23.7}} & 29.5 & \cellcolor{softergreen}{26.1} \\
ViT-so400M-SigLip & 89.4 & 50.3 & 47.4 & 42.1 & 44.2 & 40.2 & 95.5 & 36.3 & 30.0 & 42.6 & 65.1 & 30.9 & 36.1 & 36.4 & \cellcolor{softergreen}{\underline{27.4}} & 31.3 & \cellcolor{softergreen}{\textbf{26.1}} \\
\hline
\textbf{Average} & 84.4 & 58.9 & 58.6 & 57.6 & 55.2 & 54.9 & 92.9 & 61.5 & 55.1 & 52.9 & 61.0 & 52.0 & 58.1 & 53.8 & \cellcolor{softergreen}{\textbf{42.9}} & 49.2 & \cellcolor{softergreen}{\underline{43.2}} \\
\end{tabular}

    \label{tab:fpr-big-NINCO}
\end{table*}

\textbf{CIFAR} We investigate \ourMethod{} on CIFAR100 \cite{Krizhevsky09learningmultipleCIFAR}, following the OpenOOD setup with tiny ImageNet \cite{tinyImageNet}, Mnist \cite{lecun1998gradientMNIST}, SVHN \cite{netzer2011readingSVHN}, Texture \cite{cimpoi2014describing}, Places \cite{zhou2017places} and Cifar10 as OOD datasets for a range of architectures and training schemes.

We report results averaged across the OOD datasets in Table~\ref{tab:fpr-cifar100} for the most competitive methods and standard baselines (full results in Appendix~\ref{app:extended-results}). We observe that \ourMethod{} consistently outperforms the conventional Mahalanobis distance, but the differences are smaller compared to the ImageNet setup. 
We hypothesize that this is because the problems of the Mahalanobis distance are less drastic at a smaller scale, and therefore the conventional Mahalanobis distance is already fairly effective for OOD detection. 
ViM and KNN are the most competitive baseline methods, but \ourMethod{} remains the most consistent and effective method across models.

\section{Conclusion}\vspace{-0.55mm}
We showed that the frequently occurring failure cases of the Mahalanobis distance as an OOD detection method are related to violations of the method's basic assumptions. We showed that the feature norms vary much stronger than expected under a Gaussian model, that the feature distributions are strongly heavy-tailed and that feature norms correlate with the Mahalanobis score - irrespective of whether a sample is ID or OOD. These insights explain why certain models - despite impressive ID classification performance - showed strongly degraded OOD detection results with the Mahalanobis score in previous studies \cite{bitterwolf2023ninco}. We introduced \ourMethod{}, a simple remedy consisting of $\ell_2$ normalization that effectively mitigates those problems. In particular, the resulting feature distributions are more aligned with a normal distribution, less heavy-tailed, and the class variances are more similar,
leading to improved OOD detection results across a wide range of models. \ourMethod{} outperforms the conventional Mahalanobis distance in 41/44 cases, rendering it clearly the most effective method across models. It outperforms the previously best baseline ViM by 7 FPR points on average on the OpenOOD datasets, and is the best method for 4  of the 5 top models.

\section*{Impact Statement}
This paper presents work whose goal is to advance the field of 
Machine Learning. There are many potential societal consequences 
of our work, none of which we feel must be specifically highlighted here.

\section*{Acknowledgements}
We thank Yannic Neuhaus for insightful discussions about distances on unit spheres.
Further, we acknowledge support from 
the DFG (EXC number 2064/1, Project number 390727645)
and 
the Carl Zeiss Foundation
in the project 
"Certification and Foundations of Safe Machine Learning Systems in Healthcare".
Finally, we acknowledge support from the German Federal Ministry of Education and Research (BMBF) through the Tübingen AI Center (FKZ: 01IS18039A)
and the European Laboratory for Learning and Intelligent Systems (ELLIS).

\bibliography{main}
\bibliographystyle{icml2025}

\newpage
\appendix
\onecolumn
\section{Overview}
The Appendix is structured as follows:\\
In Section~\ref{app:proof-gauss} we provide the proof for Lemma~\ref{le:Gauss}.
In Section~\ref{app:featurenorm} we provide extended analysis on feature norm and normalization. In particular, 
\begin{itemize}  \setlength\itemsep{0pt}
    \item we show the feature norm distribution for more models in Figure~\ref{fig:feature-norm-dist-vits-four}
    \item we provide QQ-plots for more models in Figure~\ref{fig:qq-many}
    \item We report the feature norm distribution for ID and OOD data in Figure~\ref{fig:fnorm-OOD-larger}, showing that OOD features can be larger than ID features for off-the-shelf pretrained models
    \item we highlight that the class variances become more similar to the global variance after normalization in Figure~\ref{fig:var-class-global}
    \item we plot the correlation between feature norm and OOD-score in Figure~\ref{fig:fnorm-manymodels}
\end{itemize}
In Section~\ref{app:extended-results} we report extended results. In particular, 
\begin{itemize}
    \item we show additional ImageNet numbers (AUC for NINCO in Table~\ref{tab:auc-ninco}, OpenOOD near and far in Table~\ref{tab:fpr-OpenOOD-near} and Table~\ref{tab:fpr-OpenOOD-far}, OpenOOD averaged AUC in Table~\ref{tab:AUC-openood}) 
    \item we compare cosine-based methods on ImageNet explicitly in Table~\ref{tab:cosine-methods-fpr}
    \item we show robustness to noise distributions (unit tests) in Table~\ref{tab:unit-tests-big}
    \item we show additional CIFAR numbers (Cifar10 AUC in Table~\ref{tab:auc-cifar10} and FPR in Table~\ref{tab:fpr-cifar10}, Cifar100 AUC in Table~\ref{tab:auc-cifar100} and FPR in Table~\ref{tab:fpr-cifar100-long}
    \item we compare \ourMethod{} to SSD+ in Table~\ref{tab:ssd-comparison} to highlight the benefits of post-hoc OOD detection methods
\end{itemize}
In Section \ref{app:models} we report details on the model checkpoints used throughout the experiments (ImageNet models in Table~\ref{tab:models-imagenet} and Cifar models in Table~\ref{tab:models-cifar}). 
In Section~\ref{sec:methods} we provide details on the OOD detection methods evaluated in the main paper. 

\section{Proof of Lemma \ref{le:Gauss}}\label{app:proof-gauss}
\begin{proof} 
Let $\Sigma=U\Lambda U^T$ be the eigendecomposition of the covariance matrix with $U$ being an orthogonal matrix containing the eigenvectors of $\Sigma$ and $\Lambda$ the diagonal matrix containing the eigenvalues of $\sigma$. Let $X$ be a random variable with distribution $\mathcal{N}(\mu, \Sigma)$ (in the main paper, we denoted the features as $\Phi(X)$, here we write them as $X$ for notational simplicity).
Then it holds $Z=U^TX$ has distribution $\mathcal{N}(U^T \mu, \Lambda)$ and since $U^T$ is an orthogonal matrix: $\norm{X}^2_2=\norm{Z}^2_2$. We have
\begin{align*}
\Exp[\norm{X}^2_2]&=\Exp[\norm{Z}^2_2] = \sum_{i=1}^d \Exp[Z_i^2] = \sum_{i=1}^d \mathrm{Var}(Z_i) + \Exp[Z_i]^2 = \sum_{i=1}^d \lambda_i + \norm{U^T \mu}^2_2 = \mathrm{tr}(\Sigma)+\norm{\mu}^2_2 
\end{align*}
We note that 
\begin{align}
\mathrm{Var}(\norm{Z}^2_2) & = \Exp[\norm{Z}^4_2] - \Exp[\norm{Z}^2_2]^2\ =\Exp[\norm{Z}^4_2]-(\mathrm{tr}(\Sigma)+\norm{\mu}^2_2)^2
\end{align}
and it remains to compute $\Exp[\norm{Z}^4_2]$. We note that
\begin{align*}
\Exp[\norm{Z}_2^4] &= \sum_{i=1}^d \Exp[Z_i^4] + \sum_{i\neq j}^d \Exp[Z_i^2]\Exp[Z_j^2]=\sum_{i=1}^d (3\lambda_i^2 +6 \mu_i^2 \lambda_i +\mu_i^4)+ \left(\sum_{i=1}^d (\lambda_i+\mu_i^2)\right)^2 -\sum_{i=1}^d (\lambda_i+\mu_i^2)^2,
\end{align*}
where we have used the following calculations:
\begin{align*}
0&=\Exp[(Z_i-\mu_i)^3] =\Exp[Z_i^3]-3\mu_i\lambda_i-\mu_i^3\\
3\lambda_i^2&=\Exp[(Z_i-\mu_i)^4]=\Exp[Z_i^4]-4\mu_i \Exp[Z_i^3]+6\mu_i^2\Exp[Z_i^2]-3\mu_i^4
\end{align*}
and thus
\begin{align*}
\Exp[Z_i^3] &=3\mu_i\lambda_i + \mu_i^3\\
\Exp[Z_i^4] &=3\lambda_i^2 + 6\mu_i^2\lambda_i+\mu_i^4
\end{align*}
This yields
\begin{align*}
\mathrm{Var}(\norm{Z}^2_2) & =\sum_{i=1}^d (3\lambda_i^2+6 \mu_i^2 \lambda_i +\mu_i^4)-\sum_{i=1}^d (\lambda_i+\mu_i^2)^2
\end{align*}
Applying Chebychev's inequality yields the result.
\end{proof}

\section{Derivation of expected squared relative variance deviation}\label{app:proof-exp}
Here we want to derive the statement about the expected squared relative variance (denoting the covariance matrix as $C$ instead of $\Sigma$):
\begin{align}\Exp_u[(u^T C^{-\frac{1}{2}}(C_i-C)C^{-\frac{1}{2}}u)^2] =\frac{2 \mathrm{trace}(A^2) + \mathrm{trace}(A)^2}{d(d+2)},
\end{align}
where $u$ has a uniform distribution on the unit sphere and $A=C^{-\frac{1}{2}}(C_i-C)C^{-\frac{1}{2}}$.
We note that $A$ is symmetric and thus has an eigendecomposition $A=U\Lambda U^T$. We have
\[ \Exp_u[(u^T A u)^2] = \Exp_u[ \left( (U^Tu)^T \Lambda (U^T u)\right)^2] = \Exp_u[ (u^T \Lambda u)^2] = \sum_{i=1}^d  \lambda_i^2 \Exp_u[ u_i^4] + \sum_{i \neq j}\lambda_i \lambda_j \Exp_u[u_i^2 u_j^2] \]
It remains to compute these moments on the unit sphere. For this purpose we note that $\norm{u}^2_2 = \sum_{i=1}^d u_i^2=1$ and thus
\[ 1=\norm{u}_2^4 = \left(\sum_{i=1}^d u_i^2 \right)^2 = \sum_{i=1}^d u_i^4 +\sum_{i \neq j} u_i^2 u_j^2\]
We note that $u_i^4$ for $i=1,\ldots,d$ and $u_i^2 u_j^2$ for $i\neq j$ are all equally distributed and thus for $i\neq j$
\begin{align}\label{eq:rel}
 1 = d \Exp[u_i^4] + d(d-1)\Exp[u_i^2 u_j^2]
 \end{align}
Moreover, we note that rotations do not change the distribution for a unifom distribution on the sphere and thus 
$(u_i,u_j)$ and $\left(\frac{u_i-u_j}{\sqrt{2}},\frac{u_i+u_j}{\sqrt{2}}\right)$ have the same distribution and 
\[ \Exp[u_i^2 u_j^2] =\Exp\left[ \left(\frac{u_i-u_j}{\sqrt{2}}\right)^2\left(\frac{u_i+u_j}{\sqrt{2}}\right)^2\right] =\frac{1}{2}\Exp[u_i^4] - \frac{1}{2}\Exp[u_i^2 u_j^2].\]
This yields $\Exp[u_i^4]=3\Exp[u_i^2 u_j^2]$. Plugging this into \eqref{eq:rel} yields
\[ \Exp[u_i^4] = \frac{3}{d(d+2)}, \quad \Exp[u_i^2 u_j^2]=\frac{1}{d(d+2)}\]
Thus
\[ \Exp_u[(u^T A u)^2] = \frac{1}{d(d+2)}\left(3 \sum_{i=1}^d \lambda_i^2 + \sum_{i \neq j} \lambda_i \lambda_j\right) = \frac{1}{d(d+2)}\left(2 \sum_{i=1}^d \lambda_i^2 + \sum_{i,j=1}^d \lambda_i \lambda_j\right)\]
Using that $\mathrm{trace}(A)=\sum_{i=1}^d \lambda_i$ finishes the derivation.
\FloatBarrier
\section{Extended Analysis}\label{app:featurenorm}
Here, we report extended results on the experiments of Section~\ref{sec:featurenorm} of the main paper. In particular, we show that the observations made hold beyond the SwinV2 model. If not stated differently, all experiments are with ImageNet as ID dataset and NINCO as OOD dataset.

\begin{figure}[htb]
    \centering
    \includegraphics[width=.5\linewidth]{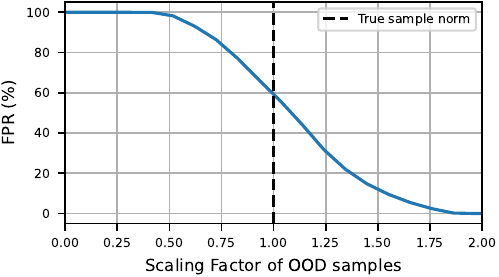}\vskip -0.1in
    \caption{\textbf{Impact of the feature norm of OOD samples on their Mahalanobis score.} When scaling down the norm of the features while leaving the feature direction unchanged, OOD samples receive a smaller Mahalanobis score and are incorrectly classified as ID samples. When the feature norm is artificially increased, the opposite happens.}
    \label{fig:scaled-ood}\vskip -0.1in
\end{figure}

\begin{figure}
    \centering
    \includegraphics[width=1.\linewidth]{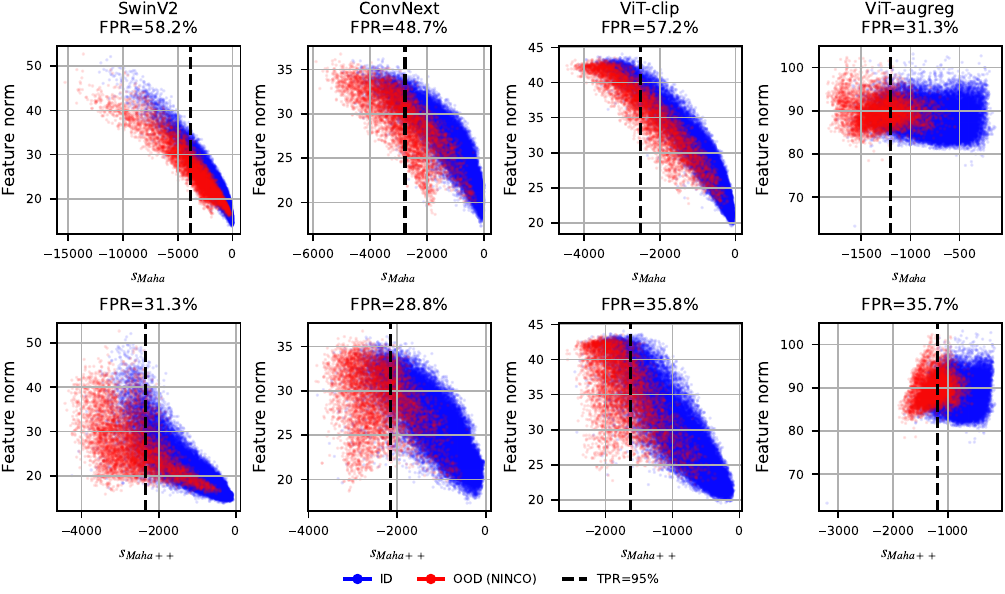}
    \caption{\textbf{\ourMethod{} resolves feature-norm dependency of Mahalanobis score.} With unnormalized features, OOD samples with small pre-logit feature norm were systematically identified as ID, but after normalization, OOD samples with small feature norm are rightfully detected as OOD, resulting in significantly improved OOD detection with \ourMethod{}. The only exception is an \textit{augreg} ViT, which does not show a correlation between feature norm and Mahalanobis score, even without normalization.}
    \label{fig:fnorm-manymodels}
\end{figure}

\textbf{Feature Norm Correlation. }
In the main paper, we showed that for a SwinV2 model, the feature norm of a sample correlates strongly with the OOD score received via the Mahalanobis distance. Here, we show this phenomenon for more models.
In Figure~\ref{fig:fnorm-manymodels}, we plot the feature norm against the OOD score assigned by Mahalanobis and \ourMethod{} for four models. For SwinV2, ConvNext and ViT-clip, the feature norms correlate strongly with the OOD score. Normalizing the features (bottom) mitigates this dependency, as OOD samples with small feature norms are detected as OOD, and thus improves OOD detection strongly. A notable exception is the augreg-ViT, for which there is no correlation between feature norms and OOD score. 
In Figure~\ref{fig:scaled-ood}, we further investigate the dependency of the Mahalanobis score on the feature norm by artificially scaling the feature norm of the OOD features with a prefactor $\alpha$. That is, for a SwinV2 model, we use $\alpha*\phi_i$ for each OOD feature and leave the ID validation features unchanged. We report the FPR values against $\alpha$ in Figure \ref{fig:scaled-ood}, and again observe a clear correlation between the scaling factor and the FPR: Perhaps unexpectedly, upscaling the features reduces the false-positive rate, up to a scaling factor of 2, where zero false positives are achieved. 
When scaling down the feature norm, the FPR increases, and for $\alpha\approx0.5$, i.e. at half the original feature norm, all OOD samples are identified as ID. Notably, this does not change for smaller $\alpha$ values, not even for $\alpha=0$, where all OOD samples collapse to the zero vector. In other words, everything in the vicinity of the origin is identified as in-distribution, which contradicts the intuition of tight Gaussian clusters centered around class means. In the main paper, we hypothesized that this might explain why the Mahalanobis distance sometimes fails to detect the unit tests since those might receive a small feature norm.
In Figure~\ref{fig:fnorm-OOD-larger}, we plot the feature norms of different datasets for a range of models with (top) and without (bottom) pretraining. We find that the feature norms of natural OOD images like those from NINCO tend to be even larger than the ImageNet feature norms. This violates basic assumptions in feature-norm-based OOD detection methods like the negative-aware-norm \cite{Park23UnderstandingFeatureNorm}, indicating that special training schemes might be necessary for those methods. However, noise distributions like the unit tests from \citet{bitterwolf2023ninco} can lead to fairly small feature norms for most models. Since we showed that small feature norms lead to small Mahalanobis distances for many models, this highlights why these supposedly easy-to-detect images were not detected with the Mahalanobis distance in previous studies.

\begin{figure}
    \centering
    \includegraphics[width=1.\linewidth]{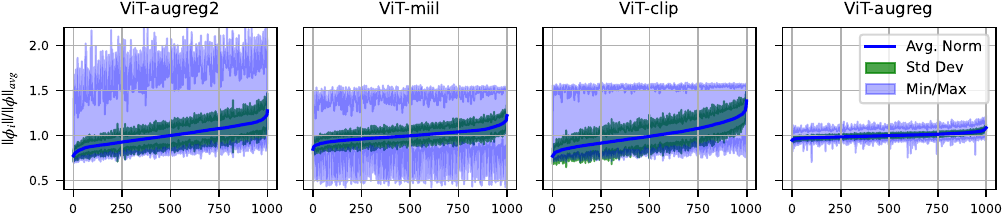}
    \caption{\textbf{For most models, the feature norms vary strongly across and within classes.} The same plot as the \textit{"observed"} part of Figure~\ref{fig:fnorm-maha} in the main paper, but normalized by the global average feature norm to make the scales of different models comparable. We thus show how strongly the feature norms vary relative to their scale. We report results for ViT-B16 models with different pretraining schemes. Only the \textit{augreg} ViT shows little variation in feature norm and is the only model that does not benefit from normalization. Interestingly, the \textit{augreg2} model was finetuned on ImageNet-1k from the \textit{same} 21k-checkpoint as the \textit{augreg} model and even achieves higher classification accuracy, but shows a very different feature norm distribution - which reflects in the OOD detection performance with Mahalanobis and \ourMethod{}: All models except for the \textit{augreg} model benefit from normalization.}    \label{fig:feature-norm-dist-vits-four}
\end{figure}

\textbf{Feature norm distribution. }In Figure~\ref{fig:feature-norm-dist-vits-four}, we plot the feature norms for four ViTs of exactly the same architecture (ViT-B16). In order to make the plots comparable, we normalize by the average feature norm per model. We observe that, like for the SwinV2 in the main paper, the norms vary strongly across and within classes - except for the augreg-ViT. This model is one of the models that performed well with Mahalanobis "out-of-the-box", i.e., not requiring normalization. 

\begin{figure}
    \centering
    \includegraphics[width=1.\linewidth]{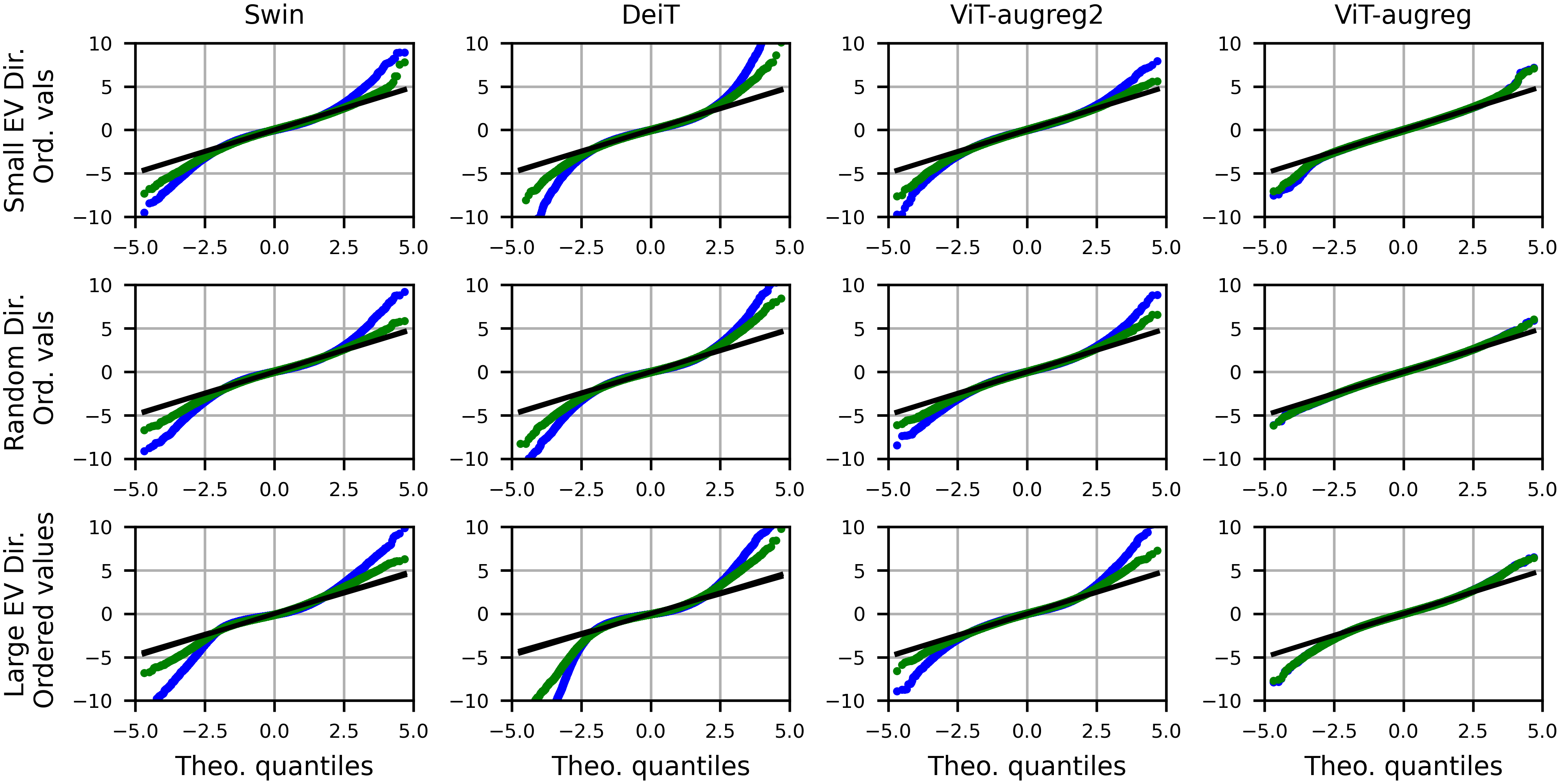}
    \caption{
    \textbf{QQ-plot: $\ell_2 -$normalization helps transform the features to be more aligned with a normal distribution.} Normalized features in green, unnormalized features in blue. For a SwinV2, DeiT3 and ViT-augreg2, the feature norms vary strongly across classes (see e.g. Fig.~\ref{fig:fnorm-variations} and Fig.~\ref{fig:feature-norm-dist-vits-four}) and normalization shifts the distribution towards a Gaussian. For a ViT-B-augreg the feature norms are similar across classes (see Fig~\ref{fig:feature-norm-dist-vits-four}) and the feature norms are already fairly normal, so $\ell_2$-normalization has almost no effect.
    }  
    \label{fig:qq-many}
\end{figure}

\begin{figure*}[htb]
    \centering
    \includegraphics[width=1.\linewidth]{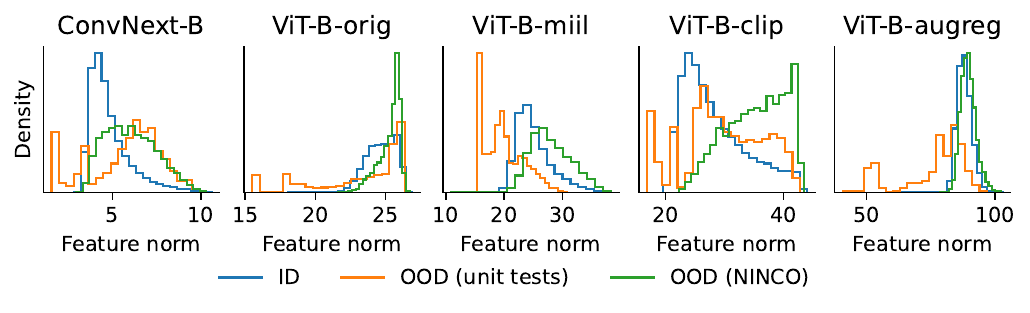}    
    \includegraphics[width=1.\linewidth]{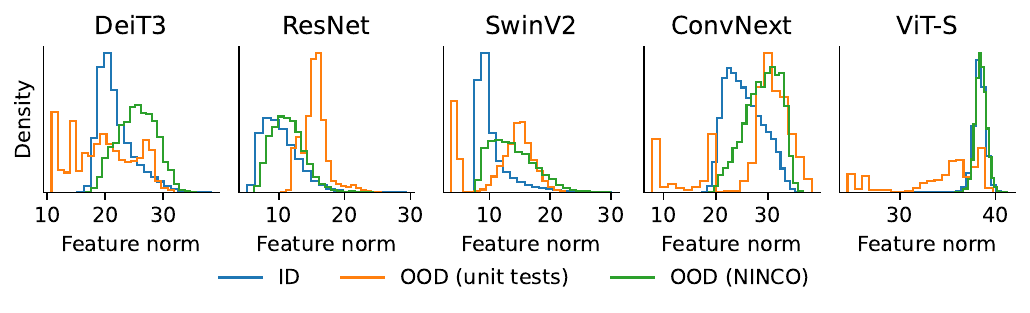}

    \caption{\textbf{Feature norm distribution.} In contrast to previous work (e.g., \cite{Park23UnderstandingFeatureNorm}), we find that the feature norm of natural OOD samples (NINCO in green) is often larger than that of ID samples (orange). Far-OOD data, like noise distributions, tend to have lower feature norms. This holds for models with (top) and without (bottom) pretraining.}
    \label{fig:fnorm-OOD-larger}
\end{figure*}

\textbf{QQ plots. }In Figure~\ref{fig:qq-many} we show QQ plots for four models along three directions in feature space. We observe that the normalized features (green) more closely resemble a normal distribution compared to the unnormalized features (blue), which is best visible via the long tail. The only exception is the augreg-ViT, for which normalized and unnormalized features are similarly close to a Gaussian distribution.

\begin{figure}
    \centering
    \includegraphics[width=.8\linewidth]{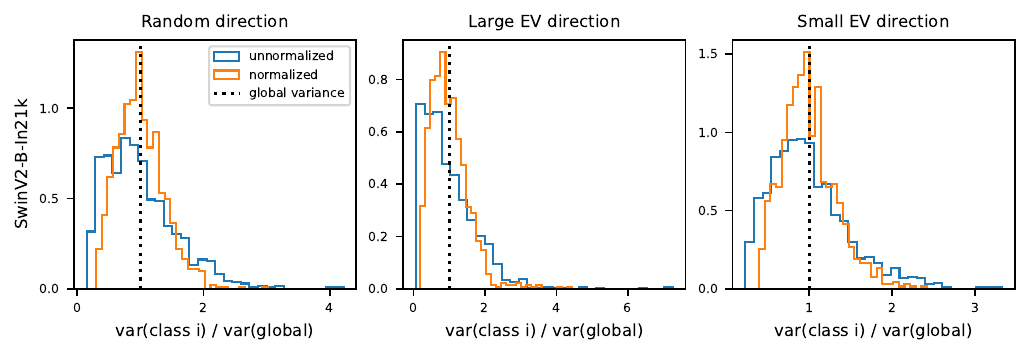}
    \includegraphics[width=.8\linewidth]{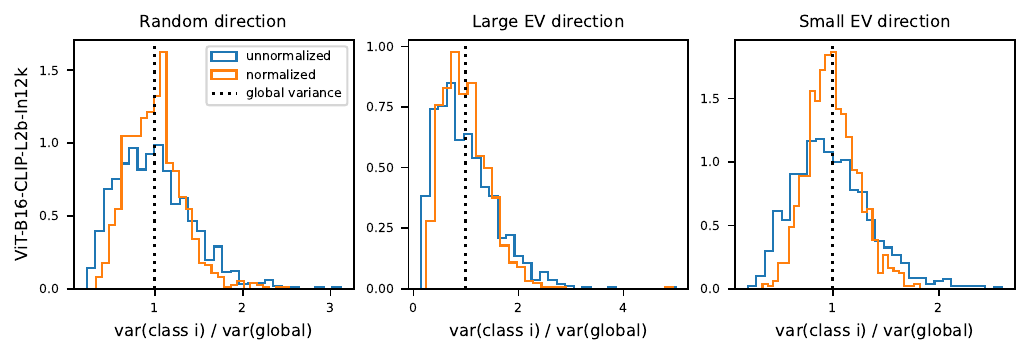}
    \includegraphics[width=.8\linewidth]{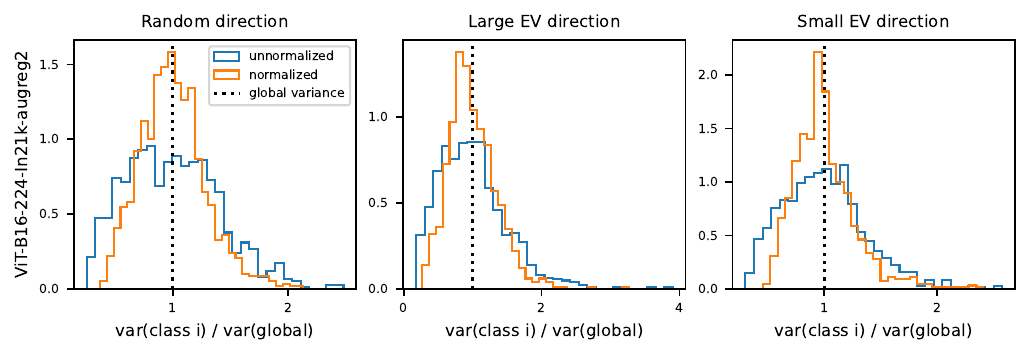}
    
    \includegraphics[width=.8\linewidth]{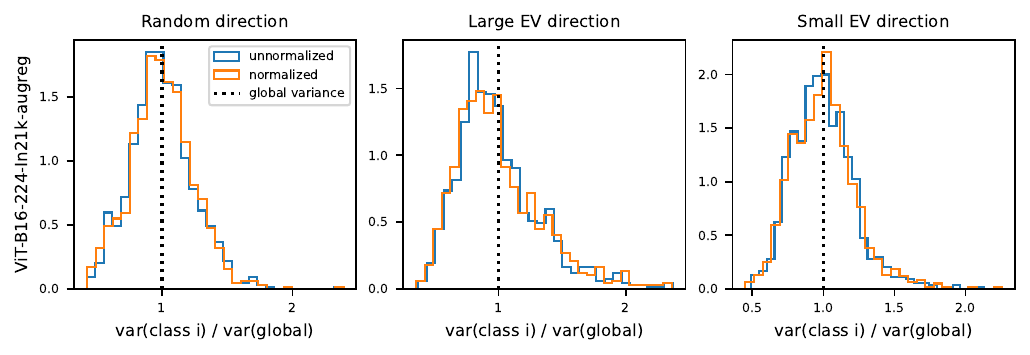}
    \caption{\textbf{\ourMethod{} aligns class-variances. } We report the distribution of the variances of the train features for each class along three directions: 1) a random direction, 2) a large eigendirection, 3) a small eigendirection. For each class, we compute the variance divided by the global variance, and plot the resulting distributions. Larger deviations from one indicate larger deviations of the class variance from the global variance. For all directions the distribution of variances is more peaked around 1 after normalization, indicating that after normalization the shared variance assumption is more appropriate - except for the ViT-augreg.}
    \label{fig:var-class-global}
\end{figure}

\textbf{Variance alignment. }
We report extended results on the expected relative deviation scores (see Eq.~\ref{eq:expectdeviation}) for more models in Table~\ref{tab:deviation-variances-avg}. We observe that for all models - except the ViT-augreg - normalization lowers the deviations, indicating a better alignment of the global variance with the individual class variances. 
In Figure~\ref{fig:var-class-global}, we illustrate this further: Instead of the score reported in Table~\ref{tab:deviation-variances-avg}, which computes an expectation over all directions, we pick three specific directions:  1) a random direction, 2) an eigendirection with a large eigenvalue, and 3) an eigendirection with a small eigenvalue. Ideally, along each direction, the 1000 class variances would coincide with the globally estimated, shared variance. For each direction, we divide the 1000 class variances by the global variance and plot the resulting distribution. Distributions peaked around 1 indicate that the global variance can capture the class variances well. We observe that the distributions of the variances after feature normalization peak more towards one for all models, except the ViT-augreg.

\setlength{\tabcolsep}{4.5pt}
\begin{table}[]
    \centering\small
    \caption{\textbf{Deviations from global variance.} We report the mean squared relative variance deviation as defined in Equation \eqref{eq:expectdeviation} for multiple models. In all cases, except for the ViT-augreg, normalization significantly improves the fit of the global covariance matrix to the covariance structure of the individual classes. As noted in the text for the ViT-augreg the features already follow very well the assumptions of the Mahalanobis score and normalization leads to no improvements.}
    \label{tab:deviation-variances-avg}
\begin{tabular}{lrr}
\toprule
                 model &  unnormalized &  normalized \\
\midrule
 ViT-B16-In21k-augreg & 0.05 &      0.05 \\
        SwinV2-B-In21k & 0.26 &      0.12 \\
ViT-B16-CLIP-L2b-In12k & 0.17 &      0.08 \\
 ViT-B16-In21k-augreg2 & 0.14 &      0.07 \\
    ViT-B16-In21k-miil & 0.12 &      0.09 \\
       DeiT3-B16-In21k & 0.24 &      0.15 \\
      ConvNeXt-B-In21k & 0.17 &      0.11 \\
       EVA02-B14-In21k & 0.21 &      0.14 \\
    ConvNeXtV2-B-In21k & 0.23 &      0.18 \\        
          ConvNeXtV2-B & 0.22 &      0.14 \\
      ConvNeXt-B-In21k & 0.17 &      0.11 \\
        ConvNeXt-B & 0.22 &      0.12 \\
\bottomrule
\end{tabular}
\end{table}

\textbf{Augreg ViTs. }
The ViTs that showed the best performance with Mahalanobis distance in previous studies were base-size ViTs pretrained on ImageNet21k and fine-tuned on ImageNet1k by \citet{steinerhowtotrainyourvit}. The training scheme is called \textit{augreg}, a carefully tuned combination of augmentation and regularization methods. In this paper, we made several observations regarding those models (applies for both base-size and large-size models with pretraining on ImageNet21k). In particular, they
\begin{itemize}
    \item show strong OOD detection performance with Mahalanobis distance without normalization, and normalization does not improve Mahalanobis-based OOD detection
    \item show little variations in feature norm compared to all other investigated models
    \item show no correlation between feature norm and Mahalanobis score (in contrast to all other investigated models)
\item show much weaker heavy tails than the other models
\item show low values for the variance deviation metric
    \item loose their advantage for unnormalized Mahalanobis-based detection when the fine-tuning scheme is changed (the augreg2 model is fine-tuned from a 21k-augreg-checkpoint, but the fine-tuning scheme differs in learning rate and augmentations)
\end{itemize}
In short, the augreg models omit all the points that we identified as problematic for Mahalanobis-based OOD detection. This indicates that the augreg training scheme induces a feature space that lends itself naturally towards a normal distribution, aligning well with the assumptions of the Mahalanobis distance as OOD detection method. Understanding the exact reason why the augreg scheme induces those features is beyond the scope of this paper. The connection of training hyperparameters and OOD detection performance was, however, investigated by \citet{mueller2024trainvitooddetection}. It should be stressed that for post-hoc OOD detection, we ideally want a method that works well with \textit{all} models, not only those obtained via a certain training scheme. We provide such a method with \ourMethod{}.

\FloatBarrier
\section{Extended results}\label{app:extended-results}
\begin{figure}
    \centering
    \includegraphics[width=0.5\linewidth]{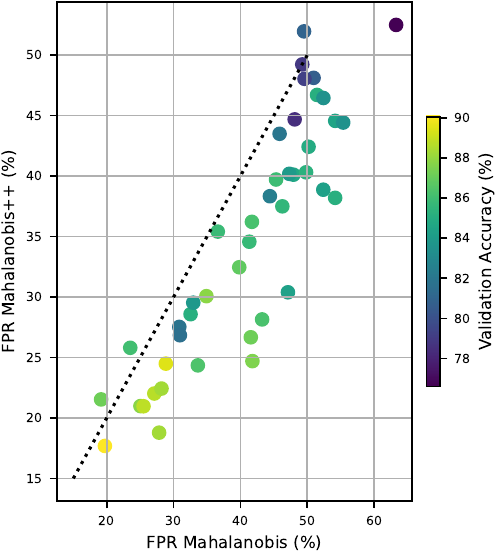}
    \caption{We plot the FPR with \ourMethod{} against the FPR with the conventional Mahalanobis score averaged over the five OpenOOD datasets. With three minor exceptions, \ourMethod{} improves OOD detection performance for all models. In particular, it significantly improves all models with high accuracy.}
    \label{fig:scatter-fpr-diffs}
\end{figure}

\setlength{\tabcolsep}{4pt}
\begin{table*}[h!]
\centering
\scriptsize 
      \caption{FPR on OpenOOD-near datasets, \colorbox{softergreen}{Green} indicates that normalized method is better than its unnormalized counterpart, \textbf{bold} indicates the best method, and \underline{underlined} indicates second best method. Maha++ improves over Maha on average by 9.6\% in FPR over all models. Similarly, rMaha++ is 5.5\% better in FPR than rMaha. 
      The lowest FPR is achieved by ViM for the EVA02-L14-M38m-In21k highlighted in {\color{blue}blue}, closely followed by Maha++ for the same model. }\vspace{1mm}
        \begin{tabular}{lccccccccccccccccc}
Model & Val Acc & MSP & E & E+R & ML & ViM & AshS & KNN & NNG & NEC & GMN & GEN & fDBD & Maha & Maha++ & rMaha & rMaha++ \\
\hline
ConvNeXt-B-In21k & 86.3 & 53.9 & 46.1 & 44.2 & 46.7 & 53.8 & 91.3 & 63.1 & 52.7 & \underline{42.2} & 53.8 & 46.7 & 58.3 & 59.6 & \cellcolor{softergreen}{\textbf{41.7}} & 51.8 & \cellcolor{softergreen}{44.7} \\
ConvNeXt-B & 84.4 & 70.0 & 89.0 & 86.8 & 75.1 & 72.1 & 99.0 & 77.5 & 72.1 & 72.1 & 72.9 & 73.3 & 76.1 & 72.0 & \cellcolor{softergreen}{\underline{59.1}} & 67.3 & \cellcolor{softergreen}{\textbf{58.1}} \\
ConvNeXtV2-T-In21k & 85.1 & 59.7 & 51.2 & 51.5 & 53.3 & \underline{45.4} & 95.7 & 62.7 & 55.6 & 48.3 & 56.2 & 54.4 & 61.7 & 52.7 & \cellcolor{softergreen}{\textbf{44.9}} & 52.1 & \cellcolor{softergreen}{48.8} \\
ConvNeXtV2-B-In21k & 87.6 & 50.8 & 39.0 & 38.8 & 41.5 & 37.3 & 95.5 & 49.9 & 42.6 & \underline{36.8} & 46.1 & 41.0 & 48.7 & 42.0 & \cellcolor{softergreen}{\textbf{34.7}} & 41.3 & \cellcolor{softergreen}{38.2} \\
ConvNeXtV2-L-In21k & 88.2 & 48.4 & 38.5 & 38.6 & 40.6 & 48.8 & 95.6 & 49.5 & 41.4 & 35.9 & 54.9 & 39.7 & 49.5 & 46.6 & \cellcolor{softergreen}{\textbf{31.2}} & 41.2 & \cellcolor{softergreen}{\underline{35.6}} \\
ConvNeXtV2-T & 83.5 & 72.7 & 79.1 & 75.8 & 72.9 & 72.6 & 98.6 & 86.7 & 81.0 & 69.7 & 74.6 & 72.4 & 78.4 & 73.3 & \cellcolor{softergreen}{\underline{60.5}} & 66.6 & \cellcolor{softergreen}{\textbf{58.2}} \\
ConvNeXtV2-B & 85.5 & 69.6 & 79.2 & 76.3 & 70.5 & 69.0 & 99.0 & 74.9 & 70.3 & 68.2 & 71.1 & 65.8 & 72.6 & 66.9 & \cellcolor{softergreen}{\underline{55.0}} & 61.5 & \cellcolor{softergreen}{\textbf{54.1}} \\
ConvNeXtV2-L & 86.1 & 69.8 & 77.4 & 73.7 & 69.9 & 70.6 & 98.4 & 70.3 & 66.8 & 68.8 & 77.8 & 64.1 & 68.2 & 62.4 & \cellcolor{softergreen}{\underline{53.6}} & 56.4 & \cellcolor{softergreen}{\textbf{51.9}} \\
DeiT3-S16-In21k & 84.8 & 73.1 & 65.8 & 64.8 & 67.4 & 68.7 & 98.8 & 70.8 & 67.1 & 65.4 & 64.5 & 65.5 & 72.6 & 70.7 & \cellcolor{softergreen}{\textbf{60.4}} & 68.4 & \cellcolor{softergreen}{\underline{60.6}} \\
DeiT3-B16-In21k & 86.7 & 67.1 & 60.6 & 57.4 & 61.2 & 65.5 & 98.9 & 63.2 & 57.4 & 58.1 & 59.1 & 53.4 & 62.2 & 62.7 & \cellcolor{softergreen}{\underline{50.8}} & 58.6 & \cellcolor{softergreen}{\textbf{49.9}} \\
DeiT3-L16-In21k & 87.7 & 66.7 & 54.3 & 51.2 & 57.1 & 57.3 & 97.2 & 53.9 & 49.4 & 52.2 & 52.9 & 48.1 & 56.9 & 54.5 & \cellcolor{softergreen}{\underline{47.3}} & 51.6 & \cellcolor{softergreen}{\textbf{46.3}} \\
DeiT3-S16 & 83.4 & 70.8 & 71.3 & 71.3 & 68.5 & 63.8 & 86.7 & 81.1 & 67.7 & 68.5 & 70.5 & 66.7 & 72.8 & 69.7 & \cellcolor{softergreen}{\underline{62.4}} & 65.4 & \cellcolor{softergreen}{\textbf{60.0}} \\
DeiT3-B16 & 85.1 & 71.8 & 89.5 & 90.1 & 76.8 & 67.3 & 98.7 & 79.6 & 83.6 & 76.1 & 71.9 & 65.4 & 74.7 & 71.8 & \cellcolor{softergreen}{\underline{64.8}} & 67.4 & \cellcolor{softergreen}{\textbf{61.7}} \\
DeiT3-L16 & 85.8 & 72.5 & 83.3 & 86.7 & 74.4 & 65.9 & 85.0 & 75.3 & 80.0 & 74.7 & 67.9 & 65.8 & 72.3 & 66.4 & \cellcolor{softergreen}{\underline{60.7}} & 61.9 & \cellcolor{softergreen}{\textbf{57.2}} \\
EVA02-B14-In21k & 88.7 & 45.1 & 37.0 & \underline{36.5} & 40.3 & 36.5 & 93.8 & 46.9 & 40.8 & 36.6 & 45.2 & 36.6 & 43.3 & 41.9 & \cellcolor{softergreen}{\textbf{34.8}} & 42.0 & \cellcolor{softergreen}{38.0} \\
EVA02-L14-M38m-In21k & 90.1 & 38.3 & 31.9 & 31.7 & 34.6 & \textbf{\color{blue}28.1} & 95.2 & 40.2 & 35.7 & 31.5 & 45.2 & 30.8 & 36.3 & 31.8 & \cellcolor{softergreen}{\underline{28.7}} & 33.3 & \cellcolor{softergreen}{32.3} \\
EVA02-T14 & 80.6 & 77.6 & 76.9 & 77.3 & 76.5 & 74.6 & 97.8 & 80.6 & 77.3 & 73.5 & \textbf{67.5} & 76.4 & 79.2 & 73.0 & \cellcolor{softergreen}{\underline{71.0}} & 72.7 & \cellcolor{softergreen}{71.2} \\
EVA02-S14 & 85.7 & 67.7 & 67.0 & 67.1 & 64.8 & 58.1 & 98.2 & 67.7 & 62.9 & 60.8 & \textbf{56.4} & 61.5 & 68.1 & 58.3 & \cellcolor{softergreen}{\underline{57.0}} & 58.7 & \cellcolor{softergreen}{57.3} \\
EffNetV2-S & 83.9 & 72.2 & 78.4 & 76.9 & 72.9 & 79.7 & 98.7 & 70.3 & 68.9 & 72.7 & 79.9 & 69.1 & 74.9 & 75.1 & \cellcolor{softergreen}{\underline{64.6}} & 67.7 & \cellcolor{softergreen}{\textbf{60.2}} \\
EffNetV2-L & 85.7 & 69.0 & 81.4 & 75.6 & 70.1 & 74.7 & 98.6 & 70.0 & 68.6 & 69.4 & 70.5 & 63.5 & 69.9 & 65.8 & \cellcolor{softergreen}{\underline{56.4}} & 59.5 & \cellcolor{softergreen}{\textbf{53.8}} \\
EffNetV2-M & 85.2 & 68.9 & 78.6 & 75.2 & 69.3 & 77.7 & 99.0 & 71.2 & 69.3 & 68.8 & 70.2 & 64.0 & 72.1 & 69.5 & \cellcolor{softergreen}{\underline{58.1}} & 61.5 & \cellcolor{softergreen}{\textbf{54.7}} \\
Mixer-B16-In21k & 76.6 & 82.4 & 87.7 & 87.8 & 84.6 & 83.3 & 94.9 & 89.4 & 87.9 & 84.8 & \underline{69.4} & 82.4 & 85.6 & 78.4 & \cellcolor{softergreen}{71.8} & 74.9 & \cellcolor{softergreen}{\textbf{68.8}} \\
SwinV2-B-In21k & 87.1 & 56.0 & 43.1 & \underline{41.9} & 45.6 & 63.1 & 84.5 & 69.8 & 57.6 & \textbf{40.7} & 60.9 & 47.0 & 62.5 & 69.8 & \cellcolor{softergreen}{46.7} & 61.1 & \cellcolor{softergreen}{48.4} \\
SwinV2-L-In21k & 87.5 & 54.0 & 45.4 & 44.4 & 46.4 & 66.6 & 87.2 & 67.9 & 57.3 & \textbf{41.6} & 72.6 & 47.4 & 62.2 & 69.3 & \cellcolor{softergreen}{\underline{44.0}} & 60.6 & \cellcolor{softergreen}{46.1} \\
SwinV2-S & 84.2 & 73.4 & 77.5 & 77.0 & 73.0 & 75.4 & 99.7 & 79.8 & 75.7 & 70.2 & 65.5 & 72.0 & 79.7 & 75.3 & \cellcolor{softergreen}{\underline{59.2}} & 71.1 & \cellcolor{softergreen}{\textbf{58.0}} \\
SwinV2-B & 84.6 & 73.9 & 77.8 & 75.3 & 73.1 & 73.6 & 98.4 & 76.0 & 73.4 & 70.6 & 66.3 & 69.6 & 76.4 & 69.5 & \cellcolor{softergreen}{\underline{59.6}} & 65.6 & \cellcolor{softergreen}{\textbf{57.8}} \\
ResNet101 & 81.9 & 78.4 & 87.0 & 99.7 & 80.4 & 82.1 & 91.5 & 82.6 & 75.0 & 81.1 & 87.4 & 78.8 & 87.6 & 73.1 & \cellcolor{softergreen}{\textbf{58.7}} & 62.8 & \cellcolor{softergreen}{\underline{59.9}} \\
ResNet152 & 82.3 & 76.9 & 85.9 & 99.7 & 79.2 & 81.4 & 90.7 & 81.0 & 72.2 & 79.8 & 84.4 & 76.8 & 86.3 & 71.3 & \cellcolor{softergreen}{\textbf{55.5}} & 60.6 & \cellcolor{softergreen}{\underline{58.6}} \\
ResNet50 & 80.9 & 80.7 & 95.0 & 99.6 & 82.5 & 84.1 & 91.5 & 88.3 & 81.9 & 83.4 & 91.0 & 80.7 & 88.5 & 75.5 & \cellcolor{softergreen}{65.4} & \underline{64.6} & \cellcolor{softergreen}{\textbf{62.4}} \\
ResNet50-supcon & 78.7 & 70.5 & \textbf{67.5} & \underline{67.7} & 67.7 & 87.4 & 71.5 & 77.3 & 69.4 & 67.8 & 85.5 & 71.2 & 74.1 & 98.2 & \cellcolor{softergreen}{71.3} & 91.3 & \cellcolor{softergreen}{68.5} \\
ViT-T16-In21k-augreg & 75.5 & 83.7 & 78.7 & 75.2 & 79.8 & 73.5 & 92.1 & 86.9 & 86.5 & 77.4 & 73.1 & 83.5 & 78.6 & \textbf{69.7} & \underline{71.6} & 74.3 & 75.0 \\
ViT-S16-In21k-augreg & 81.4 & 73.8 & 61.0 & 63.6 & 63.1 & \underline{56.6} & 88.7 & 77.2 & 72.1 & 60.6 & 62.4 & 69.3 & 66.6 & 56.7 & \cellcolor{softergreen}{\textbf{56.0}} & 61.3 & \cellcolor{softergreen}{60.8} \\
ViT-B16-In21k-augreg2 & 85.1 & 69.4 & 59.1 & 57.4 & 62.4 & 77.8 & 97.4 & 73.4 & 67.4 & 59.7 & 68.6 & 63.5 & 73.2 & 77.4 & \cellcolor{softergreen}{\underline{56.8}} & 68.1 & \cellcolor{softergreen}{\textbf{56.0}} \\
ViT-B16-In21k-augreg & 84.5 & 64.4 & 54.3 & 58.4 & 54.8 & \underline{47.9} & 95.3 & 74.9 & 68.0 & 52.8 & 55.9 & 57.7 & 58.7 & \textbf{44.6} & 49.0 & 48.3 & 49.9 \\
ViT-B16-In21k-orig & 81.8 & 60.6 & 44.8 & 44.9 & 48.0 & \underline{41.4} & 62.5 & 60.1 & 54.6 & 44.9 & 61.2 & 57.1 & 52.6 & 45.3 & \cellcolor{softergreen}{\textbf{40.6}} & 49.9 & \cellcolor{softergreen}{47.8} \\
ViT-B16-In21k-miil & 84.3 & 65.6 & 52.6 & 53.3 & 56.8 & 55.3 & 96.2 & 69.2 & 61.5 & \underline{52.1} & 62.4 & 60.1 & 67.3 & 66.3 & \cellcolor{softergreen}{\textbf{47.5}} & 60.1 & \cellcolor{softergreen}{52.1} \\
ViT-L16-In21k-augreg & 85.8 & 56.5 & 49.6 & \underline{38.3} & 48.9 & 42.6 & 96.3 & 73.2 & 65.9 & 46.5 & 54.2 & 48.7 & 50.8 & \textbf{36.4} & 41.9 & 39.2 & 40.8 \\
ViT-L16-In21k-orig & 81.5 & 53.1 & 40.2 & \underline{40.1} & 42.2 & 44.4 & 55.9 & 50.7 & 46.0 & 40.7 & 59.3 & 49.2 & 46.0 & 46.4 & \cellcolor{softergreen}{\textbf{39.7}} & 47.5 & \cellcolor{softergreen}{46.1} \\
ViT-S16-augreg & 78.8 & 78.3 & 79.3 & 80.2 & 78.6 & 85.5 & 96.6 & 86.8 & 85.0 & 78.9 & 75.6 & 78.9 & 81.2 & 69.8 & 69.9 & \underline{66.5} & \cellcolor{softergreen}{\textbf{66.4}} \\
ViT-B16-augreg & 79.2 & 77.7 & 78.1 & 75.5 & 77.4 & 79.0 & 93.2 & 83.1 & 81.4 & 77.3 & 69.2 & 78.0 & 79.8 & 69.4 & \cellcolor{softergreen}{68.3} & \underline{65.4} & \cellcolor{softergreen}{\textbf{64.7}} \\
ViT-B16-CLIP-L2b-In12k & 86.2 & 56.2 & 46.9 & 46.0 & 49.1 & 55.4 & 99.3 & 57.9 & 50.6 & \underline{45.9} & 52.5 & 49.1 & 56.6 & 65.0 & \cellcolor{softergreen}{\textbf{44.7}} & 59.2 & \cellcolor{softergreen}{49.2} \\
ViT-L14-CLIP-L2b-In12k & 88.2 & 44.1 & 34.8 & 34.3 & 37.2 & \textbf{32.0} & 96.9 & 48.6 & \underline{33.1} & 37.0 & 46.2 & 36.3 & 41.6 & 45.3 & \cellcolor{softergreen}{35.9} & 41.9 & \cellcolor{softergreen}{39.1} \\
ViT-H14-CLIP-L2b-In12k & 88.6 & 44.5 & 36.8 & 36.7 & 38.7 & \textbf{33.6} & 97.0 & 50.5 & \underline{35.5} & 38.6 & 55.5 & 36.9 & 44.1 & 44.4 & \cellcolor{softergreen}{35.8} & 41.6 & \cellcolor{softergreen}{38.9} \\
ViT-so400M-SigLip & 89.4 & 58.6 & 56.1 & 52.5 & 53.4 & 49.9 & 95.3 & 48.6 & 43.7 & 51.3 & 63.9 & 42.8 & 48.9 & 47.0 & \cellcolor{softergreen}{\textbf{38.6}} & 43.7 & \cellcolor{softergreen}{\underline{39.2}} \\
\hline
\textbf{Average} & 84.4 & 65.6 & 64.0 & 63.6 & 62.0 & 62.7 & 93.0 & 69.5 & 63.9 & 59.9 & 65.3 & 60.5 & 66.3 & 62.5 & \cellcolor{softergreen}{\textbf{52.9}} & 58.8 & \cellcolor{softergreen}{\underline{53.3}} \\
\end{tabular}

    \label{tab:fpr-OpenOOD-near}
\end{table*}

\setlength{\tabcolsep}{4pt}
\begin{table*}[h!]
\centering
\scriptsize 
 \caption{FPR on OpenOOD-far datasets, \colorbox{softergreen}{Green} indicates that normalized method is better than its unnormalized counterpart, \textbf{bold} indicates the best method, and \underline{underlined} indicates second best method. Maha++ improves over Maha on average by 6.1\% in FPR over all models. Similarly, rMaha++ is 1.2\% better in FPR than rMaha. In total, Maha++ improves the SOTA compared to the previously strongest method, ViM, by 5.1\%, which is significant. The lowest FPR is achieved by Maha++ for the EVA02-L14-M38m-In21k highlighted in {\color{blue}blue}.  }\vspace{1mm}
        \begin{tabular}{lccccccccccccccccc}
Model & Val Acc & MSP & E & E+R & ML & ViM & AshS & KNN & NNG & NEC & GMN & GEN & fDBD & Maha & Maha++ & rMaha & rMaha++ \\
\hline
ConvNeXt-B-In21k & 86.3 & 33.6 & 36.1 & 30.6 & 31.1 & \underline{13.3} & 86.6 & 20.0 & 17.9 & 24.1 & 54.5 & 23.2 & 24.3 & 16.3 & \cellcolor{softergreen}{\textbf{12.8}} & 18.2 & 19.4 \\
ConvNeXt-B & 84.4 & 55.7 & 92.2 & 86.9 & 67.0 & 39.9 & 99.8 & 46.2 & 37.2 & 62.7 & 74.5 & 51.4 & 49.8 & 42.3 & \cellcolor{softergreen}{\textbf{34.9}} & 38.5 & \cellcolor{softergreen}{\underline{37.0}} \\
ConvNeXtV2-T-In21k & 85.1 & 34.7 & 28.0 & 27.5 & 28.8 & \textbf{14.7} & 97.3 & 27.5 & 23.6 & 23.1 & 41.2 & 24.6 & 29.4 & 19.1 & \cellcolor{softergreen}{\underline{17.7}} & 23.0 & 23.1 \\
ConvNeXtV2-B-In21k & 87.6 & 26.9 & 19.4 & 18.7 & 20.7 & \underline{12.4} & 95.2 & 18.6 & 15.9 & 16.7 & 34.2 & 17.7 & 19.1 & 13.7 & \cellcolor{softergreen}{\textbf{11.9}} & 15.7 & \cellcolor{softergreen}{15.2} \\
ConvNeXtV2-L-In21k & 88.2 & 26.0 & 19.3 & 18.5 & 20.4 & 15.3 & 95.6 & 18.4 & 15.6 & 16.3 & 18.2 & 17.5 & 19.6 & 15.3 & \cellcolor{softergreen}{\textbf{10.5}} & 15.5 & \cellcolor{softergreen}{\underline{14.6}} \\
ConvNeXtV2-T & 83.5 & 52.4 & 57.4 & 47.1 & 49.6 & \underline{34.8} & 99.6 & 62.4 & 50.6 & 43.6 & 73.4 & 41.1 & 50.7 & 43.5 & \cellcolor{softergreen}{\textbf{33.7}} & 37.2 & \cellcolor{softergreen}{35.5} \\
ConvNeXtV2-B & 85.5 & 51.7 & 65.2 & 56.0 & 52.2 & 32.0 & 99.9 & 39.1 & 32.9 & 47.7 & 71.3 & 37.1 & 41.1 & 32.5 & \cellcolor{softergreen}{\textbf{25.8}} & 30.6 & \cellcolor{softergreen}{\underline{29.0}} \\
ConvNeXtV2-L & 86.1 & 51.2 & 61.7 & 51.0 & 50.6 & 33.6 & 99.6 & 34.6 & 30.0 & 47.2 & 54.5 & 34.4 & 35.3 & 27.9 & \cellcolor{softergreen}{\textbf{24.6}} & \underline{27.4} & 28.6 \\
DeiT3-S16-In21k & 84.8 & 52.1 & 44.9 & 40.8 & 45.7 & 33.6 & 99.5 & 35.8 & 34.4 & 42.4 & 43.6 & 36.0 & 42.2 & 36.5 & \cellcolor{softergreen}{\textbf{30.5}} & 35.9 & \cellcolor{softergreen}{\underline{32.2}} \\
DeiT3-B16-In21k & 86.7 & 48.5 & 50.7 & 38.2 & 45.8 & 24.9 & 99.5 & 24.8 & 24.3 & 38.6 & 37.7 & 29.5 & 29.1 & 24.6 & \cellcolor{softergreen}{\textbf{20.2}} & 24.1 & \cellcolor{softergreen}{\underline{21.2}} \\
DeiT3-L16-In21k & 87.7 & 47.2 & 39.7 & 29.6 & 40.0 & 22.4 & 98.7 & 22.5 & 20.7 & 29.3 & 26.1 & 25.6 & 25.8 & 21.9 & \cellcolor{softergreen}{\textbf{18.6}} & 21.3 & \cellcolor{softergreen}{\underline{19.0}} \\
DeiT3-S16 & 83.4 & 47.5 & 42.5 & 49.3 & 41.3 & \textbf{29.6} & 84.9 & 62.3 & 35.1 & 41.3 & 59.9 & \underline{33.1} & 41.8 & 40.9 & \cellcolor{softergreen}{35.8} & 38.3 & \cellcolor{softergreen}{34.8} \\
DeiT3-B16 & 85.1 & 51.6 & 77.4 & 87.2 & 56.1 & \textbf{29.6} & 99.6 & 57.2 & 63.1 & 55.1 & 57.8 & \underline{33.1} & 41.6 & 37.9 & \cellcolor{softergreen}{34.7} & 35.6 & \cellcolor{softergreen}{33.8} \\
DeiT3-L16 & 85.8 & 52.2 & 78.6 & 91.0 & 57.1 & 32.8 & 73.9 & 39.8 & 67.6 & 57.5 & 49.5 & 31.4 & 39.1 & 31.3 & \cellcolor{softergreen}{\textbf{25.7}} & 29.9 & \cellcolor{softergreen}{\underline{26.3}} \\
EVA02-B14-In21k & 88.7 & 23.9 & 20.0 & 19.3 & 21.0 & \underline{12.4} & 83.9 & 18.0 & 15.8 & 17.3 & 29.8 & 16.2 & 18.8 & 14.6 & \cellcolor{softergreen}{\textbf{11.8}} & 15.6 & \cellcolor{softergreen}{14.2} \\
EVA02-L14-M38m-In21k & 90.1 & 19.5 & 16.5 & 16.2 & 17.5 & \underline{11.2} & 88.2 & 16.1 & 14.2 & 15.3 & 36.4 & 13.2 & 15.6 & 11.7 & \cellcolor{softergreen}{\textbf{\color{blue}10.3}} & 13.0 & \cellcolor{softergreen}{12.4} \\
EVA02-T14 & 80.6 & 56.2 & 59.0 & 59.7 & 54.2 & \textbf{32.5} & 98.8 & 47.6 & 43.7 & 43.4 & 45.2 & 45.5 & 57.9 & 36.2 & \cellcolor{softergreen}{\underline{32.9}} & 39.3 & \cellcolor{softergreen}{37.1} \\
EVA02-S14 & 85.7 & 41.9 & 44.4 & 43.8 & 39.2 & \textbf{19.2} & 99.6 & 28.3 & 25.2 & 31.0 & 34.2 & 28.5 & 36.1 & 22.1 & \cellcolor{softergreen}{\underline{21.0}} & 24.4 & \cellcolor{softergreen}{23.1} \\
EffNetV2-S & 83.9 & 50.8 & 66.1 & 46.6 & 53.3 & 33.8 & 99.8 & 29.2 & 29.4 & 50.3 & 76.6 & 36.7 & 40.1 & 28.8 & \cellcolor{softergreen}{\textbf{23.9}} & 27.6 & \cellcolor{softergreen}{\underline{27.2}} \\
EffNetV2-L & 85.7 & 49.1 & 69.4 & 45.3 & 51.3 & 31.7 & 99.6 & 34.8 & 32.9 & 47.1 & 49.8 & 32.2 & 35.4 & 25.0 & \cellcolor{softergreen}{\textbf{20.0}} & 23.7 & \cellcolor{softergreen}{\underline{22.2}} \\
EffNetV2-M & 85.2 & 49.1 & 63.2 & 44.4 & 49.3 & 39.4 & 99.8 & 38.0 & 34.7 & 45.7 & 64.6 & 32.7 & 39.6 & 30.3 & \cellcolor{softergreen}{\textbf{23.1}} & 27.5 & \cellcolor{softergreen}{\underline{24.8}} \\
Mixer-B16-In21k & 76.6 & 64.2 & 79.8 & 80.7 & 68.5 & 64.1 & 96.4 & 70.1 & 81.2 & 69.3 & 55.8 & 57.8 & 62.6 & 53.3 & \cellcolor{softergreen}{\textbf{39.7}} & 50.0 & \cellcolor{softergreen}{\underline{42.3}} \\
SwinV2-B-In21k & 87.1 & 35.1 & 30.4 & 23.5 & 28.4 & 17.5 & 72.0 & 20.2 & \underline{16.3} & 21.1 & 55.5 & 21.7 & 22.3 & 22.7 & \cellcolor{softergreen}{\textbf{13.3}} & 20.8 & \cellcolor{softergreen}{18.8} \\
SwinV2-L-In21k & 87.5 & 31.3 & 29.7 & 22.4 & 26.6 & 20.7 & 83.7 & 19.6 & \underline{16.6} & 20.7 & 32.9 & 20.8 & 21.1 & 23.5 & \cellcolor{softergreen}{\textbf{11.9}} & 20.0 & \cellcolor{softergreen}{17.1} \\
SwinV2-S & 84.2 & 53.1 & 61.8 & 52.2 & 52.9 & 34.9 & 99.9 & 44.5 & 37.6 & 46.5 & 58.3 & 39.9 & 48.1 & 37.1 & \cellcolor{softergreen}{\textbf{25.3}} & 33.8 & \cellcolor{softergreen}{\underline{26.8}} \\
SwinV2-B & 84.6 & 54.6 & 58.5 & 46.9 & 52.1 & 34.1 & 99.6 & 41.0 & 36.2 & 45.2 & 49.3 & 36.8 & 43.9 & 33.5 & \cellcolor{softergreen}{\textbf{27.1}} & 31.6 & \cellcolor{softergreen}{\underline{27.6}} \\
ResNet101 & 81.9 & 60.5 & 79.9 & 99.5 & 64.3 & \underline{29.4} & 72.7 & 34.2 & 35.6 & 63.6 & 78.9 & 51.6 & 60.4 & \textbf{27.7} & 33.4 & 50.8 & 71.4 \\
ResNet152 & 82.3 & 59.3 & 79.5 & 99.4 & 63.8 & 28.5 & 72.8 & 32.8 & 29.9 & 62.0 & 72.3 & 49.3 & 58.0 & \textbf{26.5} & \underline{26.9} & 45.9 & 68.7 \\
ResNet50 & 80.9 & 66.2 & 96.5 & 99.2 & 71.3 & \underline{32.4} & 72.8 & 54.2 & 52.2 & 72.0 & 88.6 & 55.2 & 65.7 & \textbf{32.2} & 43.0 & 61.2 & 75.7 \\
ResNet50-supcon & 78.7 & 43.0 & 33.9 & 25.0 & 35.5 & 61.7 & \textbf{20.0} & 26.8 & \underline{23.5} & 34.4 & 74.4 & 41.7 & 30.7 & 93.7 & \cellcolor{softergreen}{26.6} & 89.5 & \cellcolor{softergreen}{60.5} \\
ViT-T16-In21k-augreg & 75.5 & 62.1 & 39.6 & \textbf{30.6} & 44.0 & 36.1 & 96.8 & 69.1 & 60.7 & 36.4 & 48.3 & 52.1 & 45.2 & 46.0 & \cellcolor{softergreen}{\underline{32.2}} & 49.1 & \cellcolor{softergreen}{46.1} \\
ViT-S16-In21k-augreg & 81.4 & 45.8 & 24.2 & 28.5 & 27.4 & \underline{18.0} & 68.8 & 41.2 & 33.4 & 23.1 & 32.2 & 31.4 & 29.0 & 23.4 & \cellcolor{softergreen}{\textbf{15.5}} & 30.9 & \cellcolor{softergreen}{27.1} \\
ViT-B16-In21k-augreg2 & 85.1 & 46.0 & 37.1 & 30.2 & 37.6 & 38.0 & 99.5 & 30.3 & \textbf{25.4} & 33.1 & 54.5 & 29.2 & 36.8 & 38.7 & \cellcolor{softergreen}{\underline{25.8}} & 33.0 & \cellcolor{softergreen}{27.8} \\
ViT-B16-In21k-augreg & 84.5 & 34.5 & 19.9 & 21.0 & 21.2 & \textbf{13.0} & 94.6 & 40.6 & 30.7 & 18.8 & 27.0 & 22.3 & 21.6 & \underline{13.0} & 14.5 & 19.1 & 19.3 \\
ViT-B16-In21k-orig & 81.8 & 33.9 & 21.3 & 21.5 & 23.2 & \underline{20.7} & 62.7 & 24.2 & 22.6 & 20.9 & 40.5 & 25.9 & 24.4 & 21.2 & \cellcolor{softergreen}{\textbf{18.8}} & 25.8 & \cellcolor{softergreen}{24.6} \\
ViT-B16-In21k-miil & 84.3 & 36.3 & 23.3 & 22.2 & 26.8 & 26.2 & 97.4 & 28.9 & 23.2 & \underline{21.7} & 53.5 & 23.7 & 29.5 & 34.4 & \cellcolor{softergreen}{\textbf{19.0}} & 32.6 & \cellcolor{softergreen}{26.3} \\
ViT-L16-In21k-augreg & 85.8 & 29.3 & 16.0 & 16.1 & 17.4 & \underline{11.0} & 93.4 & 35.6 & 24.7 & 15.7 & 33.1 & 18.7 & 16.9 & \textbf{10.7} & 11.8 & 15.8 & 15.8 \\
ViT-L16-In21k-orig & 81.5 & 32.7 & 22.1 & 22.0 & 23.6 & 21.0 & 44.9 & 22.8 & 22.0 & 21.8 & 40.3 & 25.8 & 24.3 & \underline{20.6} & \cellcolor{softergreen}{\textbf{18.3}} & 24.3 & \cellcolor{softergreen}{23.6} \\
ViT-S16-augreg & 78.8 & 55.8 & 45.4 & 47.5 & 47.6 & 56.6 & 97.1 & 61.3 & 58.2 & 47.7 & 52.5 & 50.3 & 53.8 & \underline{35.6} & \cellcolor{softergreen}{\textbf{35.5}} & 36.3 & \cellcolor{softergreen}{36.0} \\
ViT-B16-augreg & 79.2 & 55.3 & 47.2 & 43.3 & 48.6 & 53.0 & 88.2 & 53.8 & 52.5 & 48.2 & 54.3 & 50.5 & 52.7 & 36.4 & \cellcolor{softergreen}{\textbf{34.6}} & 35.8 & \cellcolor{softergreen}{\underline{34.7}} \\
ViT-B16-CLIP-L2b-In12k & 86.2 & 32.9 & 31.6 & 28.5 & 29.3 & 22.3 & 99.7 & 20.8 & \underline{19.0} & 26.0 & 34.0 & 22.9 & 25.7 & 28.7 & \cellcolor{softergreen}{\textbf{17.1}} & 23.9 & \cellcolor{softergreen}{21.1} \\
ViT-L14-CLIP-L2b-In12k & 88.2 & 23.2 & 18.8 & 18.1 & 19.3 & \underline{14.5} & 98.0 & 17.5 & 15.1 & 19.1 & 29.6 & 16.4 & 17.7 & 16.8 & \cellcolor{softergreen}{\textbf{13.4}} & 17.2 & \cellcolor{softergreen}{16.3} \\
ViT-H14-CLIP-L2b-In12k & 88.6 & 23.7 & 19.6 & 19.0 & 20.3 & \underline{14.8} & 98.3 & 18.3 & 15.3 & 20.1 & 51.3 & 16.3 & 18.7 & 15.6 & \cellcolor{softergreen}{\textbf{12.8}} & 16.9 & \cellcolor{softergreen}{15.9} \\
ViT-so400M-SigLip & 89.4 & 36.8 & 41.1 & 30.7 & 34.0 & 17.7 & 92.4 & 15.4 & \textbf{14.3} & 31.9 & 64.6 & 18.7 & 17.3 & 16.8 & \cellcolor{softergreen}{\underline{15.1}} & 16.3 & 16.4 \\
\hline
\textbf{Average} & 84.4 & 44.0 & 45.7 & 42.6 & 40.4 & \underline{28.1} & 89.1 & 35.1 & 32.1 & 36.7 & 50.3 & 32.3 & 35.4 & 29.1 & \cellcolor{softergreen}{\textbf{23.0}} & 30.5 & \cellcolor{softergreen}{29.3} \\
\end{tabular}

    \label{tab:fpr-OpenOOD-far}
\end{table*}

\setlength{\tabcolsep}{3.5pt}
\begin{table*}[h!]
\centering
\scriptsize 
    
     \caption{AUC on OpenOOD, \colorbox{softergreen}{Green} indicates that normalized method is better than its unnormalized counterpart, \textbf{bold} indicates the best method, and \underline{underlined} indicates second best method. Maha++ improves over Maha on average by 2.1\% in AUC over all models. Similarly, rMaha++ is 0.6\% better in AUC than rMaha. In total, Maha++ improves the SOTA compared to the previously strongest methods rMaha by 1.4\%, which is significant. The highest AUC is achieved by Maha++ for the EVA02-L14-M38m-In21k highlighted in {\color{blue}blue}.  }
     \vspace{1mm}
        \begin{tabular}{lccccccccccccccccc}
Model & Val Acc & MSP & E & E+R & ML & ViM & AshS & KNN & NNG & NEC & GMN & GEN & fDBD & Maha & Maha++ & rMaha & rMaha++ \\
\hline
ConvNeXt-B-In21k & 86.3 & 88.8 & 86.8 & 88.9 & 88.4 & \underline{93.3} & 52.1 & 90.1 & 92.6 & 91.6 & 88.1 & 91.5 & 90.3 & 91.9 & \cellcolor{softergreen}{\textbf{93.7}} & 92.0 & \cellcolor{softergreen}{92.5} \\
ConvNeXt-B & 84.4 & 80.9 & 61.3 & 71.2 & 74.7 & 85.4 & 19.8 & 84.4 & 86.8 & 76.4 & 81.6 & 83.2 & 84.7 & 87.1 & \cellcolor{softergreen}{\textbf{88.7}} & 87.9 & \cellcolor{softergreen}{\underline{88.7}} \\
ConvNeXtV2-T-In21k & 85.1 & 87.4 & 88.1 & 88.3 & 88.3 & \textbf{93.3} & 40.3 & 88.6 & 91.0 & 91.3 & 89.1 & 90.0 & 86.4 & 91.8 & \cellcolor{softergreen}{\underline{92.7}} & 91.2 & \cellcolor{softergreen}{91.5} \\
ConvNeXtV2-B-In21k & 87.6 & 89.9 & 90.4 & 91.0 & 90.4 & \underline{94.8} & 45.2 & 92.0 & 93.6 & 93.4 & 91.1 & 92.3 & 91.4 & 94.0 & \cellcolor{softergreen}{\textbf{94.9}} & 93.7 & \cellcolor{softergreen}{94.0} \\
ConvNeXtV2-L-In21k & 88.2 & 90.6 & 91.1 & 91.9 & 91.1 & 93.8 & 45.4 & 92.1 & 93.8 & 94.0 & 92.7 & 92.8 & 91.7 & 93.7 & \cellcolor{softergreen}{\textbf{95.3}} & 93.8 & \cellcolor{softergreen}{\underline{94.2}} \\
ConvNeXtV2-T & 83.5 & 83.5 & 78.1 & 82.2 & 81.7 & 86.7 & 21.6 & 81.2 & 85.3 & 83.5 & 82.2 & 86.6 & 84.9 & 87.0 & \cellcolor{softergreen}{\underline{88.9}} & 88.2 & \cellcolor{softergreen}{\textbf{89.1}} \\
ConvNeXtV2-B & 85.5 & 82.8 & 73.0 & 78.7 & 79.0 & 86.7 & 15.8 & 86.3 & 88.3 & 80.1 & 83.4 & 86.7 & 86.8 & 89.0 & \cellcolor{softergreen}{\textbf{90.5}} & 89.6 & \cellcolor{softergreen}{\underline{90.3}} \\
ConvNeXtV2-L & 86.1 & 83.0 & 73.0 & 79.9 & 78.9 & 84.5 & 17.3 & 87.2 & 88.8 & 78.9 & 85.8 & 87.3 & 88.0 & 89.8 & \cellcolor{softergreen}{\textbf{90.6}} & 90.3 & \cellcolor{softergreen}{\underline{90.6}} \\
DeiT3-S16-In21k & 84.8 & 80.3 & 78.7 & 81.2 & 79.5 & 87.1 & 19.5 & 86.0 & 87.4 & 81.1 & 87.9 & 85.2 & 85.5 & 87.9 & \cellcolor{softergreen}{\textbf{89.5}} & 88.1 & \cellcolor{softergreen}{\underline{89.2}} \\
DeiT3-B16-In21k & 86.7 & 82.4 & 74.8 & 82.2 & 78.4 & 90.6 & 18.4 & 89.7 & 90.4 & 82.7 & 89.8 & 87.4 & 88.7 & 91.0 & \cellcolor{softergreen}{\textbf{92.3}} & 91.1 & \cellcolor{softergreen}{\underline{92.1}} \\
DeiT3-L16-In21k & 87.7 & 84.3 & 81.5 & 86.6 & 82.7 & 92.0 & 20.5 & 91.1 & 91.4 & 88.3 & 90.9 & 89.3 & 89.4 & 92.0 & \cellcolor{softergreen}{\textbf{93.0}} & 91.9 & \cellcolor{softergreen}{\underline{92.8}} \\
DeiT3-S16 & 83.4 & 84.2 & 83.8 & 83.0 & 84.7 & 88.2 & 61.9 & 83.0 & 86.5 & 84.9 & 85.4 & 87.8 & 86.5 & 87.9 & \cellcolor{softergreen}{\underline{88.9}} & 88.6 & \cellcolor{softergreen}{\textbf{89.3}} \\
DeiT3-B16 & 85.1 & 83.5 & 69.5 & 63.3 & 79.0 & 87.6 & 22.4 & 83.1 & 79.1 & 79.8 & 85.1 & 87.6 & 86.1 & 88.1 & \cellcolor{softergreen}{\underline{88.9}} & 88.9 & \cellcolor{softergreen}{\textbf{89.6}} \\
DeiT3-L16 & 85.8 & 82.9 & 76.6 & 66.4 & 80.5 & 87.0 & 65.3 & 84.6 & 80.4 & 80.6 & 86.3 & 87.8 & 86.7 & 89.0 & \cellcolor{softergreen}{\underline{89.9}} & 89.7 & \cellcolor{softergreen}{\textbf{90.4}} \\
EVA02-B14-In21k & 88.7 & 91.0 & 90.4 & 91.3 & 90.9 & \textbf{95.0} & 54.0 & 92.2 & 93.6 & 93.4 & 92.5 & 93.3 & 92.6 & 94.0 & \cellcolor{softergreen}{\underline{94.9}} & 93.6 & \cellcolor{softergreen}{94.1} \\
EVA02-L14-M38m-In21k & 90.1 & 92.4 & 91.7 & 92.2 & 92.3 & \underline{95.9} & 48.0 & 93.6 & 94.6 & 94.4 & 92.0 & 94.3 & 94.1 & 95.5 & \cellcolor{softergreen}{\textbf{\color{blue}95.9}} & 95.1 & \cellcolor{softergreen}{95.3} \\
EVA02-T14 & 80.6 & 81.3 & 79.1 & 78.9 & 80.8 & \underline{87.2} & 43.1 & 81.7 & 85.0 & 85.0 & 86.4 & 85.0 & 80.0 & 86.6 & \cellcolor{softergreen}{\textbf{87.2}} & 86.3 & \cellcolor{softergreen}{86.7} \\
EVA02-S14 & 85.7 & 84.4 & 79.7 & 80.1 & 82.1 & \textbf{91.6} & 22.7 & 87.2 & 89.5 & 86.4 & 90.5 & 87.8 & 85.4 & 90.8 & \cellcolor{softergreen}{\underline{91.2}} & 90.4 & \cellcolor{softergreen}{90.7} \\
EffNetV2-S & 83.9 & 83.2 & 74.2 & 83.2 & 79.3 & 86.7 & 20.8 & 86.8 & 88.0 & 82.0 & 82.1 & 87.2 & 86.6 & 88.6 & \cellcolor{softergreen}{\underline{90.0}} & 89.7 & \cellcolor{softergreen}{\textbf{90.4}} \\
EffNetV2-L & 85.7 & 83.9 & 73.5 & 83.2 & 80.6 & 85.6 & 17.8 & 86.9 & 87.8 & 82.2 & 86.3 & 87.8 & 86.6 & 89.6 & \cellcolor{softergreen}{\underline{90.8}} & 90.7 & \cellcolor{softergreen}{\textbf{91.3}} \\
EffNetV2-M & 85.2 & 83.7 & 74.4 & 83.3 & 80.6 & 85.3 & 17.8 & 86.3 & 87.6 & 82.4 & 84.6 & 88.0 & 86.2 & 88.9 & \cellcolor{softergreen}{\underline{90.5}} & 90.2 & \cellcolor{softergreen}{\textbf{91.0}} \\
Mixer-B16-In21k & 76.6 & 80.4 & 78.9 & 78.7 & 79.8 & 82.5 & 48.6 & 79.3 & 79.3 & 79.9 & 84.0 & 82.2 & 81.3 & 83.1 & \cellcolor{softergreen}{\underline{86.4}} & 85.1 & \cellcolor{softergreen}{\textbf{86.4}} \\
SwinV2-B-In21k & 87.1 & 88.0 & 87.0 & 90.4 & 88.3 & \underline{92.1} & 47.5 & 88.8 & 91.8 & 91.9 & 87.0 & 91.4 & 90.1 & 90.6 & \cellcolor{softergreen}{\textbf{92.9}} & 90.8 & \cellcolor{softergreen}{91.9} \\
SwinV2-L-In21k & 87.5 & 88.7 & 86.8 & 90.8 & 88.1 & 91.9 & 38.4 & 89.8 & 92.2 & 91.8 & 88.9 & 91.6 & 90.8 & 91.1 & \cellcolor{softergreen}{\textbf{93.7}} & 91.5 & \cellcolor{softergreen}{\underline{92.7}} \\
SwinV2-S & 84.2 & 82.3 & 74.8 & 81.4 & 78.9 & 87.0 & 14.8 & 84.3 & 86.8 & 81.5 & 85.7 & 86.4 & 85.7 & 87.9 & \cellcolor{softergreen}{\textbf{90.0}} & 88.3 & \cellcolor{softergreen}{\underline{90.0}} \\
SwinV2-B & 84.6 & 82.4 & 75.9 & 82.8 & 79.4 & 86.1 & 21.2 & 85.7 & 87.2 & 82.0 & 86.7 & 87.2 & 86.1 & 88.8 & \cellcolor{softergreen}{\textbf{90.2}} & 89.0 & \cellcolor{softergreen}{\underline{90.1}} \\
ResNet101 & 81.9 & 79.9 & 68.2 & 23.1 & 75.8 & 83.7 & 56.4 & 82.8 & 85.5 & 76.7 & 74.9 & 83.0 & 79.2 & \underline{88.0} & \cellcolor{softergreen}{\textbf{89.5}} & 87.0 & 85.2 \\
ResNet152 & 82.3 & 80.4 & 67.6 & 21.0 & 75.8 & 84.2 & 56.0 & 82.7 & 85.9 & 77.6 & 77.7 & 83.6 & 80.6 & \underline{88.5} & \cellcolor{softergreen}{\textbf{90.0}} & 87.7 & 86.0 \\
ResNet50 & 80.9 & 77.5 & 52.6 & 27.6 & 73.3 & 82.0 & 60.8 & 79.7 & 81.5 & 73.3 & 65.1 & 81.9 & 77.4 & \underline{86.5} & \cellcolor{softergreen}{\textbf{87.5}} & 84.6 & 83.1 \\
ResNet50-supcon & 78.7 & 85.2 & 87.7 & \underline{88.7} & 87.6 & 82.3 & 88.6 & 86.4 & 88.6 & 87.8 & 78.9 & 86.7 & 87.1 & 52.9 & \cellcolor{softergreen}{\textbf{89.2}} & 76.5 & \cellcolor{softergreen}{85.6} \\
ViT-T16-In21k-augreg & 75.5 & 79.9 & 85.5 & \textbf{86.9} & 85.1 & 86.5 & 45.7 & 75.9 & 80.0 & 86.4 & 84.1 & 83.9 & 85.5 & 84.3 & \cellcolor{softergreen}{\underline{86.5}} & 83.8 & \cellcolor{softergreen}{84.0} \\
ViT-S16-In21k-augreg & 81.4 & 84.4 & 90.4 & 89.9 & 90.0 & \textbf{91.6} & 66.5 & 85.3 & 88.2 & 90.8 & 88.8 & 89.1 & 89.2 & 90.4 & \cellcolor{softergreen}{\underline{91.5}} & 89.1 & \cellcolor{softergreen}{89.4} \\
ViT-B16-In21k-augreg2 & 85.1 & 84.4 & 84.6 & 87.3 & 85.4 & 85.9 & 24.6 & 86.8 & 89.1 & 87.6 & 86.0 & 88.6 & 84.5 & 86.7 & \cellcolor{softergreen}{\textbf{90.3}} & 88.6 & \cellcolor{softergreen}{\underline{90.0}} \\
ViT-B16-In21k-augreg & 84.5 & 87.5 & 91.9 & 91.3 & 91.7 & \textbf{93.6} & 54.8 & 85.6 & 89.1 & 92.4 & 90.1 & 91.6 & 91.4 & \underline{93.4} & 92.5 & 92.0 & 91.9 \\
ViT-B16-In21k-orig & 81.8 & 88.0 & 93.2 & 93.1 & 92.7 & \underline{93.5} & 77.1 & 90.4 & 91.9 & 93.3 & 87.7 & 91.1 & 92.0 & 92.7 & \cellcolor{softergreen}{\textbf{93.5}} & 91.6 & \cellcolor{softergreen}{91.8} \\
ViT-B16-In21k-miil & 84.3 & 87.7 & 90.6 & 91.0 & 90.1 & \underline{91.8} & 32.1 & 88.4 & 91.1 & 91.7 & 86.6 & 91.0 & 89.1 & 89.6 & \cellcolor{softergreen}{\textbf{92.6}} & 90.2 & \cellcolor{softergreen}{91.2} \\
ViT-L16-In21k-augreg & 85.8 & 89.6 & 93.0 & 94.1 & 92.9 & \underline{94.6} & 58.1 & 87.8 & 90.6 & 93.5 & 89.9 & 93.0 & 93.1 & \textbf{94.9} & 94.0 & 93.9 & 93.7 \\
ViT-L16-In21k-orig & 81.5 & 89.5 & 93.3 & 93.3 & 93.0 & 93.4 & 86.2 & 92.0 & 92.9 & \underline{93.5} & 87.3 & 91.9 & 92.5 & 92.8 & \cellcolor{softergreen}{\textbf{93.8}} & 92.2 & \cellcolor{softergreen}{92.4} \\
ViT-S16-augreg & 78.8 & 82.3 & 84.9 & 84.6 & 84.8 & 80.4 & 46.7 & 77.9 & 81.8 & 84.6 & 83.9 & 85.1 & 83.5 & 86.9 & 86.8 & \underline{87.4} & \cellcolor{softergreen}{\textbf{87.4}} \\
ViT-B16-augreg & 79.2 & 82.7 & 85.8 & 86.4 & 85.7 & 84.1 & 52.9 & 81.2 & 84.0 & 85.6 & 84.5 & 85.6 & 84.8 & 87.4 & \cellcolor{softergreen}{87.6} & \underline{88.0} & \cellcolor{softergreen}{\textbf{88.1}} \\
ViT-B16-CLIP-L2b-In12k & 86.2 & 87.9 & 85.8 & 87.8 & 87.1 & 92.3 & 19.5 & 90.4 & \underline{92.5} & 89.6 & 91.2 & 90.8 & 90.6 & 90.5 & \cellcolor{softergreen}{\textbf{93.3}} & 91.2 & \cellcolor{softergreen}{92.2} \\
ViT-L14-CLIP-L2b-In12k & 88.2 & 90.5 & 89.9 & 90.6 & 90.4 & \underline{94.6} & 38.2 & 92.2 & 94.1 & 90.9 & 92.3 & 92.6 & 92.7 & 93.6 & \cellcolor{softergreen}{\textbf{94.8}} & 93.7 & \cellcolor{softergreen}{94.0} \\
ViT-H14-CLIP-L2b-In12k & 88.6 & 90.6 & 89.4 & 90.1 & 90.2 & \underline{94.2} & 27.6 & 92.0 & 93.6 & 90.8 & 88.9 & 92.7 & 92.1 & 93.7 & \cellcolor{softergreen}{\textbf{94.7}} & 93.7 & \cellcolor{softergreen}{94.0} \\
ViT-so400M-SigLip & 89.4 & 87.6 & 79.7 & 85.8 & 83.8 & 92.6 & 25.7 & 91.8 & 93.1 & 85.7 & 88.8 & 91.6 & 92.0 & 93.2 & \cellcolor{softergreen}{\textbf{93.8}} & 93.4 & \cellcolor{softergreen}{\underline{93.6}} \\
\hline
\textbf{Average} & 84.4 & 85.0 & 81.5 & 81.0 & 84.4 & 89.1 & 40.4 & 86.6 & 88.5 & 86.2 & 86.2 & 88.4 & 87.5 & 89.1 & \cellcolor{softergreen}{\textbf{91.2}} & 89.8 & \cellcolor{softergreen}{\underline{90.4}} \\
\end{tabular}

    \label{tab:AUC-openood}
\end{table*}

\setlength{\tabcolsep}{4pt}
\begin{table*}[h!]
\centering
\scriptsize 
    
     \caption{AUC on NINCO datasets, \colorbox{softergreen}{Green} indicates that normalized method is better than its unnormalized counterpart, \textbf{bold} indicates the best method, and \underline{underlined} indicates second best method. Maha++ improves over Maha on average by 2.6\% in AUC over all models. Similarly, rMaha++ is 1.0\% better in AUC than rMaha. In total, Maha++ improves the SOTA compared to the previously strongest methods rMaha by 1.0\%, which is significant. The highest AUC is achieved by Maha++ for the EVA02-L14-M38m-In21k highlighted in {\color{blue}blue}.  }
     \vspace{1mm}
        \begin{tabular}{lccccccccccccccccc}
Model & Val Acc & MSP & E & E+R & ML & ViM & AshS & KNN & NNG & NEC & GMN & GEN & fDBD & Maha & Maha++ & rMaha & rMaha++ \\
\hline
ConvNeXt-B-In21k & 86.3 & 88.2 & 85.7 & 87.8 & 87.6 & 92.5 & 45.4 & 88.0 & 91.5 & 90.7 & 87.7 & 90.9 & 89.5 & 91.2 & \cellcolor{softergreen}{\textbf{94.3}} & 92.3 & \cellcolor{softergreen}{\underline{93.5}} \\
ConvNeXt-B & 84.4 & 81.7 & 64.8 & 73.2 & 76.7 & 83.0 & 25.1 & 81.6 & 85.2 & 78.0 & 81.1 & 83.3 & 83.0 & 85.8 & \cellcolor{softergreen}{\underline{88.5}} & 87.2 & \cellcolor{softergreen}{\textbf{88.8}} \\
ConvNeXtV2-T-In21k & 85.1 & 86.7 & 87.2 & 87.4 & 87.5 & \underline{92.7} & 39.7 & 86.8 & 90.2 & 90.6 & 88.8 & 89.4 & 85.9 & 91.7 & \cellcolor{softergreen}{\textbf{92.9}} & 91.4 & \cellcolor{softergreen}{91.9} \\
ConvNeXtV2-B-In21k & 87.6 & 89.4 & 89.1 & 89.9 & 89.3 & 94.8 & 43.8 & 91.1 & 93.5 & 92.7 & 91.3 & 92.0 & 91.8 & 94.5 & \cellcolor{softergreen}{\textbf{95.6}} & 94.5 & \cellcolor{softergreen}{\underline{95.0}} \\
ConvNeXtV2-L-In21k & 88.2 & 90.2 & 89.3 & 90.7 & 89.7 & 93.9 & 39.8 & 91.6 & 93.9 & 93.4 & 92.6 & 92.3 & 92.8 & 94.1 & \cellcolor{softergreen}{\textbf{96.2}} & 94.7 & \cellcolor{softergreen}{\underline{95.5}} \\
ConvNeXtV2-T & 83.5 & 82.9 & 76.2 & 80.3 & 80.5 & 83.5 & 26.0 & 78.0 & 83.1 & 82.0 & 81.0 & 85.6 & 82.5 & 85.4 & \cellcolor{softergreen}{\underline{88.3}} & 87.5 & \cellcolor{softergreen}{\textbf{89.0}} \\
ConvNeXtV2-B & 85.5 & 82.6 & 73.2 & 78.2 & 79.2 & 83.6 & 20.0 & 83.7 & 86.6 & 79.8 & 82.9 & 86.1 & 85.2 & 87.9 & \cellcolor{softergreen}{\underline{90.1}} & 89.0 & \cellcolor{softergreen}{\textbf{90.3}} \\
ConvNeXtV2-L & 86.1 & 82.2 & 71.7 & 78.6 & 78.0 & 80.8 & 19.1 & 84.8 & 87.1 & 77.5 & 84.2 & 86.4 & 86.1 & 88.7 & \cellcolor{softergreen}{\underline{90.2}} & 89.9 & \cellcolor{softergreen}{\textbf{90.6}} \\
DeiT3-S16-In21k & 84.8 & 77.9 & 74.8 & 77.2 & 76.2 & 85.4 & 23.2 & 83.4 & 85.4 & 77.6 & 87.6 & 82.7 & 84.2 & 87.3 & \cellcolor{softergreen}{\textbf{89.2}} & 87.6 & \cellcolor{softergreen}{\underline{89.0}} \\
DeiT3-B16-In21k & 86.7 & 81.5 & 74.5 & 80.5 & 77.5 & 89.3 & 21.3 & 87.6 & 89.2 & 81.3 & 89.6 & 86.6 & 87.5 & 90.2 & \cellcolor{softergreen}{\underline{91.8}} & 90.6 & \cellcolor{softergreen}{\textbf{91.9}} \\
DeiT3-L16-In21k & 87.7 & 83.9 & 81.9 & 86.3 & 82.7 & 90.9 & 25.3 & 89.6 & 90.8 & 87.4 & 90.6 & 89.1 & 88.7 & 91.6 & \cellcolor{softergreen}{\underline{92.8}} & 91.9 & \cellcolor{softergreen}{\textbf{92.9}} \\
DeiT3-S16 & 83.4 & 82.9 & 81.4 & 81.2 & 83.1 & 86.3 & 64.6 & 81.1 & 85.0 & 83.0 & 84.4 & 86.4 & 85.1 & 87.5 & \cellcolor{softergreen}{\underline{88.6}} & 88.4 & \cellcolor{softergreen}{\textbf{89.3}} \\
DeiT3-B16 & 85.1 & 82.2 & 66.6 & 62.4 & 76.7 & 85.1 & 28.4 & 80.5 & 77.1 & 77.3 & 83.9 & 86.4 & 84.0 & 87.2 & \cellcolor{softergreen}{\underline{88.3}} & 88.3 & \cellcolor{softergreen}{\textbf{89.2}} \\
DeiT3-L16 & 85.8 & 81.2 & 75.5 & 68.8 & 78.6 & 84.7 & 62.3 & 81.1 & 78.8 & 78.7 & 84.8 & 86.0 & 84.9 & 88.2 & \cellcolor{softergreen}{89.1} & \underline{89.2} & \cellcolor{softergreen}{\textbf{90.1}} \\
EVA02-B14-In21k & 88.7 & 90.6 & 89.7 & 91.0 & 90.3 & \underline{94.7} & 50.1 & 91.1 & 93.2 & 92.7 & 92.6 & 93.2 & 93.0 & 94.0 & \cellcolor{softergreen}{\textbf{95.1}} & 93.9 & \cellcolor{softergreen}{94.6} \\
EVA02-L14-M38m-In21k & 90.1 & 92.6 & 91.1 & 91.8 & 92.1 & \underline{96.1} & 40.9 & 93.2 & 94.6 & 94.2 & 92.0 & 94.7 & 94.8 & 96.0 & \cellcolor{softergreen}{\textbf{\color{blue}96.4}} & 95.9 & \cellcolor{softergreen}{96.0} \\
EVA02-T14 & 80.6 & 79.2 & 75.3 & 75.0 & 77.7 & 83.6 & 41.1 & 78.3 & 81.9 & 81.5 & 83.7 & 82.3 & 78.0 & 84.2 & \cellcolor{softergreen}{\textbf{84.8}} & 84.3 & \cellcolor{softergreen}{\underline{84.7}} \\
EVA02-S14 & 85.7 & 81.8 & 76.1 & 76.5 & 78.8 & 88.6 & 27.0 & 84.3 & 87.1 & 82.8 & \underline{89.6} & 85.0 & 82.8 & 89.2 & \cellcolor{softergreen}{\textbf{89.6}} & 89.3 & \cellcolor{softergreen}{89.6} \\
EffNetV2-S & 83.9 & 81.1 & 71.1 & 77.6 & 76.3 & 81.7 & 20.7 & 83.5 & 84.9 & 78.3 & 80.6 & 84.6 & 83.7 & 85.5 & \cellcolor{softergreen}{87.1} & \underline{88.1} & \cellcolor{softergreen}{\textbf{89.4}} \\
EffNetV2-L & 85.7 & 82.6 & 73.3 & 80.3 & 79.6 & 80.2 & 20.9 & 84.0 & 85.4 & 79.7 & 85.3 & 86.4 & 84.0 & 87.4 & \cellcolor{softergreen}{89.0} & \underline{89.5} & \cellcolor{softergreen}{\textbf{90.5}} \\
EffNetV2-M & 85.2 & 82.3 & 71.8 & 79.5 & 78.6 & 81.0 & 19.3 & 84.0 & 85.4 & 79.4 & 83.9 & 86.6 & 84.1 & 87.2 & \cellcolor{softergreen}{89.1} & \underline{89.5} & \cellcolor{softergreen}{\textbf{90.6}} \\
Mixer-B16-In21k & 76.6 & 79.3 & 78.1 & 78.1 & 78.8 & 80.8 & 49.8 & 76.2 & 78.6 & 78.9 & 82.3 & 80.9 & 79.1 & 80.5 & \cellcolor{softergreen}{\underline{84.6}} & 83.8 & \cellcolor{softergreen}{\textbf{85.2}} \\
SwinV2-B-In21k & 87.1 & 87.4 & 86.0 & 89.2 & 87.5 & 90.8 & 44.1 & 86.1 & 90.5 & 91.2 & 86.7 & 90.8 & 88.6 & 89.4 & \cellcolor{softergreen}{\textbf{92.9}} & 90.4 & \cellcolor{softergreen}{\underline{92.5}} \\
SwinV2-L-In21k & 87.5 & 88.2 & 85.9 & 89.9 & 87.4 & 90.8 & 36.9 & 87.8 & 91.5 & 91.4 & 87.7 & 91.2 & 89.8 & 90.0 & \cellcolor{softergreen}{\textbf{94.1}} & 91.2 & \cellcolor{softergreen}{\underline{93.5}} \\
SwinV2-S & 84.2 & 80.9 & 74.8 & 79.7 & 78.1 & 83.9 & 17.6 & 80.7 & 83.8 & 80.3 & 84.5 & 84.7 & 82.9 & 86.0 & \cellcolor{softergreen}{\underline{88.7}} & 86.6 & \cellcolor{softergreen}{\textbf{88.9}} \\
SwinV2-B & 84.6 & 80.8 & 74.9 & 80.4 & 78.0 & 81.4 & 28.9 & 82.3 & 84.2 & 79.7 & 84.5 & 85.2 & 83.1 & 86.8 & \cellcolor{softergreen}{\underline{88.5}} & 87.3 & \cellcolor{softergreen}{\textbf{88.7}} \\
ResNet101 & 81.9 & 78.7 & 67.3 & 21.1 & 74.6 & 76.9 & 47.5 & 77.6 & 82.6 & 74.7 & 74.1 & 80.9 & 73.9 & 85.3 & \cellcolor{softergreen}{\textbf{89.1}} & 87.8 & \cellcolor{softergreen}{\underline{88.1}} \\
ResNet152 & 82.3 & 79.1 & 67.2 & 20.0 & 74.6 & 77.3 & 48.7 & 77.6 & 83.1 & 75.6 & 76.4 & 81.4 & 75.7 & 85.9 & \cellcolor{softergreen}{\textbf{89.5}} & 87.9 & \cellcolor{softergreen}{\underline{88.6}} \\
ResNet50 & 80.9 & 76.8 & 55.0 & 23.7 & 73.0 & 73.5 & 52.6 & 75.1 & 78.8 & 72.3 & 66.2 & 79.7 & 71.7 & 82.7 & \cellcolor{softergreen}{\underline{86.5}} & 86.1 & \cellcolor{softergreen}{\textbf{86.6}} \\
ResNet50-supcon & 78.7 & 84.9 & 87.2 & 87.1 & 87.2 & 79.2 & 86.0 & 84.3 & 86.8 & \underline{87.3} & 80.3 & 86.9 & 85.0 & 48.2 & \cellcolor{softergreen}{\textbf{88.1}} & 78.7 & \cellcolor{softergreen}{87.0} \\
ViT-T16-In21k-augreg & 75.5 & 78.1 & 81.0 & 82.1 & 81.3 & 82.2 & 54.4 & 73.3 & 76.2 & 82.3 & 81.0 & 80.8 & 83.0 & \textbf{84.0} & \underline{83.7} & 83.0 & 82.8 \\
ViT-S16-In21k-augreg & 81.4 & 83.4 & 88.6 & 88.4 & 88.5 & 89.7 & 61.1 & 82.0 & 85.6 & 89.2 & 86.7 & 87.7 & 88.4 & \underline{90.7} & \cellcolor{softergreen}{\textbf{90.8}} & 89.5 & 89.4 \\
ViT-B16-In21k-augreg2 & 85.1 & 83.2 & 82.7 & 85.2 & 83.9 & 83.6 & 28.7 & 84.7 & 87.6 & 86.1 & 85.5 & 87.8 & 82.8 & 86.1 & \cellcolor{softergreen}{\underline{90.7}} & 88.6 & \cellcolor{softergreen}{\textbf{90.8}} \\
ViT-B16-In21k-augreg & 84.5 & 86.3 & 91.1 & 90.3 & 91.0 & 92.4 & 56.2 & 82.8 & 87.5 & 91.5 & 89.3 & 91.1 & 90.3 & \textbf{94.1} & \underline{93.2} & 93.2 & 93.0 \\
ViT-B16-In21k-orig & 81.8 & 87.2 & 91.8 & 91.8 & 91.5 & 92.6 & 71.0 & 88.3 & 90.1 & 92.1 & 87.3 & 90.3 & 91.2 & \underline{93.1} & \cellcolor{softergreen}{\textbf{93.8}} & 92.7 & \cellcolor{softergreen}{92.9} \\
ViT-B16-In21k-miil & 84.3 & 86.5 & 88.1 & 88.6 & 88.0 & 91.2 & 31.3 & 87.2 & 89.7 & 90.0 & 86.5 & 89.8 & 88.2 & 89.9 & \cellcolor{softergreen}{\textbf{93.1}} & 91.1 & \cellcolor{softergreen}{\underline{92.2}} \\
ViT-L16-In21k-augreg & 85.8 & 89.5 & 92.7 & 94.9 & 92.7 & 93.9 & 48.1 & 83.9 & 88.8 & 93.2 & 88.6 & 93.2 & 92.4 & \textbf{95.3} & 94.4 & \underline{95.1} & 94.8 \\
ViT-L16-In21k-orig & 81.5 & 89.3 & 91.7 & 91.8 & 91.6 & 91.9 & 81.7 & 89.7 & 91.3 & 92.1 & 86.1 & 91.1 & 91.6 & 92.4 & \cellcolor{softergreen}{\textbf{93.5}} & 92.9 & \cellcolor{softergreen}{\underline{93.1}} \\
ViT-S16-augreg & 78.8 & 80.8 & 81.8 & 81.5 & 82.2 & 74.8 & 44.4 & 73.8 & 78.2 & 81.9 & 81.4 & 83.0 & 81.0 & 85.3 & 85.1 & \textbf{86.6} & \underline{86.6} \\
ViT-B16-augreg & 79.2 & 81.0 & 83.3 & 83.8 & 83.4 & 81.0 & 52.4 & 77.8 & 81.1 & 83.3 & 82.0 & 83.9 & 82.5 & 85.7 & \cellcolor{softergreen}{85.7} & \underline{86.9} & \cellcolor{softergreen}{\textbf{86.9}} \\
ViT-B16-CLIP-L2b-In12k & 86.2 & 86.6 & 82.9 & 85.5 & 84.9 & 91.1 & 21.5 & 88.4 & 91.1 & 87.9 & 90.6 & 89.6 & 90.1 & 89.4 & \cellcolor{softergreen}{\textbf{92.9}} & 90.6 & \cellcolor{softergreen}{\underline{92.3}} \\
ViT-L14-CLIP-L2b-In12k & 88.2 & 90.0 & 87.9 & 89.0 & 89.0 & 94.4 & 33.4 & 91.2 & 93.3 & 89.6 & 92.6 & 92.2 & 93.4 & 93.9 & \cellcolor{softergreen}{\textbf{95.2}} & 94.2 & \cellcolor{softergreen}{\underline{94.8}} \\
ViT-H14-CLIP-L2b-In12k & 88.6 & 89.7 & 87.2 & 88.2 & 88.5 & 93.9 & 27.6 & 91.0 & 92.4 & 89.2 & 89.8 & 91.8 & 92.7 & 94.2 & \cellcolor{softergreen}{\textbf{95.3}} & 94.3 & \cellcolor{softergreen}{\underline{94.8}} \\
ViT-so400M-SigLip & 89.4 & 87.3 & 79.5 & 85.1 & 83.2 & 92.1 & 26.1 & 91.8 & 93.3 & 84.7 & 88.3 & 91.7 & 92.4 & 93.3 & \cellcolor{softergreen}{\underline{94.6}} & 94.2 & \cellcolor{softergreen}{\textbf{94.8}} \\
\hline
\textbf{Average} & 84.4 & 84.1 & 80.2 & 79.3 & 83.1 & 86.6 & 39.9 & 84.1 & 86.7 & 84.6 & 85.4 & 87.3 & 85.9 & 88.1 & \cellcolor{softergreen}{\textbf{90.7}} & 89.7 & \cellcolor{softergreen}{\underline{90.7}} \\
\end{tabular}

    \label{tab:auc-ninco}
\end{table*}

\setlength{\tabcolsep}{4pt}
\begin{table*}[h!]   
\centering
\scriptsize 
    
    \caption{FPR on NINCO for cosine-based methods, \colorbox{softergreen}{Green} indicates that the normalized method is better than its unnormalized counterpart, \textbf{bold} indicates the best method, and \underline{underlined} indicates the second best method. \ourMethod{} consistently outperforms other cosine-based methods. In only 2 out of 44 models, another method (once NNguide and once Cosine) is better than Maha++. }
    \vspace{1mm}
        \begin{tabular}{ll|rrrrr}
\toprule
                    Model & Accuracy & Maha++ &  Cosine &  KNN \cite{sun2022knnood} &       NNguide \cite{park2023nnguide} &  SSC \cite{techapanurak2020hyperparameter} \\
\midrule

            ConvNeXt-B &   84.434 &     \textbf{50.5} &          60.6 & 70.1 &          62.2 & 69.3 \\
      ConvNeXt-B-In21k &   86.270 &     \textbf{28.8} &          42.2 & 51.6 &          41.2 & 51.3 \\
          ConvNeXtV2-B &   85.474 &     \textbf{44.7} &          57.1 & 67.3 &          60.3 & 66.6 \\
    ConvNeXtV2-B-In21k &   87.642 &     \textbf{22.4} &          31.1 & 40.9 &          31.7 & 40.4 \\
          ConvNeXtV2-L &   86.120 &     \textbf{43.0} &          53.8 & 62.4 &          56.9 & 59.7 \\
    ConvNeXtV2-L-In21k &   88.196 &     \textbf{18.4} &          27.0 & 38.7 &          29.9 & 39.2 \\
          ConvNeXtV2-T &   83.462 &     \textbf{52.3} &          69.4 & 82.3 &          73.9 & 72.8 \\
    ConvNeXtV2-T-In21k &   85.104 &     \textbf{32.8} &          45.8 & 54.1 &          45.0 & 55.1 \\
             DeiT3-B16 &   85.074 &     \textbf{57.2} &          67.1 & 74.5 &          80.1 & 65.7 \\
       DeiT3-B16-In21k &   86.744 &     \textbf{38.8} &          46.9 & 52.6 &          46.4 & 53.2 \\
             DeiT3-L16 &   85.812 &     \textbf{50.4} &          62.2 & 67.2 &          77.9 & 68.7 \\
       DeiT3-L16-In21k &   87.722 &     \textbf{33.9} &          38.9 & 43.8 &          37.8 & 46.3 \\
             DeiT3-S16 &   83.434 &     \textbf{53.5} &          67.6 & 75.6 &          57.8 & 63.6 \\
       DeiT3-S16-In21k &   84.826 &     \textbf{50.8} &          58.4 & 62.9 &          59.3 & 65.2 \\
       EVA02-B14-In21k &   88.694 &     \textbf{23.8} &          29.1 & 37.6 &          30.0 & 34.4 \\
  EVA02-L14-M38m-In21k &   90.054 &     \textbf{18.6} &          22.7 & 30.3 &          26.1 & 28.8 \\
             EVA02-S14 &   85.720 &     \textbf{48.0} &          53.6 & 60.0 &          54.0 & 63.8 \\
             EVA02-T14 &   80.630 &     \textbf{64.0} &          69.8 & 74.5 &          71.0 & 75.5 \\
       Mixer-B16-In21k &   76.598 &     \textbf{65.4} &          78.2 & 85.8 &          83.7 & 79.5 \\
             ResNet101 &   81.890 &     \textbf{50.4} &          61.5 & 74.9 &          66.4 & 87.2 \\
             ResNet152 &   82.286 &     \textbf{46.5} &          62.1 & 72.0 &          61.6 & 85.8 \\
              ResNet50 &   80.856 &     \textbf{61.0} &          64.0 & 83.7 &          75.0 & 88.2 \\
       ResNet50-supcon &   78.686 &              59.6 &          58.9 & 65.8 & \textbf{58.4} & 74.1 \\
        SwinV2-B-In21k &   87.096 &     \textbf{31.3} &          45.9 & 57.2 &          42.7 & 53.7 \\
              SwinV2-B &   84.604 &     \textbf{52.2} &          63.4 & 69.4 &          65.2 & 71.6 \\
        SwinV2-L-In21k &   87.468 &     \textbf{28.3} &          42.8 & 55.1 &          41.7 & 53.6 \\
              SwinV2-S &   84.220 &     \textbf{49.8} &          65.2 & 73.1 &          66.8 & 75.2 \\
            EffNetV2-L &   85.664 &     \textbf{47.8} &          57.0 & 62.5 &          60.1 & 62.4 \\
            EffNetV2-M &   85.204 &     \textbf{50.0} &          58.0 & 63.1 &          60.6 & 64.4 \\
            EffNetV2-S &   83.896 &              59.9 & \textbf{58.8} & 60.9 &          59.6 & 68.7 \\
 ViT-B16-In21k-augreg2 &   85.096 &     \textbf{45.9} &          56.4 & 64.0 &          57.0 & 64.8 \\
        ViT-B16-augreg &   79.152 &     \textbf{61.3} &          73.1 & 77.6 &          75.9 & 75.7 \\
  ViT-B16-In21k-augreg &   84.528 &     \textbf{35.7} &          53.4 & 67.7 &          59.0 & 54.0 \\
    ViT-B16-In21k-orig &   81.790 &     \textbf{31.6} &          46.0 & 52.7 &          47.6 & 45.2 \\
    ViT-B16-In21k-miil &   84.268 &     \textbf{35.4} &          49.7 & 59.6 &          51.7 & 62.1 \\
ViT-B16-CLIP-L2b-In12k &   86.172 &     \textbf{35.8} &          43.5 & 49.4 &          42.3 & 48.8 \\
ViT-H14-CLIP-L2b-In12k &   88.588 &     \textbf{23.7} &          31.5 & 41.7 &          27.4 & 36.5 \\
ViT-L14-CLIP-L2b-In12k &   88.178 &     \textbf{25.4} &          30.9 & 39.5 & \textbf{25.4} & 33.3 \\
  ViT-L16-In21k-augreg &   85.840 &     \textbf{28.9} &          50.0 & 68.6 &          58.9 & 48.2 \\
    ViT-L16-In21k-orig &   81.508 &     \textbf{32.4} &          39.2 & 45.8 &          40.7 & 42.0 \\
        ViT-S16-augreg &   78.842 &     \textbf{63.1} &          75.8 & 82.1 &          80.0 & 78.1 \\
  ViT-S16-In21k-augreg &   81.388 &     \textbf{44.6} &          60.3 & 70.9 &          64.0 & 61.5 \\
     ViT-so400M-SigLip &   89.406 &     \textbf{27.4} &          29.0 & 36.3 &          30.0 & 35.6 \\
  ViT-T16-In21k-augreg &   75.466 &     \textbf{63.2} &          77.2 & 81.7 &          81.9 & 74.2 \\
\bottomrule
\end{tabular}

\label{tab:cosine-methods-fpr}
\end{table*}

\setlength{\tabcolsep}{4pt}
\begin{table*}[h!]
\centering
\scriptsize 
 \caption{AUROC for CIFAR10, \colorbox{softergreen}{Green} indicates that the normalized method is better than its unnormalized counterpart, \textbf{bold} indicates the best method, and \underline{underlined} indicates the second best method. Maha++ is clearly the best method. Only for the WRN28-10 Maha is better (but not significantly). Maha++ improves in all cases over the previously beset methods ViM.  We highlight the best AUC achieved by Maha++ for the ViT-B16-21k-1k in {\color{blue}blue}.}
 \vspace{1mm}
\resizebox{\textwidth}{!}{%
    \begin{tabular}{lcccccccccccccccccc}

Model & Ash & Dice & Ebo & KlM & KNN & ML & MSP & O-Max & React & She & NNguide & T-Scal & ViM & Neco & rMD & rMD++ & MD & MD++ \\
\hline
SwinV2-S-1k & 69.96 & 92.85 & 95.61 & 98.04 & 99.25 & 95.83 & 96.60 & 97.02 & 96.83 & 96.88 & 67.51 & 96.61 & \underline{99.53} & 98.86 & 98.83 & 98.79 & 99.50 & \cellcolor{softergreen}{\textbf{99.57}} \\
ViT-B16-21k-1k & 82.75 & 99.33 & 99.42 & 96.98 & 99.64 & 99.41 & 98.88 & 97.66 & 99.45 & 98.99 & 87.30 & 99.06 & \underline{99.67} & 99.56 & 99.03 & \cellcolor{softergreen}{99.04} & 99.60 & \cellcolor{softergreen}{\color{blue}\textbf{99.71}} \\
RN18 & 87.15 & 89.60 & 91.09 & 79.62 & \underline{91.58} & 90.97 & 89.93 & 89.04 & 90.78 & 87.62 & 63.57 & 90.32 & 91.12 & 90.67 & 89.92 & \cellcolor{softergreen}{90.06} & 86.87 & \cellcolor{softergreen}{\textbf{91.69}} \\
RN34 & 78.29 & 84.84 & 87.26 & 82.75 & 92.15 & 87.20 & 88.11 & 87.38 & 87.50 & 81.40 & 55.07 & 88.07 & \underline{92.50} & 86.39 & 90.34 & \cellcolor{softergreen}{90.49} & 91.53 & \cellcolor{softergreen}{\textbf{93.61}} \\
RNxt29-32 & 78.33 & 71.90 & 88.45 & 83.19 & 90.46 & 88.20 & 87.98 & 85.65 & 85.27 & 87.90 & 29.57 & 87.97 & \underline{91.36} & 89.62 & 89.84 & 88.69 & 90.70 & \cellcolor{softergreen}{\textbf{91.56}} \\
\hline
Average & 79.29 & 87.70 & 92.37 & 88.11 & 94.62 & 92.32 & 92.30 & 91.35 & 91.97 & 90.56 & 60.60 & 92.41 & \underline{94.84} & 93.02 & 93.59 & 93.41 & 93.64 & \cellcolor{softergreen}{\textbf{95.23}} \\
\hline
RN50-SC & --- & --- & --- & --- & \underline{96.76} & --- & --- & --- & --- & --- & --- & --- & --- & --- & 94.46 & 94.30 & 59.00 & \cellcolor{softergreen}{\textbf{96.80}} \\
RN34-SC & --- & --- & --- & --- & \underline{96.15} & --- & --- & --- & --- & --- & --- & --- & --- & --- & 94.72 & 94.24 & 64.21 & \cellcolor{softergreen}{\textbf{96.77}} \\
\end{tabular}

    }
   
    \label{tab:auc-cifar10}
\end{table*}

\setlength{\tabcolsep}{4pt}
\begin{table*}[h!]
\centering
\scriptsize 
 \caption{FPR for CIFAR10, \colorbox{softergreen}{Green} indicates that the normalized method is better than its unnormalized counterpart, \textbf{bold} indicates the best method, and \underline{underlined} indicates the second best method. Maha++ is the best method on average. We highlight the best FPR achieved by Maha++ for the ViT-B16-21k-1k in {\color{blue}blue}.}
 \vspace{1mm}
\resizebox{\textwidth}{!}{%

    \begin{tabular}{lcccccccccccccccccc}

Model & Ash & Dice & Ebo & KlM & KNN & ML & MSP & O-Max & React & She & NNguide & T-Scal & ViM & Neco & rMD & rMD++ & MD & MD++ \\
\hline
SwinV2-S-1k & 93.43 & 19.51 & 8.82 & 6.02 & 4.03 & 8.07 & 6.74 & 5.97 & 7.21 & 12.08 & 63.18 & 6.73 & \underline{2.17} & 3.66 & 3.42 & \cellcolor{softergreen}{3.18} & 2.35 & \cellcolor{softergreen}{\textbf{2.16}} \\
ViT-B16-21k-1k & 60.41 & 2.42 & 1.93 & 6.23 & 1.75 & 1.97 & 3.27 & 3.15 & 1.99 & 4.48 & 41.88 & 2.91 & \underline{1.29} & 1.57 & 2.59 & 2.66 & 1.66 & \cellcolor{softergreen}{\color{blue}\textbf{1.24}} \\
RN18 & 47.52 & 41.18 & \textbf{39.22} & 56.48 & 45.55 & 40.28 & 56.42 & 76.47 & \underline{40.22} & 45.89 & 77.65 & 52.58 & 51.99 & 41.10 & 52.51 & 54.09 & 69.46 & \cellcolor{softergreen}{46.15} \\
RN34 & 46.12 & 44.20 & \textbf{38.05} & 52.89 & 46.24 & 39.14 & 51.99 & 78.47 & 42.36 & 45.81 & 76.79 & 48.98 & 48.15 & 41.93 & 52.67 & 53.67 & 54.45 & \cellcolor{softergreen}{\underline{38.36}} \\
RNxt29-32 & 97.43 & 63.57 & 47.36 & 56.54 & 55.91 & 50.41 & 53.15 & 89.85 & 56.26 & 38.13 & 99.97 & 53.36 & \underline{36.13} & 51.32 & 58.07 & 61.31 & 41.17 & \cellcolor{softergreen}{\textbf{34.64}} \\
\hline
Average & 68.98 & 34.18 & \underline{27.08} & 35.63 & 30.70 & 27.97 & 34.31 & 50.78 & 29.61 & 29.28 & 71.89 & 32.91 & 27.95 & 27.92 & 33.85 & 34.98 & 33.82 & \cellcolor{softergreen}{\textbf{24.51}} \\
\hline
RN50-SC & --- & --- &--- & --- & \underline{19.48} & --- & --- & --- & --- & --- & --- & --- & --- & --- & 33.23 & 35.26 & 81.77 & \cellcolor{softergreen}{\textbf{18.59}} \\
RN34-SC & --- & --- & --- & --- & \underline{22.47} & --- & --- & --- & --- & --- & --- & --- & --- & --- & 30.52 & 32.89 & 78.65 & \cellcolor{softergreen}{\textbf{17.55}} \\

\end{tabular}

    }
   
    \label{tab:fpr-cifar10}
\end{table*}

\setlength{\tabcolsep}{4pt}
\begin{table*}[h!]
\centering
\scriptsize 
\caption{AUROC for CIFAR100, \colorbox{softergreen}{Green} indicates that the normalized method is better than its unnormalized counterpart, \textbf{bold} indicates the best method, and \underline{underlined} indicates the second best method. Maha++ is clearly the best method. Only for the RNxt29-32 She is slightly better. Maha++ improves in all cases over the previously best methods ViM, Maha and KNN.  We highlight the best AUC achieved by Maha++ for the ViT-B32-21k in {\color{blue}blue}.}
\vspace{1mm}
\resizebox{\textwidth}{!}{%
    \begin{tabular}{lcccccccccccccccccc}

Model & Ash & Dice & Ebo & KlM & KNN & ML & MSP & O-Max & React & She & NNguide & T-Scal & ViM & Neco & rMD & rMD++ & MD & MD++ \\
\hline
SwinV2-S-1k & 48.67 & 63.60 & 84.72 & 82.52 & 90.06 & 85.20 & 85.68 & 85.82 & 87.53 & 89.66 & 71.36 & 85.93 & \underline{91.34} & 90.38 & 89.77 & 89.30 & 90.29 & \cellcolor{softergreen}{\textbf{92.99}} \\
Deit3-S-21k & 49.99 & 44.78 & 85.69 & 81.59 & 88.06 & 86.18 & 86.21 & 84.52 & 88.88 & 87.47 & 55.19 & 86.43 & \underline{90.41} & 89.86 & 87.44 & \cellcolor{softergreen}{87.85} & 88.30 & \cellcolor{softergreen}{\textbf{90.54}} \\
ConvN-T-21k & 63.80 & 53.50 & 77.76 & 80.48 & 86.60 & 78.51 & 79.09 & 82.60 & 80.17 & 82.94 & 62.98 & 79.22 & 87.67 & 81.31 & 85.00 & 84.89 & \underline{87.95} & \cellcolor{softergreen}{\textbf{89.55}} \\
ViT-B32-21k & 59.23 & 88.31 & 90.28 & 89.13 & 94.87 & 89.99 & 85.36 & 88.00 & 88.59 & 94.10 & 87.17 & 86.73 & 94.62 & 90.70 & 92.37 & \cellcolor{softergreen}{92.82} & \underline{95.59} & \cellcolor{softergreen}{\color{blue}\textbf{96.84}} \\
ViT-S16-21k & 65.78 & 84.35 & 89.85 & 84.23 & 93.97 & 89.44 & 83.87 & 88.09 & 88.45 & 92.32 & 80.08 & 85.38 & \underline{95.91} & 90.80 & 93.09 & \cellcolor{softergreen}{93.32} & 95.63 & \cellcolor{softergreen}{\textbf{96.81}} \\
RN18 & 74.20 & 79.77 & 80.31 & 74.11 & 81.22 & 80.31 & 79.70 & 68.22 & 80.27 & 79.18 & 81.06 & 80.02 & 78.50 & 80.66 & \underline{81.27} & 80.91 & 78.46 & \cellcolor{softergreen}{\textbf{81.71}} \\
RN34 & 65.82 & 78.86 & 79.88 & 75.63 & 81.51 & 79.76 & 79.13 & 73.14 & 80.48 & 77.15 & 74.13 & 79.56 & \underline{82.13} & 80.61 & 81.22 & 80.94 & 82.03 & \cellcolor{softergreen}{\textbf{82.16}} \\
RNxt29-32 & 79.46 & 82.01 & 78.58 & 70.79 & 80.89 & 78.47 & 78.37 & 66.11 & 78.36 & \textbf{82.59} & 73.21 & 78.22 & 75.33 & 79.68 & 76.87 & \cellcolor{softergreen}{77.06} & 76.18 & \cellcolor{softergreen}{\underline{82.48}} \\
\hline
Average & 63.37 & 71.90 & 83.38 & 79.81 & \underline{87.15} & 83.48 & 82.18 & 79.56 & 84.09 & 85.68 & 73.15 & 82.69 & 86.99 & 85.50 & 85.88 & \cellcolor{softergreen}{85.89} & 86.80 & \cellcolor{softergreen}{\textbf{89.14}} \\
\hline
RN34-SC & --- & --- & --- & --- & \underline{83.76} & --- & --- & --- & --- & --- & --- & --- & --- & --- & 76.80 & \cellcolor{softergreen}{80.03} & 53.30 & \cellcolor{softergreen}{\textbf{84.83}} \\
RN50-SC & --- & --- & --- & --- & \underline{82.41} & --- &---& --- & --- & --- & --- & --- & --- & --- & 77.90 & \cellcolor{softergreen}{79.67} & 59.01 & \cellcolor{softergreen}{\textbf{82.44}} \\

\end{tabular}
}

\label{tab:auc-cifar100}
\end{table*}

\setlength{\tabcolsep}{4pt}
\begin{table*}[h!]
\centering
\scriptsize 
  \caption{FPR for CIFAR100, \colorbox{softergreen}{Green} indicates that the normalized method is better than its unnormalized counterpart, \textbf{bold} indicates the best method, and \underline{underlined} indicates the second best method. Maha++ is improving in all cases over Maha and is on average the best method. We highlight the best FPR achieved by Maha++ for the ViT-S16-21k in {\color{blue}blue}.}
  \vspace{1mm}
\resizebox{\textwidth}{!}{%
    \begin{tabular}{lcccccccccccccccccc}

Model & Ash & Dice & Ebo & KlM & KNN & ML & MSP & O-Max & React & She & NNguide & T-Scal & ViM & Neco & rMD & rMD++ & MD & MD++ \\
\hline
SwinV2-S-1k & 92.66 & 75.98 & 40.95 & 49.65 & 36.27 & 40.96 & 47.28 & 67.04 & 39.54 & 39.64 & 80.29 & 45.58 & 34.02 & \underline{33.59} & 41.40 & 47.14 & 40.10 & \cellcolor{softergreen}{\textbf{26.01}} \\
Deit3-S-21k & 94.47 & 96.34 & 41.61 & 47.86 & \underline{36.81} & 42.37 & 48.92 & 66.00 & 40.46 & 40.93 & 96.35 & 47.15 & 39.99 & 37.12 & 41.02 & 41.36 & 41.99 & \cellcolor{softergreen}{\textbf{31.72}} \\
ConvN-T-21k & 92.11 & 89.10 & 57.67 & 65.50 & \underline{51.16} & 57.44 & 60.60 & 66.86 & 58.23 & 53.76 & 91.25 & 60.04 & 51.18 & 53.92 & 62.79 & \cellcolor{softergreen}{61.66} & 52.48 & \cellcolor{softergreen}{\textbf{42.69}} \\
ViT-B32-21k & 93.98 & 46.59 & 30.51 & 43.24 & 26.49 & 31.28 & 48.02 & 53.68 & 32.53 & 33.25 & 64.86 & 40.74 & 27.14 & 28.61 & 33.80 & \cellcolor{softergreen}{31.03} & \underline{26.28} & \cellcolor{softergreen}{\textbf{18.94}} \\
ViT-S16-21k & 80.45 & 56.38 & 36.06 & 50.09 & 31.91 & 37.63 & 52.17 & 57.38 & 36.48 & 38.89 & 77.85 & 46.68 & \underline{24.90} & 33.24 & 34.10 & \cellcolor{softergreen}{32.83} & 25.51 & \cellcolor{softergreen}{\color{blue}\textbf{18.58}} \\
RN18 & 78.98 & 80.53 & 80.19 & 78.85 & 76.61 & 79.87 & 80.59 & 97.36 & 80.18 & 80.46 & \textbf{68.16} & 80.25 & 79.61 & 79.89 & 76.14 & 77.49 & 79.48 & \cellcolor{softergreen}{\underline{72.92}} \\
RN34 & 78.27 & 78.31 & 75.19 & 78.08 & \underline{74.44} & 75.33 & 76.93 & 94.07 & 74.51 & 78.76 & 75.07 & 76.20 & 77.17 & \textbf{74.25} & 75.82 & 76.22 & 76.63 & \cellcolor{softergreen}{74.51} \\
RNxt29-32 & 72.59 & \textbf{67.03} & 82.22 & 87.56 & 73.17 & 82.30 & 82.31 & 96.32 & 81.87 & 69.42 & 81.89 & 82.60 & 76.40 & 80.54 & 86.58 & \cellcolor{softergreen}{84.39} & 77.67 & \cellcolor{softergreen}{\underline{67.71}} \\
\hline
Average & 85.44 & 73.78 & 55.55 & 62.60 & \underline{50.86} & 55.90 & 62.10 & 74.84 & 55.47 & 54.39 & 79.47 & 59.90 & 51.30 & 52.65 & 56.46 & 56.51 & 52.52 & \cellcolor{softergreen}{\textbf{44.13}} \\
\hline
RN34-SC & --- & --- & --- & --- & \underline{66.87} & --- & --- & --- & --- & --- & --- & --- & --- & --- & 90.02 & \cellcolor{softergreen}{74.37} & 93.76 & \cellcolor{softergreen}{\textbf{63.51}} \\
RN50-SC & --- & --- & --- & --- & \textbf{66.69} & --- & --- & --- & --- & --- & --- & --- & --- & --- & 83.53 & \cellcolor{softergreen}{78.15} & 82.38 & \cellcolor{softergreen}{\underline{67.95}} \\
\end{tabular}

    }
  
    \label{tab:fpr-cifar100-long}
\end{table*}

\setlength{\tabcolsep}{8pt}
\begin{table}[]
    \centering
    \caption{\textbf{Normalization improves robustness against noise distributions.} We report the number of failed unit tests (noise distributions with FPR values $\geq10\%$) from \cite{bitterwolf2023ninco}. Normalization improves the brittleness of Mahalanobis-based detectors.}
    \label{tab:unit-tests-big}
    \begin{tabular}{llc}
\toprule
                      model & Maha & Maha++ \\
\midrule
                 ConvNeXt-B &          16 &       \textbf{15} \\
           ConvNeXt-B-In21k &           4 &        \textbf{0} \\
               ConvNeXtV2-B &          14 &        \textbf{6} \\
         ConvNeXtV2-B-In21k &           5 &        \textbf{0} \\
               ConvNeXtV2-L &          13 &        \textbf{4} \\
         ConvNeXtV2-L-In21k &           2 &        \textbf{0} \\
               ConvNeXtV2-T &          17 &        \textbf{9} \\
         ConvNeXtV2-T-In21k &           6 &        \textbf{0} \\
                  DeiT3-B16 &  \textbf{14} &               15 \\
            DeiT3-B16-In21k &           6 &        \textbf{3} \\
                  DeiT3-L16 &   \textbf{8} &        \textbf{8} \\
            DeiT3-L16-In21k &           1 &        \textbf{0} \\
                  DeiT3-S16 &          15 &       \textbf{10} \\
            DeiT3-S16-In21k &          17 &       \textbf{11} \\
            EVA02-B14-In21k &           3 &        \textbf{0} \\
       EVA02-L14-M38m-In21k &   \textbf{0} &        \textbf{0} \\
                  EVA02-S14 &           8 &        \textbf{0} \\
                  EVA02-T14 &          11 &        \textbf{0} \\
            Mixer-B16-In21k &          17 &       \textbf{10} \\
                  ResNet101 &   \textbf{0} &                1 \\
                  ResNet152 &   \textbf{0} &        \textbf{0} \\
                   ResNet50 &   \textbf{0} &                1 \\
            ResNet50-supcon &          17 &        \textbf{0} \\
             SwinV2-B-In21k &          10 &        \textbf{0} \\
                   SwinV2-B &          12 &                6 \\
                   SwinV2-S &          15 &                4 \\
                 EffNetV2-L &          13 &        \textbf{7} \\
                 EffNetV2-M &          13 &        \textbf{4} \\
                 EffNetV2-S &          11 &        \textbf{3} \\
  ViT-B16-224-In21k-augreg2 &          16 &        \textbf{7} \\
         ViT-B16-224-augreg &          11 &        \textbf{4} \\
     ViT-B16-224-In21k-orig &           2 &        \textbf{0} \\
     ViT-B16-224-In21k-miil &          17 &        \textbf{0} \\
     ViT-B16-CLIP-L2b-In12k &          14 &        \textbf{0} \\
     ViT-H14-CLIP-L2b-In12k &           4 &        \textbf{0} \\
     ViT-L14-CLIP-L2b-In12k &           7 &        \textbf{0} \\
     ViT-L16-224-In21k-orig &           5 &        \textbf{0} \\
         ViT-S16-224-augreg &           2 &        \textbf{1} \\
          ViT-so400M-SigLip &           8 &        \textbf{0} \\
\bottomrule
\end{tabular}

\end{table}

\begin{table}[h]\caption{\textbf{Comparison to SSD+. } SSD+ consists of a) training with contrastive loss (implicitly normalizing the features), b) estimating cluster means in the normalized feature space via k-means, c) centering the train features with the closest class mean and estimating a shared covariance matrix, and d) using the Mahalanobis distance at inference time for OOD detection. SSD+ is therefore not readily applicable as post-hoc OOD detection method. To highlight the benefits of post-hoc methods, we report the performance of SSD+ with a ResNet-50, which was trained for 700 epochs with supervised contrastive loss, and compare it to a ConvNext model and an EVA model with varied pretraining schemes. The latter models outperform SSD+ clearly, underlining the importance of post-hoc methods for OOD detection.}
\centering
\begin{tabular}{l c}
\hline
\textbf{Model} & \textbf{FPR (\%)} \\
\hline
SSD+ w. 100 clusters     & 66.0 \\
SSD+ w. 500 clusters     & 65.7 \\
SSD+ w. 1000 clusters    & 67.8 \\
CnvNxtV2-L + Maha++      & 18.4 \\
EVA02-L14 + Maha++       & 18.6 \\
\hline
\end{tabular}

\label{tab:ssd-comparison}
\end{table}

\FloatBarrier
\section{Models}\label{app:models}
\begin{table}
    \centering
    \caption{Imagenet model checkpoints.}
\begin{tabular}{lll}
 \\
\hline
Modelname & Checkpoint & source \\
\midrule
ViT-B16-In21k-augreg & vit\_base\_patch16\_224.augreg\_in21k\_ft\_in1k & timm / huggingface \\
ViT-L16-In21k-augreg & vit\_large\_patch16\_224.augreg\_in21k\_ft\_in1k & timm / huggingface \\
ViT-T16-In21k-augreg & vit\_tiny\_patch16\_224.augreg\_in21k\_ft\_in1k & timm / huggingface \\
ViT-S16-In21k-augreg & vit\_small\_patch16\_224.augreg\_in21k\_ft\_in1k & timm / huggingface \\
ViT-B16-augreg & vit\_base\_patch16\_224.augreg\_in1k & timm / huggingface \\
ViT-S16-augreg & vit\_small\_patch16\_224.augreg\_in1k & timm / huggingface \\
ViT-so400M-SigLip & vit\_so400m\_patch14\_siglip\_378.webli\_ft\_in1k & timm / huggingface \\
ViT-H14-CLIP-L2b-In12k & vit\_huge\_patch14\_clip\_336.laion2b\_ft\_in12k\_in1k & timm / huggingface \\
ViT-L14-CLIP-L2b-In12k & vit\_large\_patch14\_clip\_336.laion2b\_ft\_in12k\_in1k & timm / huggingface \\
ViT-B16-In21k-orig & vit\_base\_patch16\_224.orig\_in21k\_ft\_in1k & timm / huggingface \\
ViT-L16-In21k-orig & vit\_large\_patch32\_384.orig\_in21k\_ft\_in1k & timm / huggingface \\
ViT-B16-In21k-miil & vit\_base\_patch16\_224\_miil.in21k\_ft\_in1k & timm / huggingface \\
ViT-B16-In21k-augreg2 & vit\_base\_patch16\_224.augreg2\_in21k\_ft\_in1k & timm / huggingface \\
ViT-B16-CLIP-L2b-In12k & vit\_base\_patch16\_clip\_224.laion2b\_ft\_in12k\_in1k & timm / huggingface \\
EVA02-L14-M38m-In21k & eva02\_large\_patch14\_448.mim\_m38m\_ft\_in22k\_in1k & timm / huggingface \\
EVA02-B14-In21k & eva02\_base\_patch14\_448.mim\_in22k\_ft\_in22k\_in1k & timm / huggingface \\
EVA02-S14 & eva02\_small\_patch14\_336.mim\_in22k\_ft\_in1k & timm / huggingface \\
EVA02-T14 & eva02\_tiny\_patch14\_336.mim\_in22k\_ft\_in1k & timm / huggingface \\
DeiT3-B16 & deit3\_base\_patch16\_224 & timm / huggingface \\
DeiT3-B16-In21k & deit3\_base\_patch16\_224\_in21ft1k & timm / huggingface \\
DeiT3-L16-In21k & deit3\_large\_patch16\_384.fb\_in22k\_ft\_in1k & timm / huggingface \\
DeiT3-B16-In21k & deit3\_base\_patch16\_384.fb\_in22k\_ft\_in1k & timm / huggingface \\
DeiT3-L16 & deit3\_large\_patch16\_384.fb\_in1k & timm / huggingface \\
DeiT3-B16 & deit3\_base\_patch16\_384.fb\_in1k & timm / huggingface \\
DeiT3-S16-In21k & deit3\_small\_patch16\_384.fb\_in22k\_ft\_in1k & timm / huggingface \\
DeiT3-S16 & deit3\_small\_patch16\_384.fb\_in1k & timm / huggingface \\
SwinV2-S & swinv2\_small\_window16\_256.ms\_in1k & timm / huggingface \\
SwinV2-B & swinv2\_base\_window16\_256.ms\_in1k & timm / huggingface \\
SwinV2-L-In21k & swinv2\_large\_window12to24\_192to384.ms\_in22k\_ft\_in1k & timm / huggingface \\
SwinV2-B-In21k & swinv2\_base\_window12to24\_192to384.ms\_in22k\_ft\_in1k & timm / huggingface \\
ResNet50 & resnet50.tv2\_in1k & timm / huggingface \\
ResNet101 & resnet101.tv2\_in1k & timm / huggingface \\
ResNet152 & resnet152.tv2\_in1k & timm / huggingface \\
ResNet50-supcon & rn50supcon & \hskip -.4in \tiny{\url{github.com/roomo7time/nnguide/} } \\
ConvNeXt-B & convnext\_base.fb\_in1k & timm / huggingface \\
ConvNeXt-B-In21k & convnext\_base.fb\_in22k\_ft\_in1k & timm / huggingface \\
ConvNeXtV2-L-In21k & convnextv2\_large.fcmae\_ft\_in22k\_in1k\_384 & timm / huggingface \\
ConvNeXtV2-B-In21k & convnextv2\_base.fcmae\_ft\_in22k\_in1k\_384 & timm / huggingface \\
ConvNeXtV2-T-In21k & convnextv2\_tiny.fcmae\_ft\_in22k\_in1k\_384 & timm / huggingface \\
ConvNeXtV2-T & convnextv2\_tiny.fcmae\_ft\_in1k & timm / huggingface \\
ConvNeXtV2-B & convnextv2\_base.fcmae\_ft\_in1k & timm / huggingface \\
ConvNeXtV2-L & convnextv2\_large.fcmae\_ft\_in1k & timm / huggingface \\
Mixer-B16-In21k & mixer\_b16\_224.goog\_in21k\_ft\_in1k & timm / huggingface \\
EffNetV2-M & tf\_efficientnetv2\_m.in1k & timm / huggingface \\
EffNetV2-S & tf\_efficientnetv2\_s.in1k & timm / huggingface \\
EffNetV2-L & tf\_efficientnetv2\_l.in1k & timm / huggingface \\
\bottomrule
\end{tabular}

    \label{tab:models-imagenet}
\end{table}

\begin{table}
    \centering
    \caption{Cifar model checkpoints.}
\begin{tabular}{ll}
 \\
\hline

SwinV2-S-1k & ft from timm model \\
Deit3-S-21k & ft from timm model \\
ConvN-T-21k & ft from timm model \\
ViT-B32-21k & https://github.com/google-research/big\_vision \\
ViT-S16-21k & https://github.com/google-research/big\_vision \\
RN18 & https://huggingface.co/edadaltocg/ \\
RN34 & https://huggingface.co/edadaltocg/ \\
RN34-SC & https://huggingface.co/edadaltocg/ \\
RN50-SC & https://huggingface.co/edadaltocg/ \\
RNxt29-32 & self trained \\
\end{tabular}

    \label{tab:models-cifar}
\end{table}
\FloatBarrier
\section{Methods}\label{sec:methods}

We describe OOD detection methods evaluated in our work. Let a neural network \( n_{\theta}(x) = g(\phi(x)) \) decompose into a feature extractor \(\phi\) and linear layer $ g(\phi_i) = \mathbf{W}^T \phi_i + \mathbf{b} $. For input \( x \), \(\phi(x)\) denotes the feature embedding, and \( g(\phi(x)) \) produces logits \(\mathbf{o}\), which can be transformed to a probability vector \(\mathbf{p}\) via the softmax function.

\textbf{MSP} \cite{hendrycks2017MSP}: Classifer confidence, i.e. max-softmax-probability
\begin{equation*}
s=\max_{c}({p}_{c})
\end{equation*}

\textbf{Max-Logit} (ML or MLS) \cite{hendrycks22Scaling}: Max-Logit returns the largest entry of the logit-vector $\mathbf{o}$, i.e.  
\begin{equation*}
s=\max_{c}({o}_{c})
\end{equation*}
\textbf{Energy} (E) \cite{liu2020energy}:
Energy or log-sum-exp of logits:
\begin{equation*}
s=\log\sum_{c}^{C}\exp{(o_{c})}
\end{equation*}
\textbf{KL-Matching} (KLM) \cite{hendrycks22Scaling}: KL-Matching computes the average probability vector $\mathbf{d}_c$ for each of the $C$ classes. For a test input, the KL-distances of all $\mathbf{d}_c$ vectors to its probability vector $\mathbf{p}$ are computed, and the OOD-score is the negative of the smallest of those distances:
\begin{equation*}
s=-\min_{c}\text{KL}[\mathbf{p}||\mathbf{d}_{c}]
\end{equation*}
In the original paper by \cite{hendrycks22Scaling}, the average for $\mathbf{d}_{c}$ is computed over an additional validation set. Since none of the other methods leverages extra data and we are interested in fair comparison, we deploy KL-Matching like in \cite{wang2022vim,yang2022openood}, where the average is computed over the train set.

\textbf{KNN} (KNN) \cite{sun2022knnood}: Computes the k-nearest neighbour in the normalized feature-space: 
The feature vector of a test input is normalized to $\mathbf{z}=\mathbf{\phi}/||\mathbf{\phi}||_{2}$ and the pairwise distances $r_{i}(\mathbf{z})=||\mathbf{z}-\mathbf{z}_{i}||_{2}$ to the normalized features $\mathcal{Z}=\{\mathbf{z}_{1},...,\mathbf{z}_{N}\}$ of all samples of the training set are computed.
The distances $r_{i}(\mathbf{z})$ are then sorted according to their magnitude and the $K^{\text{th}}$ smallest distance, denoted $r^{K}(\mathbf{z})$ is used as negative OOD-score:
\begin{equation*}
s=-r^{K}(\mathbf{z})
\end{equation*}
Like suggested in \cite{sun2022knnood}, we use $K=1000$.\\

\textbf{ReAct} (E+R) \cite{sun2021react}: The authors propose to perform a truncation of the feature vector, ${\Bar{\phi}}=\min(\phi,r)$, where the $\min$ operation is to be understood element-wise and $r$ is the truncation threshold.  The truncated features can then be converted to so-called rectified logits via $\mathbf{\Bar{o}}=g({\Bar{\phi}})=\mathbf{W}^{T}{\Bar{\phi}}+\mathbf{b}$. While the rectified logits can now be used with a variety of existing detection methods, we follow \cite{sun2021react} and use the rectified Energy as OOD-score:
\begin{equation*}
s=\log\sum_{c}^{C}\exp{(\Bar{o}_{c})}
\end{equation*}
As suggested in \cite{wang2022vim}, we set the threshold $r$ such that 1\% of the activations from the train set would be truncated.

\textbf{Virtual Logit Matching} (ViM) \cite{wang2022vim}: The idea behind ViM is that meaningful features are thought to lie in a low-dimensional manifold, called the principal space $P$, whereas features from OOD-samples should also lie in $P^{\perp}$, the space orthogonal to $P$. $P$ is the $D$-dimensional subspace spanned by the eigenvectors with the largest $D$ eigenvalues of the matrix $\mathbf{F}^{T}\mathbf{F}$, where $\mathbf{F}$ is the matrix of all train features offsetted by $\mathbf{u}=-(\mathbf{W^{T}})^{+}\mathbf{b}$ ($+$ denotes the Moore-Penrose inverse). A sample with feature vector $\phi$ is then also offset to $\mathbf{\Tilde{h}}=\phi-\mathbf{u}$ and can be decomposed into $\mathbf{\Tilde{h}}=\mathbf{\Tilde{h}}^{P}+\mathbf{\Tilde{h}}^{P^{\perp}}$, and $\mathbf{\Tilde{h}}^{P^{\perp}}$ is referred to as the \textit{Residual} of $\phi$. ViM leverages the Residual and converts it to a virtual logit $o_{0}=\alpha||\mathbf{\Tilde{h}}^{P^{\perp}}||_{2}$, where 
\begin{equation*}
\alpha=\frac{\sum_{i=1}^{N}\max_{c}o_{i}^{c}}{\sum_{i=1}^{N}||\phi^{P^{\perp}}_{i}||_{2}}
\end{equation*}
is designed to match the scale of the virtual logit to the scale of the real train logits. The virtual logit is then appended to the original logits of the test sample, i.e. to $\mathbf{o}$, and a new probability vector is computed via the softmax function. The probability corresponding to the virtual logit is then the final OOD-score:
\begin{equation*}\label{vim}
s=-\frac{\exp{(o_0)}}{\sum_{c=1}^{C}\exp{(o_{c})}+\exp{(o_0)}}
\end{equation*}
Like suggested in \cite{wang2022vim}, we use $D=1000$ if the dimensionality of the feature space $d$ is $d\geq2048$, $D=512$ if $2048\geq d\geq 768$, and $D=d/2$ rounded to integers otherwise. 

\textbf{Cosine} (Cos) \cite{techapanurak2020hyperparameter,anonymous2023COOD}: This method computes the maximum cosine-similarity between the features of a test-sample and embedding vectors $\Tilde{\mathbf{u}}_{c}$ (sometimes also called concept-vector):
\begin{equation}
    s=\max_{c}\frac{\Tilde{\mathbf{u}}_{c}^{T}\phi}{||\Tilde{\mathbf{u}}_{c}^{T}||_{2}||\phi||_{2}}
\end{equation}

\textbf{Ash} (Ash) \cite{djurisic2023ash}: Ash applies activation shaping at inference time by pruning acitvations below a certain threshold, and then binarizing (Ash-b) or scaling (Ash-s) the remaining activations, which are then processed as usually in the network. As suggested by the authors, we apply ash to the pre-logit feature activations. 

\textbf{Softmax-scaled-Cosine} (SSC) \cite{tack2020csi}:
Normalize the rows of the weight matrix $\mathbf{w}_i$ and the features, and compute the cosine between the two:
\[\cos \theta_i \equiv \frac{\mathbf{w}_i \cdot \mathbf{\phi}}{\|\mathbf{w}_i\| \|\mathbf{\phi}\|}
\]
Then scale by a scalar $t$ and apply the softmax, to finally use the max-softmax as OOD score:
\[
s=\max_i \left(\text{softmax}(t*\theta)_i\right) \]
In \citet{tack2020csi} the scalr $s$ is learned, for our post-hoc setup we set $s=1$.

\textbf{NeCo} (Nec) \cite{ammar2023neco}: Compute the covariance matrix of the feature space, and project to the $d$ eigenvectors with largest eigenvalues with the corresponding projection matrix $P$. The difference in norm of the projected features and the original features is then scaled with the max-logit and serves as OOD score. 
\[
s = \frac{\|P \phi (x)\|}{\|\phi (x)\|}*\max_i o_i = \sqrt{\frac{\phi (x)^\top P P^\top \phi (x)}{\phi (x)^\top \phi (x)}}*\max_i o_i
\]
Like suggested by the authors, we standardize data and select $d$ such that 90\% of the train variance are explained.  

\textbf{Gaussian Mixture Model} (GMM or GMN) \cite{MukhotiCVPR23DDU}: Estimate a Gaussian mixture model on the train features $\phi(x)$, and use the log-probabilities as OOD score. We use GMN for Gaussian mixture model with normalized features, and GMM for Gaussian mixture model with regular features. 

\textbf{NNguide} (NNG) \cite{park2023nnguide}: "Guide" the energy score by a nearest-neighbor score: 
\[
s=s_{Energy}*s_{KNN}
\]
where $s_{KNN}$ is a KNN score in the normalized feature space, estimated on a subset of the train features. Like suggested by the authors, we use $1\%$ of the train features and $K=10$ neighbors for ImageNet experiments. We also tried $K=1000$, as increasing $K$ showed promising results in an ablation by the authors (Figure 4 in the paper), but found that it performs worse on average than $K=10$. 

\textbf{Relative Mahalanobis distance} (rMaha) \cite{RenRelMaha2021}: A modification of the Mahalanobis distance method, thought to improve near-OOD detection, is to additionally fit a global Gaussian distribution to the train set without taking class-information into account: 
\begin{equation*}
\hat{\mu}_{\text{global}}=\frac{1}{N}\sum_{i}\phi_{i},\hspace{15pt} \hat{\mathbf{\Sigma}}_{\text{global}}=\frac{1}{N}\sum_{i}(\phi_{i}-\hat{\mu}_{\text{global}})(\phi_{i}-\hat{\mu}_{\text{global}})^{T}
\end{equation*}
The OOD-score is then defined as the difference between the original Mahalanobis distance and the Mahalanobis distance w.r.t. the global Gaussian distribution:
\begin{equation*}
s=-\min_{c}\left( (\phi-\hat{\mu}_{c})\hat{\mathbf{\Sigma}}^{-1}(\phi-\hat{\mu}_{c})^{T}-(\phi-\hat{\mu}_{\text{global}})\hat{\mathbf{\Sigma}}_{\text{global}}^{-1}(\phi-\hat{\mu}_{\text{global}})^{T}\right)
\end{equation*}

\end{document}